\documentclass{article} 
\usepackage{times}

\usepackage{graphicx}
\usepackage{amsmath,amssymb,amsfonts,amsthm}
\usepackage[ruled,vlined,algo2e]{algorithm2e}
\usepackage{hyperref}
\usepackage{url}
\usepackage{tabularx}
\usepackage{booktabs}
\usepackage{todonotes}
\usepackage{multirow}
\usepackage{makecell}
\usepackage{float}

\usepackage{cleveref}
\usepackage{multirow}
\usepackage{array}
\usepackage{arydshln}
\usepackage[dvipsnames]{xcolor} 
\usepackage{pifont}
\usepackage[english]{babel}
\usepackage{listings}
\usepackage{tabularray}
\usepackage{fancyhdr}
\usepackage{caption}
\usepackage{subcaption}
\usepackage{colortbl}
\usepackage{enumitem}
\usepackage{microtype}
\usepackage{threeparttable}
\usepackage{wrapfig,epsfig}
\usepackage{pifont}
\usepackage{stmaryrd}
\usepackage{mathtools}
\usepackage{tabularray}
\usepackage{xcolor}
\usepackage[parfill]{parskip}
\usepackage[citestyle=authoryear-comp,sorting=nyt,maxbibnames=99,backend=bibtex]{biblatex}
\newcommand{\citep}{\parencite}
\newcommand{\citet}{\textcite}
\usepackage{amsmath,amsfonts,bm}

\newcommand{\Algname}{SonicMoE}
\newcommand{\algname}{SonicMoE}

\newcommand{\round}[1]{\ensuremath{\lfloor#1\rceil}_{\tileM}}
\newcommand{\roundup}[1]{\ensuremath{\lceil#1\rceil}_{\tileM}}
\newcommand{\rounddown}[1]{\ensuremath{\lfloor#1\rfloor}_{\tileM}}

\newcommand{\model}[1]{\emph{#1}}

\newcommand{\tflops}{TFLOPS}

\newcommand{\squishlist}{
  \begin{list}{$\bullet$}
    { \setlength{\itemsep}{0.3em}      \setlength{\parsep}{3pt}
      \setlength{\topsep}{3pt}       \setlength{\partopsep}{0pt}
      \setlength{\leftmargin}{1.5em} \setlength{\labelwidth}{1em}
      \setlength{\labelsep}{0.5em} } }
\newcommand{\reallysquishlist}{
  \begin{list}{$\bullet$}
    { \setlength{\itemsep}{0.3em}    \setlength{\parsep}{0pt}
      \setlength{\topsep}{0pt}     \setlength{\partopsep}{0pt}
      \setlength{\leftmargin}{0.2em} \setlength{\labelwidth}{0.2em}
      \setlength{\labelsep}{0.2em} } }

 \newcommand{\squishend}{
     \end{list} 
 }

\definecolor{lightplot0}{HTML}{5dc4d4} 
\definecolor{plot0}{HTML}{1F77B4} 
\definecolor{plot1}{HTML}{FF7F0E} 
\definecolor{plot2}{HTML}{2CA02C} 
\definecolor{plot3}{HTML}{D62728} 
\definecolor{plot4}{HTML}{9467BD} 
\definecolor{plot5}{HTML}{8C564B} 
\definecolor{plot6}{HTML}{E377C2} 
\definecolor{plot7}{HTML}{7F7F7F} 
\definecolor{plot8}{HTML}{BCBD22} 
\definecolor{plot9}{HTML}{17BECF}

\newcommand{\moe}[1]{#1}
\newcommand{\note}[1]{\textcolor{gray}{\footnotesize #1}}

\def\eqref#1{equation~\ref{#1}}

\def\1{\bm{1}}

\newcommand{\tileM}{M_\mathrm{tile}}
\newcommand{\tileN}{N_\mathrm{tile}}
\newcommand{\tileK}{K_\mathrm{tile}}

\definecolor{BrightRed}{HTML}{FF0000} 
\definecolor{BrightGreen}{HTML}{00C853}

\newcommand{\cmark}{\textcolor{BrightGreen}{\checkmark}}
\newcommand{\xmark}{\textcolor{BrickRed}{\ding{55}}}

\colorlet{shadecolor}{gray!40}
\newcommand{\grayline}{\arrayrulecolor{shadecolor}\hline\arrayrulecolor{black}}

\newcolumntype{C}[1]{>{\centering\arraybackslash}p{#1}}

\def\vs{{\mathbf{s}}}

\newcommand{\W}[1]{W_{#1}}

\newcommand{\rX}{X}
\newcommand{\bX}{X}
\newcommand{\rY}{Y}
\newcommand{\bY}{Y}
\newcommand{\rS}{S}
\newcommand{\bS}{S}
\newcommand{\rpi}{\pi}
\newcommand{\bpi}{\pi}
\newcommand{\rW}{W}
\newcommand{\bW}{W}

\newcommand{\rA}{A}
\newcommand{\bA}{A}
\newcommand{\rH}{H}
\newcommand{\bH}{H}
\newcommand{\rO}{O}
\newcommand{\bO}{O}
\newcommand{\rdO}{dO}

\DeclareMathAlphabet{\mathsfit}{\encodingdefault}{\sfdefault}{m}{sl}
\SetMathAlphabet{\mathsfit}{bold}{\encodingdefault}{\sfdefault}{bx}{n}

\newcommand{\R}{\mathbb{R}}

\usepackage{subcaption} 

\usepackage{etoolbox} 
\usepackage{authblk}

\usepackage[normalem]{ulem}

\newlength{\defbaselineskip}
\usepackage{wrapfig}
\usepackage{algorithm}
\usepackage[noend]{algpseudocode} 

\usepackage[bottomfloats,belowfloats]{footmisc}

\title{\algname: Accelerating MoE with IO and Tile-aware Optimizations}

\author[$^1$]{Wentao Guo}
\author[$^2$]{Mayank Mishra}
\author[$^1$]{Xinle Cheng}
\author[$^2$]{Ion Stoica}
\author[$^{1,3}$]{Tri Dao}
\affil[$^1$]{Princeton University}
\affil[$^2$]{University of California, Berkeley}
\affil[$^3$]{Together AI}
\affil[ ]{\small{Correspondence to: \texttt{wg0420@princeton.edu}}, {\texttt{tri@tridao.me}}}

\date{\today}

\SetCommentSty{mycommfont}

\newif\ifcomments
\commentstrue
\ifcomments
    \providecommand{\TD}[1]{{\color{magenta}{/* TD: #1 */}}}
    \providecommand{\ion}[1]{{\color{blue}{/* ion: #1 */}}}

\else
    \providecommand{\TD}[1]{}
    \providecommand{\ion}[1]{}    
    
\fi

\begin{document}

\maketitle

\begin{abstract}

Mixture of Experts (MoE) models have emerged as the de facto architecture for scaling up language models without significantly increasing the computational cost. Recent MoE models demonstrate a clear trend towards high expert granularity (smaller expert intermediate dimension) and higher sparsity (constant number of activated experts with a higher number of total experts), which improve model quality per FLOP. However, fine-grained MoEs suffer from increased activation memory footprint and reduced hardware efficiency due to higher IO costs, while sparser MoEs suffer from wasted computations due to padding in Grouped GEMM kernels. In response, we propose a memory-efficient algorithm to compute the forward and backward passes of MoEs with minimal activation caching for the backward pass. We also design GPU kernels that overlap memory IO with computation, benefiting all MoE architectures. Finally, we propose a novel ``token rounding'' method that minimizes the wasted compute due to padding in Grouped GEMM kernels. As a result, our method \Algname\ \textbf{reduces activation memory by 45\% and achieves a 1.86x compute throughput improvement} on Hopper GPUs compared to ScatterMoE's BF16 MoE kernel for a fine-grained 7B MoE. Concretely, \Algname\ on 64 H100s achieves a training throughput of 213 billion tokens per day, comparable to ScatterMoE's 225 billion tokens per day on 96 H100s for a 7B MoE model training with FSDP-2 using the lm-engine codebase\footnote{\small \url{https://github.com/open-lm-engine/lm-engine}}. \textbf{On Blackwell GPUs, \Algname~also achieves a 25\% and 15\% relative speedup on the forward and backward pass respectively compared to a highly optimized DeepGEMM baseline on OLMoE-sized 7B MoE models.} Under high MoE sparsity settings, our tile-aware token rounding algorithm yields an additional \textbf{1.16x} speedup on kernel execution time compared to vanilla top-$K$ routing while maintaining similar downstream performance on Hopper GPUs. We open-source all our kernels\footnote{\small \url{https://github.com/Dao-AILab/sonic-moe}} to enable faster MoE model training.


\end{abstract}


\section{Introduction}
Mixture of Experts (MoE) \citep{shazeer2017outrageouslylargeneuralnetworks} models have emerged as a key technique for scaling up parameters \citep{deepseekv3insights,kimiteam2025kimik2openagentic} without increasing the training computational requirements. Modern transformers often have layers comprised of a sequence mixer block (e.g. Multi-head Attention \citep{vaswani2023attentionneed}), followed by a channel mixer block (e.g. dense MLPs) where MoEs are an excellent substitute for dense MLPs for FLOPs efficiency. An MoE block is typically composed of a token router and multiple smaller and often equal-sized subnetworks, called ``experts". MoEs can reduce FLOPs consumption during training by only activating a subset of all experts per token. However, reducing FLOPs does not directly translate to better hardware utilization since MoE computation features more dynamic IO accesses when each expert needs to gather tokens from different positions, and also scatter the results back to the original positions. Moreover, such hardware-unfriendliness becomes worse as experts become more \emph{granular} (experts have smaller intermediate sizes) and \emph{sparser} (experts are increased while keeping the number of activated experts constant), as shown in Table~\ref{tab:moe-scaling}.







MoE scaling laws~\citep{clark2022unified,krajewski2024scaling,tian2025towards} predict
better model quality per FLOP with increasing expert granularity (ratio between the model's embedding dimension and each expert's intermediate size) and sparsity.
Recent MoE models like DeepSeek V3 \citep{deepseekai2025deepseekv3technicalreport}, Qwen3 MoE \citep{qwen3_2024} and gpt-oss-120b \citep{openai2025gptoss120bgptoss20bmodel}, have demonstrated the superior performance of ``fine-grained" MoEs over ``coarse-grained" MoEs at scale. Besides granularity, the pursuit of MoEs with better model quality while keeping computational requirements constant has also led to modern MoEs becoming sparser. For example, Kimi K2 \citep{kimiteam2025kimik2openagentic} has the same amount of activated parameters as DeepSeek V3 \citep{deepseekai2025deepseekv3technicalreport} but much larger total parameter count. Overall, granularity and sparsity for MoEs have only increased over time as shown in Table~\ref{tab:moe-scaling}. We note that the pursuit of granularity and sparsity is also adopted by recent alternative architectures to MoE such as PEER \citep{he2024mixture}, Memory Layers \citep{berges2024memory}, and Ultra-Mem \citep{huangultra}. 


Though more granular and sparser MoEs increase model quality per FLOP, they suffer from hardware inefficiency due to: (1) larger activation memory footprint for granular MoE models as activation size typically scales linearly with the number of activated experts, (2) lower arithmetic intensity and increased IO cost due to granular experts and (3) wasted computations due to tile quantization effects of grouped GEMM for highly sparse MoEs. The high granularity and sparsity both push MoE training towards the memory-bound regime, requiring carefully designed MoE kernels to hide the increased IO costs. Existing state-of-the-art MoE kernels such as ScatterMoE~\citep{tan2024scattered} and MoMoE~\citep{costin2025momoe} are not designed to handle these high IO costs and they suffer significant training throughput degradation.


\begin{figure}[!ht]
    \centering
    \includegraphics[width=\linewidth]{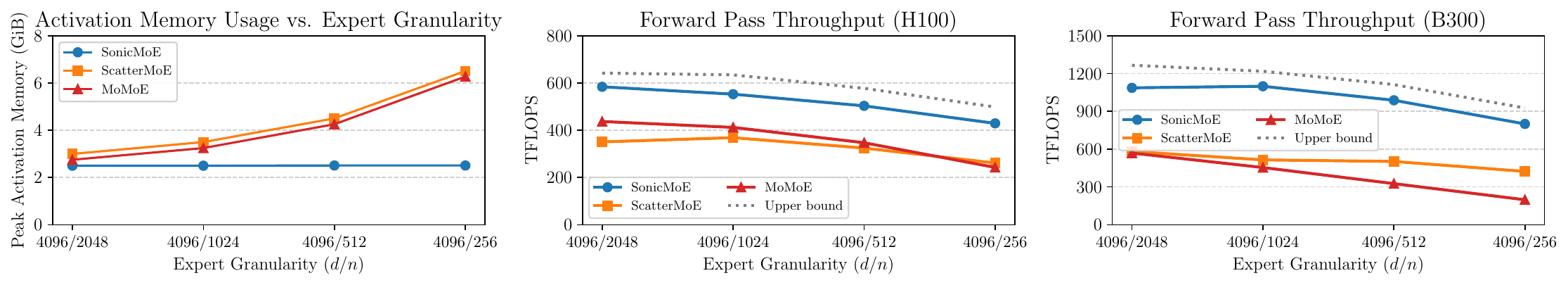}
    \vspace{-2em}
    \caption{\small \Algname's per-layer activation memory footprint (left) stays constant even when expert granularity ($d/n$ where $d$ is the embedding dimension and $n$ is the expert intermediate dimension) increases, and is 20\%-159\% more memory-efficient than other baselines. \Algname's forward computation throughput reaches an average of 88\% of the upper bound (cuBLAS BMM + activation + cuBLAS BMM + aggregation) on both H100 (mid) and B300 (right) GPUs. Note that the cuBLAS upper bound baseline does \emph{not} include the router computation. Here we use a 30B MoE configuration with microbatch size of 32768 tokens for both H100 and B300 GPUs, and we vary the activated experts / total number of experts as 2/32, 4/64, 8/128, and 16/256 from left to right.}
    \label{fig:teasor}
\end{figure}

 We propose to co-design the MoE architecture with a GPU kernel tailored to NVIDIA Blackwell and Hopper generation GPUs and a novel routing method. (1) We derive an algorithm to compute the MoE backward pass more efficiently leading to a much smaller activation memory footprint that does not increase with increasing expert granularity. (2) We leverage new hardware features on Blackwell and Hopper GPUs to overlap memory IO with computation which can benefit all MoEs, and, in particular, fine-grained MoEs. (3) We propose a hardware-aware token rounding routing method where the routed number of tokens to an expert is always a multiple of the GEMM tile size. Using extensive experiments, we show that token rounding routing is 16\% faster than the baseline token-choice routing when we scale up the number of experts 4 times from a 30B MoE. We also validate that TR preserves the MoE inference quality on 1.8B parameter scale. With (1) and (2), we can increase the end-to-end training throughput of a 7B MoE model by 42\% (without changing the top-$K$ token choice routing). Our token rounding routing method further improves training throughput by 16\% when we scale up the number of experts without any accuracy loss.

\textbf{Summary of contributions.} We propose \Algname, a hardware and model architecture co-design solution to address MoE training efficiency problems, making the following contributions:

\begin{itemize}
    \item \textbf{MoE training with minimum possible activation memory footprint without increasing FLOPs: } We analyze the impact of MoE granularity on the MoE layer's forward and backward passes and observe that increasing MoE granularity while maintaining constant FLOPs leads to a linear increase in activation memory required by the backward pass. Leveraging this observation, we carefully redesign the computation graph to avoid caching the activations for the router gradient computation while maintaining the mathematical equivalence to the original MoE formulation. As a result, for a fine-grained 7B MoE, \Algname\ reduces activation memory usage per layer by up to 45\% on H100 GPUs.
    
    \item \textbf{Efficient MoE kernel that overlaps IO with computation to yield SOTA training throughput:} We show that increasing both granularity and sparsity leads to MoEs becoming increasingly memory bandwidth bound. To alleviate this bottleneck, we exploit the asynchrony of the GEMM and IO operations by overlapping them to maximize throughput. For the same fine-grained 7B MoE model on H100 GPUs, our approach increases the \tflops~by 43\% on the forward pass compared to a highly optimized DeepGEMM baseline, and by 83\% and 115\% on the backward pass compared to the state-of-the-art MoE baselines ScatterMoE and MoMoE, respectively. On B300 GPUs, our approach also achieves 25\% more \tflops~on the forward pass and 15\% on the backward pass compared to a highly optimized DeepGEMM baseline on OLMoE-sized 7B MoE models. To evaluate the performance of these techniques, we conduct an extensive performance analysis through comprehensive kernel-level profiling and an IO-aware exploration of the MoE computational paths.

    \item \textbf{Token rounding routing that eliminates wasted FLOPs from sparse MoEs:}
    We introduce a drop-in routing algorithm that rounds the per-expert token counts to multiples of the tile size (e.g., 128) used by grouped GEMM in MoE kernels. This rounding reduces compute wasted on padding while preserving the original token-to-expert assignment as much as possible. The algorithm ensures that, for each expert, the maximum deviation from the original top-$K$ token-choice result is bounded by one tile. This method effectively eliminates padding waste in grouped GEMM while maintaining the same total number of tokens in expectation, and it delivers robust token-choice accuracy even under highly sparse MoE training regimes. We validate the performance of this token-rounding strategy in a 1.4B-parameter sparse training setting, demonstrating that its compute throughput consistently exceeds that of the vanilla top-$K$ token-choice routing. In highly sparse regimes, the improvement reaches up to 16\% higher TFLOPS for end-to-end MoE computation on H100 GPUs.


\end{itemize}



We release \Algname, mainly written in CuTe-DSL~\citep{cutedsl} with a PyTorch interface and a permissive license to benefit researchers and practitioners. The GitHub link is {\small \url{https://github.com/Dao-AILab/sonic-moe}}. A preliminary version of this paper is accepted to ICLR'26: {\small \url{https://openreview.net/pdf?id=KzTJ1raEgB}}.

\section{Background}

We first provide an overview of the MoE architecture and a standard MoE kernel employing grouped GEMM in Section~\ref{sec:background:grouped_gemm}. In Section~\ref{sec:background:arithmetic}, we discuss how granularity and MoE sparsity will affect MoE's training efficiency. We then examine the impact of the MoE routing method on the MoE model quality and training efficiency in Section~\ref{sec:background:routing-methods}.

\subsection{MoE using Grouped GEMM}\label{sec:background:grouped_gemm}

Modern GPUs support Tensor Cores; specialized hardware units with high matrix multiplication throughput \citep{nvidia2022_h100}. A GEMM (general matrix multiply)~\citep{Blas} kernel often has 3 stages: prologue (start input loading), mainloop (keep loading inputs and compute GEMM) and epilogue (miscellaneous IO/math operations on GEMM outputs). The kernel tiles computations (dividing large matrices into small tiles), and optionally pads dimensions so computation aligns with hardware-permissible tile sizes. In this paper, we follow standard GEMM notations in most BLAS~\citep{Blas} libraries: we have $A \in \R^{\mathbf{M}\times \mathbf{K}}$, $B \in \R^{\mathbf{K}\times \mathbf{N}}$, $C \in \R^{\mathbf{M}\times \mathbf{N}}$ for $C = A B$ with problem shape ($\mathbf{M},\mathbf{N},\mathbf{K}$). This notation is adopted by CUTLASS~\citep{nvidia_cutlass} which implements efficient GEMM on CUDA.

\begin{figure}[!ht]
  \centering
  \begin{minipage}[t]{0.46\textwidth}
    \vspace{-2pt} 
    \input{algorithm/swiglu_computation}
  \end{minipage}
  \hfill
  \begin{minipage}[t]{0.52\textwidth}
    \vspace{0pt} 
    \centering
    \includegraphics[width=0.75\linewidth]{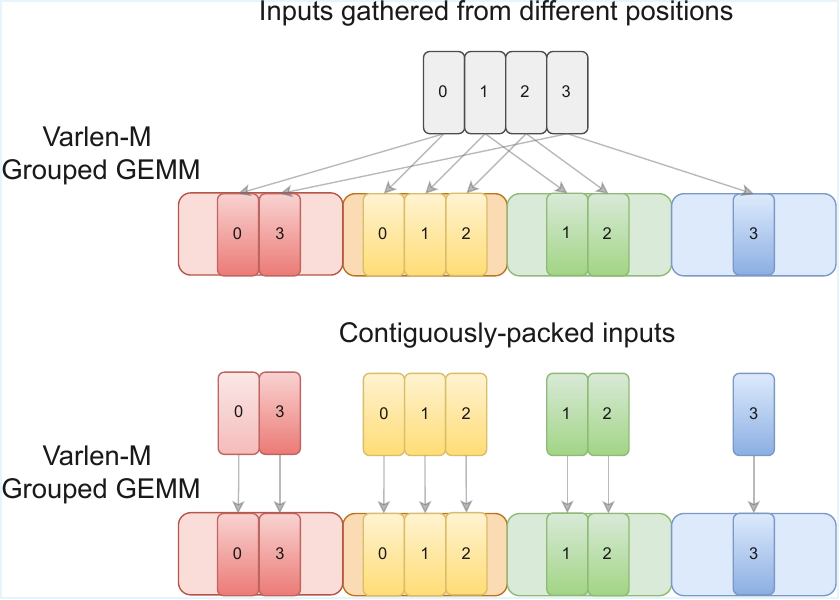}
    \captionof{figure}{\small MoE computation often requires a Grouped GEMM. Each expert gathers inputs from different positions on an input tensor (top) or reads a contiguous chunk on a grouped input array (bottom). This figure is adapted from \citet{tan2024scattered}'s Figure 2. \label{fig:input-format}}
  \end{minipage}
\end{figure}


On both NVIDIA Hopper and Blackwell GPUs, GEMM is performed asynchronously with a producer-consumer paradigm \citep{shah2024flashattention3fastaccurateattention} where producers are dedicated to loading a tile of data from High Bandwidth Memory (HBM), or global memory (GMEM) logically, to shared memory (SMEM) while consumer warpgroups are responsible for GEMM computation~\citep{shah2024flashattention3fastaccurateattention}. In prologue and mainloop, producer warpgroups fetch a tile of data and cache it to a dedicated pipeline while the consumer warpgroups read from the cached tile in this pipeline, perform tiled matrix multiply (MMA) and accumulate over the $\mathbf{K}$ dimension of GEMM. After the mainloop, we enter the epilogue stage where the consumer warpgroups apply post-processing (activation function and write results back to HBM) on the final MMA results.

An MoE block is typically composed of a token router and multiple smaller and often equal-sized subnetworks, called ``experts". The router is responsible for dispatching tokens to the experts which are subsequently used by the specific expert for computation. The outputs from all experts in the layer are then aggregated and passed onto the next layer. MoE computation\footnote{\label{footnote:moe_main_computation}We refer to the computation that decides the activated expert for each token and relevant routing metadata as \textit{\textbf{MoE routing}}, and how each expert processes the routed tokens and expert aggregation as \textit{\textbf{MoE computation}}. Algorithm~\ref{alg:our-forward-moe},~\ref{alg:our-down-bwd-moe}, and~\ref{alg:up-bwd-moe} are \Algname's MoE computation components that are compatible with arbitrary routing algorithms. } can be performed using Grouped GEMM (a list of GEMMs with possibly different $\{\mathbf{M}, \mathbf{N}, \mathbf{K}\}$ dimensions). Algorithm~\ref{alg:moe-swiglu-grouped-gemm} illustrates running MoE forward with Grouped GEMM. 

As shown in Algorithm~\ref{alg:moe-swiglu-grouped-gemm}, during the forward pass (and backward activation gradient computation), we have a variable number of tokens routed to every expert. A Grouped GEMM operation with fixed $(\mathbf{N},\mathbf{K})$ dim (as the expert weight matrix) but variable $\mathbf{M}$ (token dim) is then performed. We refer to this Grouped GEMM as ``varlen-$\mathbf{M}$ Grouped GEMM''. During the backward weight gradient computation, the embedding dimension ($\mathbf{M}$ for backward) and intermediate hidden size ($\mathbf{N}$ for backward) are constant and instead, we reduce over the token dimension ($\mathbf{K}$), which we refer to as ``varlen-$\mathbf{K}$ Grouped GEMM''. For each Grouped GEMM, we often have inputs gathered from different positions or contiguously-packed, as illustrated in Figure~\ref{fig:input-format}. For example in Algorithm~\ref{alg:moe-swiglu-grouped-gemm}, the inputs to up-proj are gathered while the inputs to down-proj are already contiguously-packed.

\subsection{MoE computation}\label{sec:background:arithmetic}

Arithmetic intensity, defined as the ratio of FLOPs over the number of transferred bytes (IO), is a metric to quantify whether a kernel is memory-bound (kernel runtime dominated by memory IO cost) or compute-bound (kernel runtime dominated by compute throughput).

The standard MoE computation for an expert $e$ with SwiGLU activation can be broken down into the following components:

\vspace{-3em}
\begin{align}
    H_e = \text{up-projection}(X_e) = X_e W_{1,e}:& \mathbb{R}^{T_e \times d} \to \mathbb{R}^{T_e \times 2n}\\
    A_e = \text{SwiGLU}(H_e):& \mathbb{R}^{T_e \times 2n} \to \mathbb{R}^{T_e \times n}\\
    Y_e = \text{down-projection}(A_e) = A_e W_{2,e}:& \mathbb{R}^{T_e \times n} \to \mathbb{R}^{T_e \times d}
\end{align}
\vspace{-2em}

where $X_e \in \R^{T_e\times d}$ denotes the input received by expert $e$.


Here, the up-projection uses $2 T_e \cdot 2n \cdot d$ FLOPs and $2 T_e d + 2 \cdot 2n \cdot d + 2 T_e n$ HBM memory transfer bytes (we ignore the writes for $H_e$ here). Similarly, down-projection uses $2 T_e n d$ FLOPs with $2 T_e n + 2 n d + 2 T_e d$  bytes. Defining $\rho = \frac{K}{E}$ as the MoE activation ratio, $G = \frac{d}{n}$ as the granularity and uniform routing i.e., $T_e = T\rho$, the arithmetic intensity (ignoring the writes for $H_e$) for the forward pass of an expert is
\vspace{-.5em}
\begin{align}    \label{eqn:arithmetic_intensity}
    \dfrac{2 T_e \cdot 2n \cdot d +  2 T_e n d}{4 T_e n + 6 n d + 4 T_e d} = \dfrac{3}{\frac{2}{d} + \frac{2}{n} + \frac{3}{T_e}} = \frac{3}{\frac{2 + 2G}{d} + \frac{3}{T\rho}}
\end{align}
\vspace{-1.5em}



For a specific model size (constant $d$), it can be seen that increasing granularity (increasing $G$) or increasing sparsity (decreasing $\rho$) leads to a decreasing arithmetic intensity. This is caused by the linear scaling of IO cost w.r.t. expert granularity. Therefore, for the case of fine-grained MoEs (high $G$)\footnote{Here we refer to ``fine-grained MoE'' as MoE with small intermediate size i.e., $n$ is smaller than $d$. We assume the setting of both iso-FLOPs and iso-params.}, it becomes increasingly important to address the increased IO cost by maximally reducing IO access and hiding IO latency. We examine a memory-efficient MoE kernel design in Section~\ref{sec:moe:memory-efficiency} and discuss techniques to reduce IO access and latency in Section~\ref{sec:moe:hide-io}.



\paragraph{Existing MoE kernel designs.} There are multiple MoE implementations available: ScatterMoE \citep{tan2024scattered}, MoMoE \citep{costin2025momoe}, MegaBlocks \citep{gale2023megablocks}, and Megatron \citep{shoeybi2019megatron}. However, they do not specialize for the setting of fine-grained MoEs that have linearly-increasing IO cost w.r.t. increasing expert granularity. In contrast, our kernel design, \Algname, minimizes the impact of IO cost on the training throughput. \textbf{In Figure~\ref{fig:speed} and~\ref{fig:B300-speed}, we show that when expert granularity $G$ increases, \Algname~demonstrates a greater relative speedup over existing MoE kernel designs due to the IO-aware optimizations.} We elaborate on the technical differences between \Algname~and prior MoE kernels in Appendix~\ref{sec:appendix:kernel_comparison} and include an overview in Table~\ref{tab:kernel-comparison}.

\subsection{MoE routing methods}\label{sec:background:routing-methods}

In MoE, routing determines which experts to activate for each token. Token choice (TC) routing where each token independently selects the activated expert is often the default routing method for MoE models~\citep{shazeer2017outrageouslylargeneuralnetworks}. We often have top-$K$ TC routing where the routing decision for token $t$ is $\mathrm{TopK}_{e\in[E]}(S_{t, e}, K)$ and $S_{t, e}$ is the expert score for token $t$. Besides top-$K$, \citet{huang2024harder} introduce token-choice top-$P$ routing to flexibly allocate compute during training. However, it introduces nondeterminism in the number of activated experts and consumed FLOPs per token. \citet{zeng-etal-2024-adamoe} also propose a similar idea that uses ``null experts'' to dynamically adjust the number of activated experts. 

Besides TC routing, expert choice (EC) routing is developed to avoid load imbalance for expert parallelism \citep{zhou2022mixture} by letting experts choose the tokens. However, EC routing is not directly usable for inference because it is incompatible with autoregressive decoding, and switching back to TC at inference time leads to a mismatch. In addition, EC breaks causality by future token information leakage~\citep{wang2024auxiliary}. To address the inference issue of EC routing, \citet{raposo2024mixture} introduce an auxiliary loss to promote the agreement between TC and EC routing results, or train an auxiliary router to explicitly predict the routing result of EC router and use this auxiliary router during inference.


In this paper, we propose a novel Grouped GEMM tile-aware token rounding method that rounds the number of received tokens per expert (``expert frequency'') to nearby multiples of Grouped GEMM tile sizes and alters at most one tile of tokens per expert. This approach effectively reduces wasted FLOPs caused by Grouped GEMM padding during sparse MoE training while preserving inference quality of trained MoE models. There are similar works that propose to drop and reroute tokens, including Rectify-Router \citep{zeng2024turn}, but they do not focus on the tile structure of Grouped GEMM. Other works such as TMA-adaptive FP8 Grouped GEMM \citep{tmaadatpivefp8} focus on reducing padding-related load traffic but the FLOPs wasted by non-aligned tile size in GEMM computation is not addressed.




\section{Memory-efficient MoE algorithm} \label{sec:moe:memory-efficiency}
\begin{figure}[!ht]
    \centering
    \includegraphics[width=\linewidth]{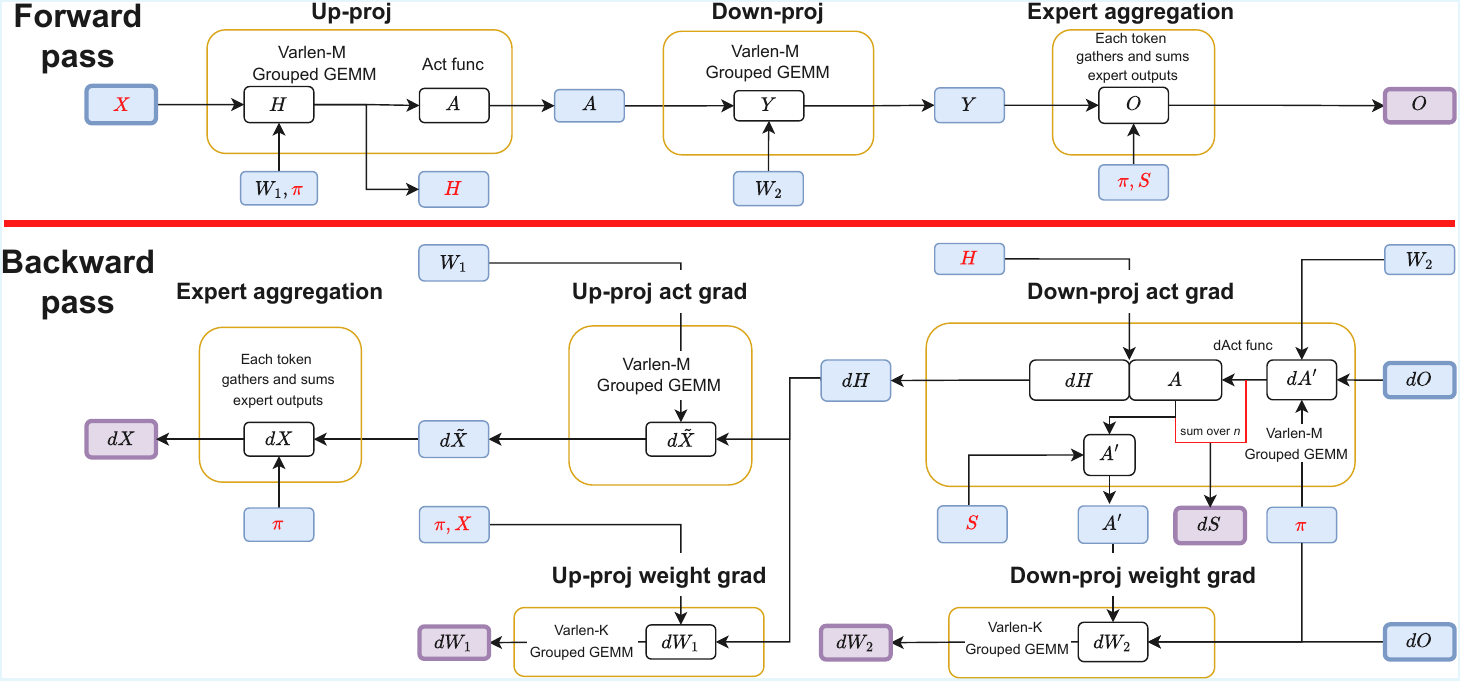}
    \caption{\small Computational workflow of \Algname's 8 launched kernels, grouped by yellow boxes. 3 and 5 kernels are launched during forward and backward computation respectively. The incoming arrows to a yellow circle indicate a variable loaded from HBM to SRAM, and an outgoing arrow represents a variable stored to HBM. We color the boxes of all variables on HBM, with purple boxes indicating the output of forward and backward while blue boxes indicate intermediate variables or weights ($\W{1}$, $\W{2}$). We color all cached activations $X$, $H$, $\pi$, $S$ in red. Algorithm~\ref{alg:our-forward-moe} formally describes \Algname's forward pass, and Algorithm~\ref{alg:our-down-bwd-moe} and~\ref{alg:up-bwd-moe} describe the backward pass.}
    \label{fig:workflow}
\end{figure}

We first describe \algname's high-level kernel design in Section~\ref{sec:high-level} that illustrates \Algname's MoE computation\textsuperscript{\ref{footnote:moe_main_computation}} as shown in Algorithm~\ref{alg:our-forward-moe}, \ref{alg:our-down-bwd-moe}, and \ref{alg:up-bwd-moe}. We then focus on the activation memory usage of \algname~in Section~\ref{sec:mem-efficiency}.

\begin{figure}[t]
\begin{minipage}{0.46\textwidth}
\begin{algorithm}[H]
\caption{\Algname's MoE kernel forward pass. Variables stored in HBM are colored \textcolor{blue}{blue}. $\mathrm{load}$ and $\mathrm{store}$ means load from / store into HBM respectively.}
\DontPrintSemicolon
\label{alg:our-forward-moe}
 \fontsize{8pt}{10pt}\selectfont

\SetKwInOut{Input}{Input}\SetKwInOut{Output}{Output}
\Input{$\rX, \ \rS, \ \rpi, \ \rW_{1}, \ \rW_{2}$ same as Algorithm~\ref{alg:moe-swiglu-grouped-gemm}. 
}

\Output{MoE layer output $\rO$}

\underline{\textcolor{red}{Up-proj $A$ kernel} $(\textcolor{blue}{X, \ \W{1}, \ \pi} ) \rightarrow (\textcolor{blue}{H, \ A})$}:  

\note{// Gather + varlen-$\mathbf{M}$ Grouped GEMM + SwiGLU}\;

\Indp \textbf{Parallel} \For{$e \in [E]$}{

 $\bX_e, \ \bW_{1, e},\ \bpi_{:, e} \gets \mathrm{load}(\textcolor{blue}{\bX_e, \ \rW_{1, e}, \ \bpi_{:, e}})$\;

$\bX_e \gets \mathrm{Gather}(\rX, \ \bpi_{:, e})$\;


$\bH_e \gets \bX_e \bW_{1, e}$

\note{// apply activation function, e.g. $\mathrm{SwiGLU}$}

$\bA_{e} \gets \mathrm{act\_func}(\bH_e)$ \; 

$ \textcolor{blue}{\rH_e, \ \rA_{e}} \gets \mathrm{store}(\bH_e, \ \bA_{e})$

}

\Indm
\underline{\textcolor{red}{Down-proj $Y$ kernel} $(\textcolor{blue}{A, \ \W{2}}) \rightarrow \textcolor{blue}{Y}$}: \;

\note{// varlen-$\mathbf{M}$ Grouped GEMM}

\Indp
\textbf{Parallel} \For{$e \in [E]$}{

 $\bA_{e},\bW_{2, e} \gets \mathrm{load}(\textcolor{blue}{\bA_{e},\ \rW_{2, e}})$

$\bY_{e} \gets \bA_{e} \bW_{2, e}$

 $\textcolor{blue}{\rY_{e}} \gets \mathrm{store}(\bY_{e})$
}

\Indm
\underline{\textcolor{red}{Expert aggregation $O$ Kernel} $(\textcolor{blue}{Y, \ S, \ \pi}) \rightarrow \textcolor{blue}{O}$}: \;

\note{// Gather and sum}\;

\Indp
\textbf{Parallel} \For{$t \in [T]$}{
 $  \bY_{e,t},\ \bS_{t, e},\ \bpi_{t, e}\gets \mathrm{load}(\textcolor{blue}{  \bY_{e,t},\ \bS_{t, e},\ \bpi_{t, e}})$\;

$\bO_t \gets \sum_{e \in [E]} \bpi_{t, e} \bS_{t, e} \bY_{e,t}$\;

 $\textcolor{blue}{\rO_t} \gets \mathrm{store}(\bO_t)$
}


\end{algorithm}
\end{minipage}
\hfill\begin{minipage}{0.51\textwidth}
    \begin{algorithm}[H]
    \caption{\Algname's MoE kernel backward pass of down projection.}
    \DontPrintSemicolon
    \label{alg:our-down-bwd-moe}
     \fontsize{8pt}{10pt}\selectfont
    
    \SetKwInOut{Input}{Input}\SetKwInOut{Output}{Output}
    \Input{$\rS, \ \rpi, \ \rW_{2}, \ \rdO$.}
    
    \Output{$dH, \ d\W{2}, \ dS$.}

    \underline{\textcolor{red}{Down-proj act $dH$ kernel} $(\textcolor{blue}{dO, \ \W{2}, \ S, \ \pi}) \rightarrow (\textcolor{blue}{d H, \ dS, \ A'})$}: \;

    \note{// Gather + varlen-$\mathbf{M}$ Grouped GEMM + dSwiGLU + $dS$}\;

    \note{// Appendix~\ref{sec:appendix:proof} elaborates this algorithm in more detail} \;
    
    \Indp

    \textbf{Parallel} \For{$e \in [E]$}{
    $dO_{e}, \ \W{2,e}, \ S, \ \pi_{:,e}, \ H_e \gets \mathrm{load}(\textcolor{blue}{dO_{e}, \ \W{2,e}, \ S, \ \pi_{:,e}}, \ H_e)$ \;
    
    $dO_{e} \gets \mathrm{Gather}(dO, \ \pi_{:,e})$ \; 

    \note{// $dA'$ is a temp variable for computing $dA$, $dS$, and $A'$} \;

    $dA_{e}' \gets dO_{e} \ \W{2,e}^\top$ \tcp*[f]{\scriptsize $dA_{e}'\in\mathbb{R}^{T_e\times n}$}\;

    $\vs_e \gets \mathrm{Gather}(S, \ \pi_{:,e})$ \;

    $dA_{e} \gets \mathrm{Broadcast}(\vs_e) \ dA_{e}'$ \;

    \note{// compute fwd act and bwd act grad simultaneously in $\mathrm{dAct}$ call} \;


    $A_{e}, \ dH_{e} \gets \mathrm{dAct\_func}(dA_{e}, \ H_e)$ \; 


    $A_{e}' \gets \mathrm{Broadcast}(\vs_e)\ A_{e}$ \tcp*[f]{\scriptsize $A' \in \mathbb{R}^{T_e \times n}$, input for $d\W{2}$}\;


    $d S_{e,t} \gets \langle dA_{e,t}', \ A_{e,t}\rangle$  \tcp*[f]{\scriptsize reduce over $n$ dim}\;
            
    $\textcolor{blue}{d H_e, \ dS, \ A_{e}' }\gets \mathrm{store}(d H_e, \ dS, \ A_{e}')$
 
    }

    \Indm
    
    \underline{\textcolor{red}{Down-proj weight $d\W{2}$ kernel} $(\textcolor{blue}{dO, \ A', \ \pi}) \rightarrow \textcolor{blue}{d\W{2}}$}:  \;

    \note{// Gather + varlen-$\mathbf{K}$ Grouped GEMM}\;

    \Indp

    \textbf{Parallel} \For{$e \in [E]$}{
    $dO_{e},\ A_{e}',\ \pi_{:,e} \gets \mathrm{load}(\textcolor{blue}{dO_{e},\ A_{e}',\ \pi_{:,e}})$\;
    
    $dO_{e} \gets \mathrm{Gather}(dO, \ \pi_{:,e})$ \; 

   $d\W{2,e} \gets A_{e}'^{\top} \, dO_{e}$  \;

   $\textcolor{blue}{d\W{2,e}}\gets \mathrm{store}(d\W{2,e})$
    }

    

    
    

    
    

    \Indm

    \end{algorithm}

\end{minipage}
\end{figure}

\subsection{Overview of \Algname's MoE kernels}\label{sec:high-level}

The MoE computation\textsuperscript{\ref{footnote:moe_main_computation}} in \Algname~launches 8 kernels: during forward, we have up-proj ($A$), down-proj ($Y$), and expert aggregation ($O$) kernels; during backward, we have activation gradient kernels for $dH$ (down-proj), $d\tilde{X}$ (up-proj), $dX$ (aggregating $d\tilde{X}$ across experts), and weight gradient kernels $d\W{1}$ and $d\W{2}$. Figure~\ref{fig:workflow} illustrates the computational workflow of these 8 kernels. We provide an efficient TC top-$K$ router, and an interface that accepts arbitrary routing input. However, it should be noted that \Algname's MoE computation is independent of the MoE router choice and is thus compatible with arbitrary router logic.

The implementation of \Algname's MoE computation is highly modularized: it only consists of (1) an optimized grouped GEMM kernel with modularized fusion and (2) an optimized expert aggregation kernel. The host dispatches to the best GEMM config and load/store strategies to launch the 8 kernels listed above. Besides such high modularity, \Algname~still exhibits state-of-the-art training throughput and minimum activation memory usage which we describe below.

\subsection{Activation memory efficiency}\label{sec:mem-efficiency}
The FLOPs of MoE forward and backward computation is $(6+12)T n K d$. For a given $T,d$\footnote{Embedding dimension $d$ is often picked independently of MoE layer so we always assume $d$ is fixed.}, we need to keep $nK$ constant for constant FLOPs. Therefore, increasing granularity requires decreasing $n$ and proportionally increasing $K$. Hence, any activations with memory $O(TKd)$ should not be cached for backward computation to avoid activation memory scaling with granularity. For current MoE kernels like ScatterMoE, activations scale linearly with expert granularity.
Activations $Y$ (down-proj output) and $X_e$ (gathered $X$) have size $T K d$ and avoiding caching them eliminates activation memory dependency on granularity. We avoid writing $dY$ (gradient for $Y$) and $dO_e$ (gathered $dO$) to HBM as they increase the peak activation memory during the backward computation:


\begin{itemize}[nosep]
    \item For $X$ and $dO$, fusing the gather operation with the HBM load eliminates the need for materialization and activation caching in HBM. We show in Figures~\ref{fig:moe_breakdown} and~\ref{fig:gather-speed} that this gather fusion significantly improves the throughput for fine-grained MoEs.

    \item A naive implementation to compute $dS$ and $dH$ would need $Y$ and $dY$. Instead, we identify an alternative computation path to compute $dS$ and $dH$ without increasing FLOPs. This is achieved via expanding $dS$ and $dH$ into an equation that does not involve using $Y$ and $dY$, as illustrated in Appendix~\ref{sec:appendix:proof}. \Algname's $dH$ kernel is shown in Algorithm~\ref{alg:our-down-bwd-moe}.
\end{itemize}





As a result, we only cache $X$ and $H$ along with routing metadata for a total size $2 T d + 4 T K n$ bytes per layer. \textbf{This activation memory usage is the same as a dense model with the same number of activated parameters, which is the minimum activation memory required for backward computation without doing activation recomputation with GEMM.}\footnote{Although \Algname~still materializes a temporary $Y$ variable, we can recycle $Y$ after each layer. As long as the number of MoE layers (typically 32+ for 7B+ MoE) is larger than $K$,  the transient memory usage of $Y$ will be overshadowed. Removing such materialization requires an atomic add (Figure~\ref{fig:expert-agg} right) to global memory which creates new issues with determinism \citep{determinism}, numerical accuracy (for BF16 atomic add), and incompatibility with all2all or all-gather communication.} In Figure~\ref{fig:mem}, we profile \Algname's activation memory for a 7B MoE training configuration and demonstrate that the activation memory of \Algname~is independent of expert granularity.





\section{IO-aware kernel design} \label{sec:moe:hide-io} 

The expressivity of fine-grained MoE comes from the diversity of every token's expert selection, which in turn leads to linearly-scaled IO cost w.r.t. expert granularity. To sustain high throughput, we need to maximally (1) reduce IO access via fusion (2) overlap the IO latency with compute. We first examine the token gather fusion with computation, and math and IO fusion with epilogue in Section~\ref{sec:moe:prologue} and~\ref{sec:moe:epilogue} respectively. We then describe the techniques to overlap MMA with IO in Section~\ref{sec:moe:wgsp}. In Appendix~\ref{sec:appendix:kernel_comparison}, we compare \Algname~with other MoE kernel designs with a summary in Table~\ref{tab:kernel-comparison}.

\begin{table}[!ht]
    \centering

    \setlength{\tabcolsep}{2pt}
    \renewcommand{\arraystretch}{1.15}
    \fontsize{8pt}{10pt}\selectfont
    
    \begin{tabular}{@{\extracolsep{\fill}}l!{\vrule width 0.4pt}cccccc}
    \toprule
    \textbf{Features \textbackslash \ Methods} &
    \textbf{\Algname} &
    \textbf{ScatterMoE} &
    \textbf{MoMoE} &
    \textbf{MegaBlocks} &
    \textbf{Megatron} &
    \textbf{DeepGEMM} \\
    \midrule
    Gather fused with GMEM-to-SMEM load  (Sec.~\ref{sec:moe:prologue})             & \cmark & fwd \cmark, bwd \xmark  & fwd \cmark, bwd \xmark & \xmark & \xmark & \xmark \\

    SwiGLU and dSwiGLU fused with epilogue (Sec.~\ref{sec:moe:epilogue})  & \cmark & \xmark & \cmark & \xmark & \cmark & NA \\

    $dS$ computed as $\langle dA'_{e,t},\ A_{e,t}\rangle$ (Sec.~\ref{sec:moe:epilogue}, App.~\ref{sec:appendix:dS_computation}) & \cmark & \xmark & \xmark & \xmark & \cmark & NA \\

    Backward epilogue that computes $dH$, $dS$ together (Sec.~\ref{sec:moe:epilogue})    & \cmark & \xmark & \xmark & \xmark & \xmark & NA \\
    
    Overlap MMA with epilogue/IO (Sec.~\ref{sec:moe:wgsp})  & \cmark & \xmark & \xmark & \xmark & \xmark & \xmark \\
    


    Do \emph{not} need a separate scatter kernel    & \cmark & \cmark & \cmark & \xmark & NA & NA \\


    Efficient top-$K$ sorting (App.~\ref{sec:appendix:top-k}) & \cmark & \xmark & \xmark & \xmark & \xmark & NA \\


    Do \emph{not} need shape-alignment efforts outside GEMM kernels   & \cmark & \cmark & \cmark & \xmark & \xmark & \xmark \\

    \bottomrule
    \end{tabular}
    
    \caption{\small Comparison between \Algname~and prior MoE kernels. \cmark~means that the kernel implements the feature or a functionality similar in semantics, and \xmark~means the feature is missing from the kernel. ``NA'' means that the feature is out of the expected scope. We use the $\mathrm{GroupedMLP}$ for Megatron and $\mathrm{ParallelDroplessMLP}$ for MegaBlocks. More discussion is included in Appendix~\ref{sec:appendix:kernel_comparison}.} \label{tab:kernel-comparison}
\end{table}

\subsection{\Algname's Grouped GEMM}

\Algname~is built on top of an efficient varlen-$\mathbf{M}$ and varlen-$\mathbf{K}$ Grouped GEMM. Inside the Grouped GEMM, we fuse the gather operations with the activation loading (\ref{sec:moe:prologue}), and fuse SwiGLU/dSwiGLU/$dS$ with epilogue (\ref{sec:moe:epilogue}). The gather fusion helps \Algname~to be faster than MoE kernel designs that require a separate gather kernel such as MegaBlocks, Megatron, and DeepGEMM++, an optimized MoE forward pass implementation built on top of the DeepGEMM library \citep{deepgemm}. The epilogue fusion boosts \Algname~to be faster than ScatterMoE in the backward pass. These fusions reduce unnecessary IO access and can be overlapped with compute MMA, as we discuss in Section~\ref{sec:moe:wgsp}.




\subsubsection{Gather fusion with GMEM-to-SMEM load} \label{sec:moe:prologue}

\Algname's Grouped GEMM accepts either contiguously-packed inputs or inputs gathered from different positions, illustrated in Figure~\ref{fig:input-format}. For the latter case, we fuse the input gather with the input loads from global memory (GMEM, often the HBM) to shared memory (SMEM) so we can batch them to perform GEMM on Tensor Core \citep{tan2024scattered,costin2025momoe}. This involves (1) fetching the routed token indices for each expert and then (2) using these indices to gather activations via the $\mathrm{cp.async}$ instruction. 



\begin{figure}[h]
    \centering
    \begin{subfigure}{\linewidth}
        \includegraphics[width=\linewidth]{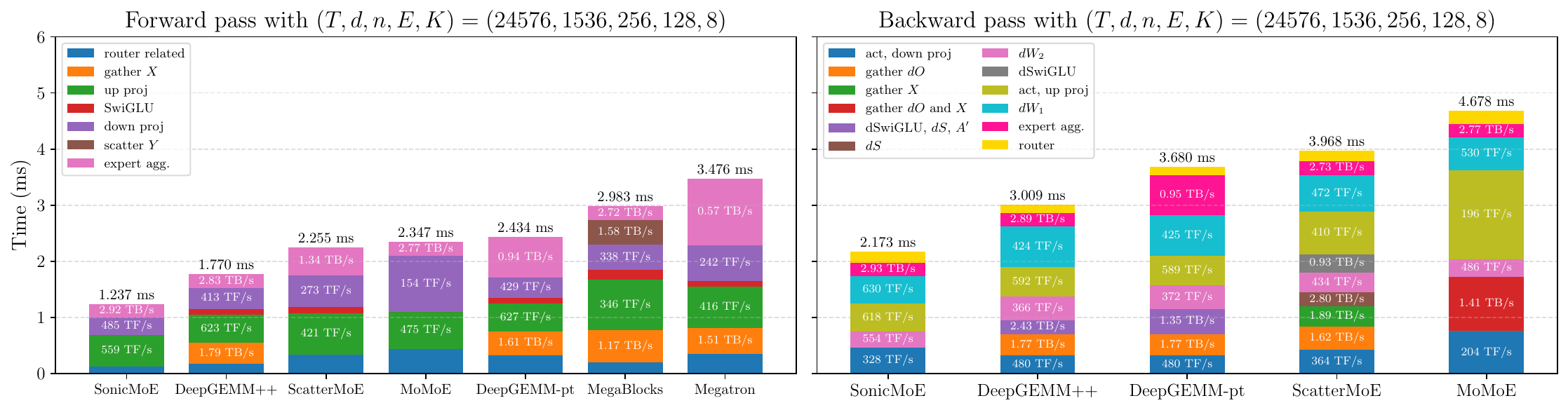}
        \caption{\small Runtime breakdown of 7B MoE training on H100 GPUs. } 
        \label{fig:moe_breakdown}
    \end{subfigure}
    \begin{subfigure}{\linewidth}
        \includegraphics[width=\linewidth]{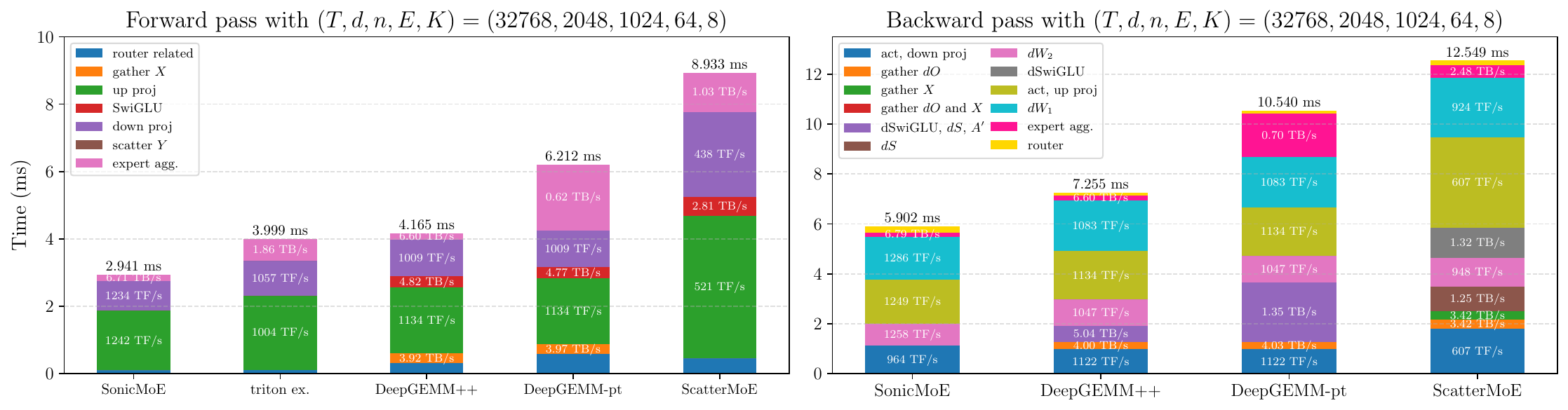}
        \caption{\small Runtime breakdown of OLMoE-sized 7B \citep{muennighoff2024olmoe} MoE training on B300 GPUs. Triton official example does not implement the backward pass.}
        \label{fig:moe_breakdown-B300}
        \vspace{-.5em}
    \end{subfigure}
    \caption{\small Runtime breakdown of different MoE kernels (ms $\downarrow$) on the 7B MoE training. We annotate the model memory bandwidth (TB/s $\uparrow$) for memory-bound kernels (gather, SwiGLU/dSwiGLU, and expert aggregation kernel) and compute throughput (TFLOPS $\uparrow$, abbr as TF/s) for grouped GEMM kernels. Note that this profile is grouped by kernel runtime semantics and one block can contain multiple actual kernel timing results. For example, the ``router related'' on left subfigure includes both router GEMM and routing metadata computation time. In addition, we do not consider the CUDA stream bubble time across kernels in this figure. We use the $\mathrm{GroupedMLP}$ for Megatron, and $\mathrm{ParallelDroplessMLP}$ for MegaBlocks. DeepGEMM does \textit{not} provide an efficient router implementation, gather and expert aggregation kernels during the forward pass, where we use a standard PyTorch implementation (``DeepGEMM-pt'') or our highly optimized kernels (``DeepGEMM++'') for them. During the backward pass, both ``DeepGEMM++'' and ``DeepGEMM-pt'' use the same computational path as \Algname, except we launch separate kernel(s) that compute $dS$, $A'$, and $\mathrm{dSwiGLU}$ together. DeepGEMM++ is effectively the best possible MoE implementation built on top of DeepGEMM SM90/SM100 BF16 Grouped GEMM kernels without modifying DeepGEMM's source code. On B300 GPUs, we also compare with triton official example (``triton ex.'') for MoE that is optimized for inference as it does not store pre-activation results $H$. }
    \label{fig:moe_breakdown-total}
\end{figure}

As shown in Figure~\ref{fig:moe_breakdown} and~\ref{fig:moe_breakdown-B300}, Gather fusion provides \Algname~with a major advantage over existing MoE kernel designs on both H100 and B300 GPUs such as DeepGEMM. Although DeepGEMM's varlen-$\mathbf{M}$ Grouped GEMM kernel is highly optimized, DeepGEMM assumes the inputs are already contiguously packed and padded to multiples of 128, which requires a separate kernel launch for gather and pad before the Grouped GEMM. 

In the backward pass, weight gradients for up-proj and down-proj ($d\W{1}$ and $d\W{2}$ respectively) need to gather $X$ and $dO$, and the activation gradient for $dH$ also needs to gather $dO$. Despite the backward having more kernels requiring the gather operation, existing approaches including ScatterMoE \citep{tan2024scattered} and MoMoE \citep{costin2025momoe} fuse the gather during forward but still launch a separate gather kernel during backward. Fusing this gather reduces the IO cost by $2TKd$ bytes and cuts down a major portion of fine-grained MoE training time.

\begin{wrapfigure}{r}{0.44\textwidth}
\begin{minipage}{0.44\textwidth}
    \centering
    \vspace{-2.5em}
    \includegraphics[width=\linewidth]{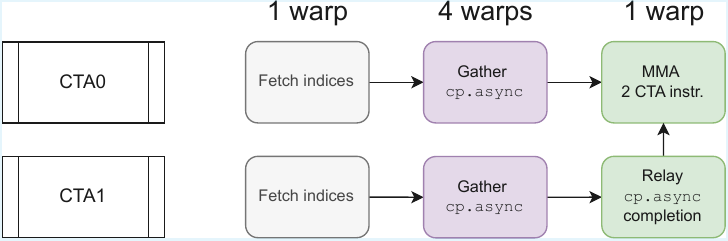}
    \caption{\small Pipeline structure for gather fusion with $\mathrm{cp.async}$ on Blackwell GPUs using 2-CTA clusters.
    }
    \label{fig:blackwell-gather-pipeline}
\end{minipage}
\end{wrapfigure}

\paragraph{Blackwell GPUs.} On Blackwell GPUs, the gather fusion with $\mathrm{cp.async}$ encounters an architectural challenge when using 2-CTA clusters (Figure~\ref{fig:blackwell-gather-pipeline}) for GEMM computation. The $\mathrm{cp.async}$ instruction, introduced in the Ampere generation, can only signal completion within the same CTA. However, Blackwell's 2-CTA GEMM requires the MMA instruction in the leader CTA (CTA 0) to wait for gather completion from both CTAs. To work around this limitation, CTA 1 requires a dedicated relay warp that receives the $\mathrm{cp.async}$ completion signal and forwards it to CTA 0's MMA warp using cluster-level synchronization primitives (e.g., mbarrier with cluster scope). This relay mechanism adds scheduling complexity but enables efficient gather fusion across the 2-CTA cluster, maintaining high throughput for Grouped GEMM.





\subsubsection{Epilogue fusion}\label{sec:moe:epilogue}

We exploit the epilogue computation to maximally reduce unnecessary IO accesses with the following design choices:

\begin{itemize}
    \item \textbf{SwiGLU and dSwiGLU fusion}: We fuse the SwiGLU and backward of SwiGLU with the epilogue of forward up-proj and backward down-proj activation gradient kernel respectively \citep{costin2025momoe}. 
    
    
    \item \textbf{Computing $dH$ and $dS$ in backward down-proj activation gradient ($dH$) kernel's epilogue}: This heavy epilogue fusion helps \Algname's $dH$ kernel to produce the same output with far less total time than ScatterMoE's down-proj act, $dS$, and dSwiGLU combined together in Figure~\ref{fig:moe_breakdown} and~\ref{fig:moe_breakdown-B300}. 
    

    In Appendix~\ref{sec:appendix:dS_computation}, we show that \Algname's $dS = \langle dA, \ A'\rangle = \langle dA, \mathrm{Broadcast}(\vs) A\rangle$ is the computationally and activation memory-efficient choice for fine-grained MoEs. However, both ScatterMoE and MoMoE choose to compute $dS$ as $\langle dO, \ Y \rangle$, which requires an additional $2TKd$ HBM load cost and requires caching $2TKd$ bytes of activation memory. In Figure~\ref{fig:moe_breakdown} (right subfigure), ScatterMoE launches a separate kernel for $dS$ while MoMoE fuses $dS$ with up-proj activation gradient which takes much longer time than \Algname's up-proj activation gradient.


\end{itemize}

The throughput of heavy epilogue fusion on backward down-proj activation gradient $dH$ kernel is boosted by the overlap of asynchronous IO and MMA, which we will elaborate in Section~\ref{sec:moe:wgsp}. Such overlap helps \Algname~to sustain a reasonable training throughput and memory bandwidth simultaneously even with the heavy epilogue fusion (load $H$ and $S$, compute $dH$, $dS$, and $A'$ as inputs to $d\W{2}$) in $dH$ kernel.

\subsection{GEMM MMA Overlapping with Asynchronous IO} \label{sec:moe:wgsp}

\begin{figure}[h]
    \centering
    \begin{subfigure}{\linewidth}
    \centering
        \includegraphics[width=0.85\linewidth]{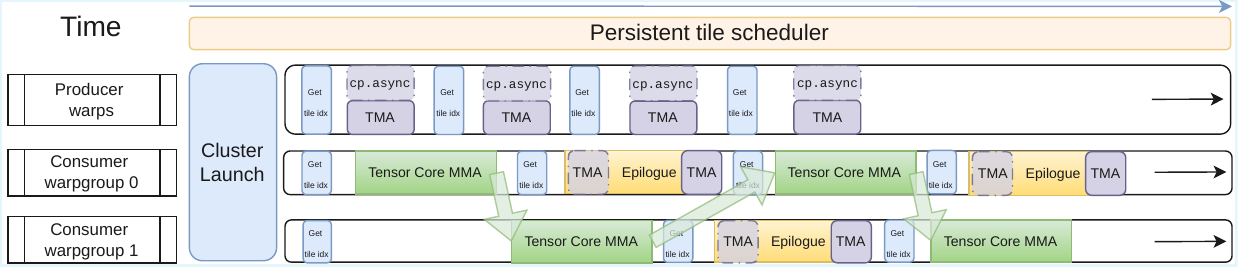}
        \caption{\small \Algname's Ping-Pong warpgroup scheduling on Hopper GPUs. The green arrows indicate that a consumer warpgroup signals the start of the epilogue and the other consumer warpgroup can proceed with the MMA. Once this step is complete, the roles of 2 consumer warpgroups are switched. \Algname~mainly uses Ping-Pong for forward down-proj $Y$ kernel and backward down-proj activation gradient $dH$ kernel as they both have heavy epilogue. In $dH$ kernel, \Algname~has an asynchronous TMA load during epilogue. This figure is adapted from \citet{torch_cutlass_pingpong}'s blog on Ping-Pong scheduling.}
        \label{fig:pingpong-hopper}
    \end{subfigure}
    \begin{subfigure}{\linewidth}
    \centering
        \includegraphics[width=0.85\linewidth]{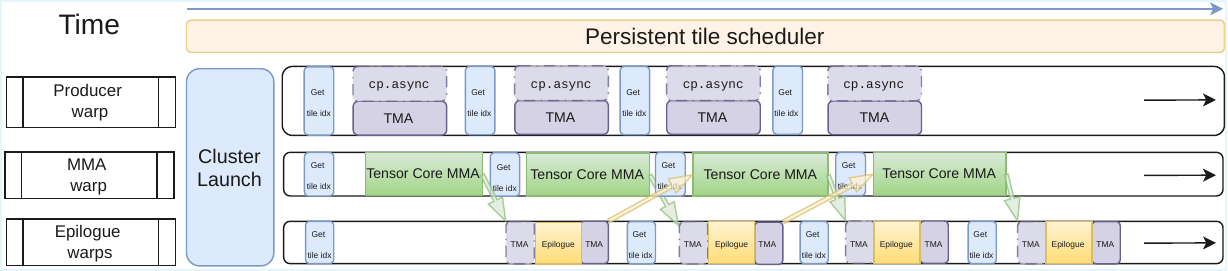}
        \caption{\small \Algname's strategy to overlap GEMM MMA with epilogue IO on Blackwell GPUs. The green arrows indicate that the MMA warp signals the epilogue warps that the MMA accumulation result of a work tile is ready for epilogue. The yellow arrows indicate that the epilogue warps finish the epilogue of a work tile and signal the MMA warp that the epilogue warps' owned Tensor Memory (TMEM, an on-chip memory on Blackwell GPUs) stages now become available. Figure~\ref{fig:tmem-blackwell} illustrates this TMEM stage ownership transfer process.}
        \label{fig:pingpong-blackwell}
        \vspace{-.5em}
    \end{subfigure}
    \caption{\small \Algname's strategy to overlap GEMM MMA with asynchronous IO on Hopper and Blackwell GPUs.}
\end{figure}

\paragraph{Hopper GPUs.}

In NVIDIA Hopper GPUs, GEMM is performed asynchronously with a producer-consumer paradigm \citep{shah2024flashattention3fastaccurateattention}. Suppose we have 2 consumer warpgroups, we can overlap the IO of 1 warpgroup with the GEMM of another warpgroup. Once this is finished, we switch the roles of the warpgroups (effectively interleaving IO and GEMM). This is often referred to as \textit{Ping-Pong scheduling} \citep{wright2024cutlass,shah2024flashattention3fastaccurateattention} on Hopper GPUs in Figure~\ref{fig:pingpong-hopper}.



Ping-Pong scheduling is particularly useful to maintain high Tensor Core throughput with heavy epilogue. For example, the down-proj forward $Y$ kernel's epilogue has heavy HBM store IO ($2TKd$ bytes) relative to the mainloop. In the down-proj activation gradient ($dH$) kernel's epilogue, we need to load $H$ and execute multiple activation and reduction operations to compute and store $dH$, $dS$, and $A'$ as inputs for $d\W{2}$. We note that the concept of overlapping MMA with IO and Ping-Pong scheduling is known in other places such as Flash Attention 3 \citep{shah2024flashattention3fastaccurateattention}, but the application of Ping-Pong scheduling to address the increasing IO costs of fine-grained MoE kernel design is novel.



Besides Ping-Pong scheduling, \Algname~also relies on asynchronous TMA operations to perform GMEM-to-SMEM load and SMEM-to-GMEM store. We overlap the following asynchronous IO with the MMA operations:

\begin{itemize}[noitemsep]
    \item \textbf{Asynchronous TMA load during $dH$ kernel's epilogue}: In the $dH$ kernel's epilogue, we need to load $H$ to compute $dH$ from $dA$. We create a dedicated pipeline for asynchronous TMA load of $H$ to overlap with other epilogue operations across epilogue stages. 
    

    \item \textbf{Asynchronous TMA store in forward down-proj $Y$ and backward up-proj activation gradient $d\tilde{X}$ kernel}: in forward down-proj and backward up-proj activation gradient, \Algname~does \emph{not} fuse the scatter with HBM store where ScatterMoE and MoMoE both choose to fuse the HBM store with scatter. This is primarily because the scatter fusion requires a synchronous SMEM-to-GMEM store instruction on Hopper GPUs.\footnote{\textbf{Synchronous $\mathrm{st.global}$ is the \emph{only} available PTX instruction for scatter fusion with HBM store on Hopper GPUs if we do not use TMA 1D store.} This is different from the gather case as $\mathrm{cp.async}$ is asynchronous but it cannot be used for SMEM-to-GMEM store. Although asynchronous $\mathrm{st.async.release.global}$ instruction is available on Blackwell GPUs, the repeated index fetching would still make scatter a less favorable option.} The synchronous GMEM store blocks the execution of MMA of next tile and largely degrades the TFLOPS ($\sim$20\%) in the case of heavy HBM store, as illustrated by Figure~\ref{fig:expert-agg-why-slow}. 
\end{itemize}


\paragraph{Blackwell GPUs.}

\begin{wrapfigure}{r}{0.45\textwidth}
\begin{minipage}{0.45\textwidth}
    \centering
    \vspace{-3.5em}
    \includegraphics[width=\linewidth]{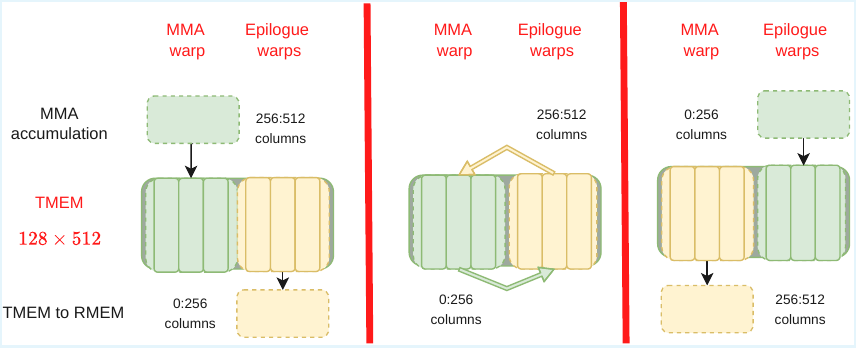}
    \caption{\small Illustration that shows MMA warp would coordinate with epilogue warps to fully utilize TMEM resources while overlapping MMA with asynchronous IO.}
    \label{fig:tmem-blackwell}
\end{minipage}
\vspace{-1.2em}
\end{wrapfigure}

In NVIDIA Blackwell GPUs, GEMM kernels use the same ``Ping-Pong'' scheduling in spirit, but the implementation differs from Hopper. Blackwell introduces Tensor Memory (TMEM), a dedicated 256KB on-chip memory per SM organized as 128 rows $\times$ 512 columns of 32-bit cells \citep{nvidia2022_gb200,colfax_cutlass_tmem}. The accumulator results from matrix multiplication are stored directly in TMEM rather than in registers, with the 512-column structure naturally enabling a two-stage accumulator pipeline. Each stage uses 256 columns: while one stage performs MMA operations via the new UMMA (Unified Matrix Multiply-Accumulate) instruction, the other stage executes the epilogue. This process is illustrated in Figure~\ref{fig:tmem-blackwell}. Unlike Hopper's WGMMA which required warpgroup-level coordination and consumed significant register memory, Blackwell's UMMA is a single-threaded asynchronous operation that eliminates register pressure for accumulation. This architectural change allows epilogue warps to read and process results from one TMEM stage concurrently with MMA warps accumulating into the other stage, enabling better overlap of epilogue and MMA operations compared to Hopper's ping-pong scheduling.

\section{Token rounding routing} \label{sec:token_rounding}

In this section, we analyze the hardware efficiency under sparse MoE training regime and identify that as MoEs become sparser, the wasted compute on padded GEMM tiles accumulates to a nontrivial amount, known as ``tile quantization" effects. In response, we propose a novel routing method ``token rounding'' to eliminate tile quantization effects.

\subsection{Training efficiency of sparse MoE}\label{sec:sparse_moe}

\begin{figure}[!ht]
  \centering
  \begin{minipage}[t]{0.47\textwidth}
    \vspace{0pt} 
    \input{algorithm/token_rounding} 
  \end{minipage}
  \hfill
  \begin{minipage}[t]{0.52\textwidth}
    \vspace{0pt} 
    \centering
    
    \includegraphics[width=0.37\linewidth]{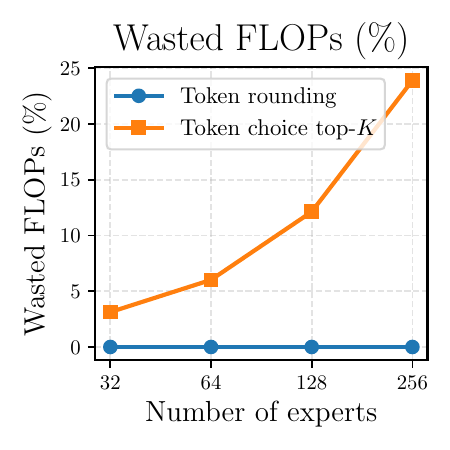}
    \vspace{-1em}     
    \captionof{figure}{\small Wasted FLOPs by padding during MoE forward \& backward pass with $T=16k$, $d=4k$, $n=1k$, $K=4$ as illustrated in the bottom right 2 subfigures of Figure~\ref{fig:token_rounding_speed}. \label{fig:wasted_flops}}

    \includegraphics[width=0.7\linewidth]{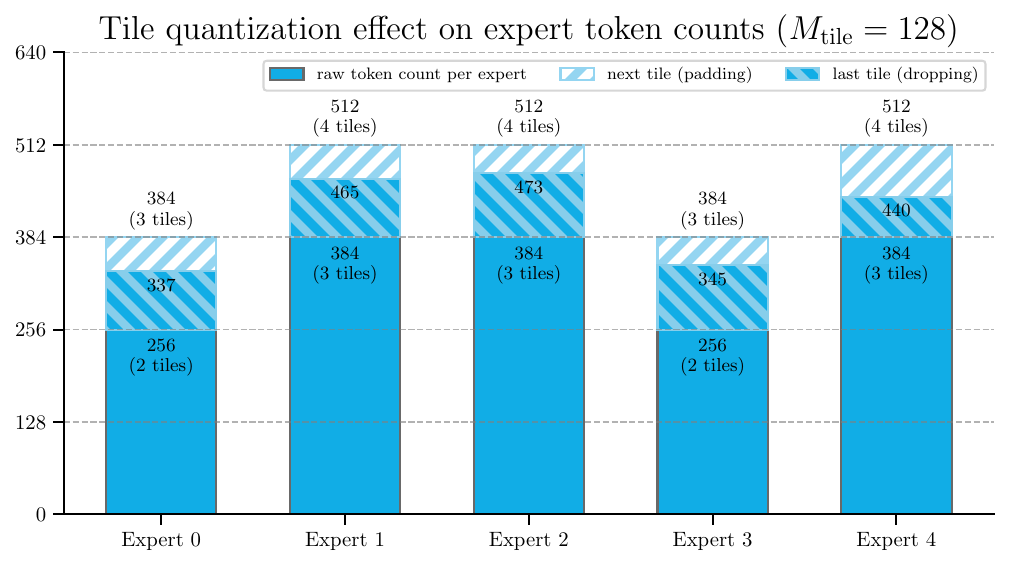}
    \captionof{figure}{\small A demonstration of tile quantization effect for sparse MoE. The rounding subroutine in TR makes a binary decision for discarding or padding tokens to guarantee that each expert receives an $\tileM$-multiple number of tokens.}  \label{fig:tile-quant}


  \end{minipage}
\end{figure}

Besides granularity, the arithmetic intensity of MoE also depends on the MoE activation ratio $\rho$ as shown in Equation~\ref{eqn:arithmetic_intensity}. When we scale down $\rho$, the expected number of received tokens per expert $\mathbb{E}_{e\in[E]} T_e = \bar{T}_e =  T \rho$ will also decrease linearly and the GEMM computation shifts towards a memory-bound regime.

\paragraph{Tile quantization effect.} GEMM on modern GPUs is often computed in tiles \citep{nvidia2022_h100} and we always need to pad to the next tile-sized multiples if any dimensions of $\mathbf{M},\mathbf{N},\mathbf{K}$ are not fully divisible by tile sizes. Once the size of input (e.g. token dimension per expert) is small, the wasted TFLOPS by padding can be nontrivial.




Therefore, we propose to use token rounding to avoid launching such extra tiles, thereby leading to more efficient training. We also empirically show that our token rounding method does not affect model quality while achieving much higher training throughput.

\subsection{Token rounding routing}\label{sec:tr}

As such, we propose to use the token rounding (TR) method as a 2-step sorting algorithm as shown in Algorithm~\ref{alg:token_rounding}. The token rounding algorithm first computes the vanilla token-choice (TC) routing results and applies a sorting of the router score over each expert's tokens, similar to the EC sorting step. We then choose to either discard tokens selected in the first step of TC top-$K$ routing or pad additional tokens in the second step of sorting. Between these 2 steps, we process the routing weight matrix such that the TC tokens are always preferred over EC tokens. This is done so that both discarding or padding only affects the last input tile for each expert.


Token rounding requires a $\mathrm{round\_and\_sparsify}$ subroutine for making a binary decision about discarding or padding. Our default choice for such a subroutine is to round expert frequency to the nearest $\tileM$ multiples: we choose to pad EC selected tokens if $\roundup{f_e} - f_e$ is smaller than $f_e - \rounddown{f_e}$. \footnote{For simplicity, we always use $\tileM$ as 128 in Table~\ref{tab:exp-sparse} and Figure~\ref{fig:token_rounding_speed}.} We further conduct an ablation in Table~\ref{tab:exp-ablation} and find that (1) our TR algorithm is quite robust w.r.t. the underlying rounding subroutine (2) this simple strategy of nearest rounding on expert frequency is often sufficient to yield excellent task performance. More detailed discussion on different rounding subroutines is included in Appendix~\ref{sec:appendix:ablation}.

\paragraph{MoE training \& inference quality.}

This simple algorithm guarantees that for each expert, the maximum deviation from token-choice routing is at most 1 tile. We find that this property has a surprisingly robust performance even under sparse MoE training regime and can serve as a substitute for token-choice under sparse MoE training settings, which is shown in Table~\ref{tab:exp-sparse}. We also conduct an ablation study on the effect of microbatch size $T$ and tile size $\tileM$ on the quality of trained MoE model with TR in Table~\ref{tab:exp-T-ablation} and~\ref{tab:exp-tileM-ablation}, and we find token rounding routing is generally robust when $\bar{T}_e/\tileM \geq 2$.



\paragraph{MoE training throughput.}

TR guarantees no tile quantization effects and in Section~\ref{sec:exp:token_round_throughput}, we show that TR training throughput over vanilla TC top-$K$ is consistently higher when in the highly sparse MoE training regime and can achieve 16\% higher TFLOPS for the kernel runtime as we scale up $E$ while keeping $K$ constant. 


\section{Experiments}





\begin{figure}[!ht]
    \centering
    \includegraphics[width=\linewidth]{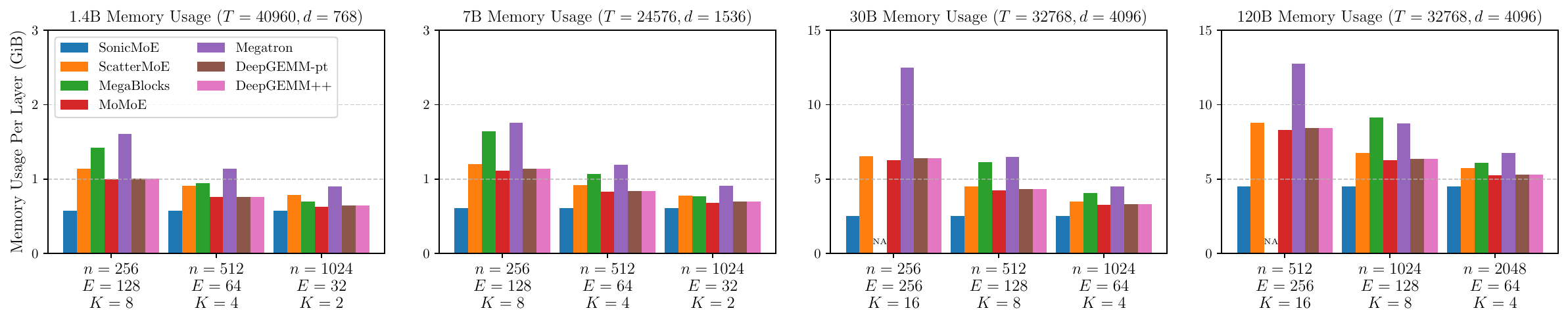}
    \caption{\small Peak activation memory usage per layer across different model scales (1.4B–120B) on H100 GPUs. MegaBlocks does not support small $n$. The benchmark configurations are listed in Table~\ref{tab:config_benchmark}. For ``DeepGEMM++'' and ``DeepGEMM-pt'', we only cache $X$, gathered $X_e$, $H_e$ for each expert $e$ and routing metadata which is the minimum amount of activation memory required for backward computation without GEMM recomputation.}
    \label{fig:mem}
\end{figure}



We evaluate \Algname's activation memory footprint (Section~\ref{sec:exp:mem}) and training throughput (Section~\ref{sec:exp:speed}) compared to other baseline MoE implementations. We also demonstrate the efficacy of the token rounding routing strategy and show that it is possible to use token choice as a drop-in replacement after training with token rounding in Section~\ref{sec:exp:token_round_perf}. We also show that token rounding can maintain the training throughput under sparse MoE configuration in Section~\ref{sec:exp:token_round_throughput}.

\subsection{\Algname's activation memory} \label{sec:exp:mem}
We demonstrate that \Algname\ achieves the lowest peak activation memory footprint for a single MoE layer as shown in Figure~\ref{fig:mem} across all scales on H100 GPUs. For the 7B model with $n=256$, our approach reduces memory usage by \textbf{45\%} compared to ScatterMoE, and more significantly compared to MoMoE. For 30B and 120B models, the gap becomes even wider: at 120B scale, our method saves more than \textbf{3GiB memory per layer} compared to MoMoE. We also validate that \Algname's activation memory stays constant w.r.t. expert granularity as shown in Figure~\ref{fig:teasor}.


\subsection{\Algname's training throughput}  \label{sec:exp:speed}

\subsubsection{Entire forward and backward throughput}
\begin{figure}[!ht]
    \centering
    \begin{subfigure}{\linewidth}
        \includegraphics[width=\linewidth]{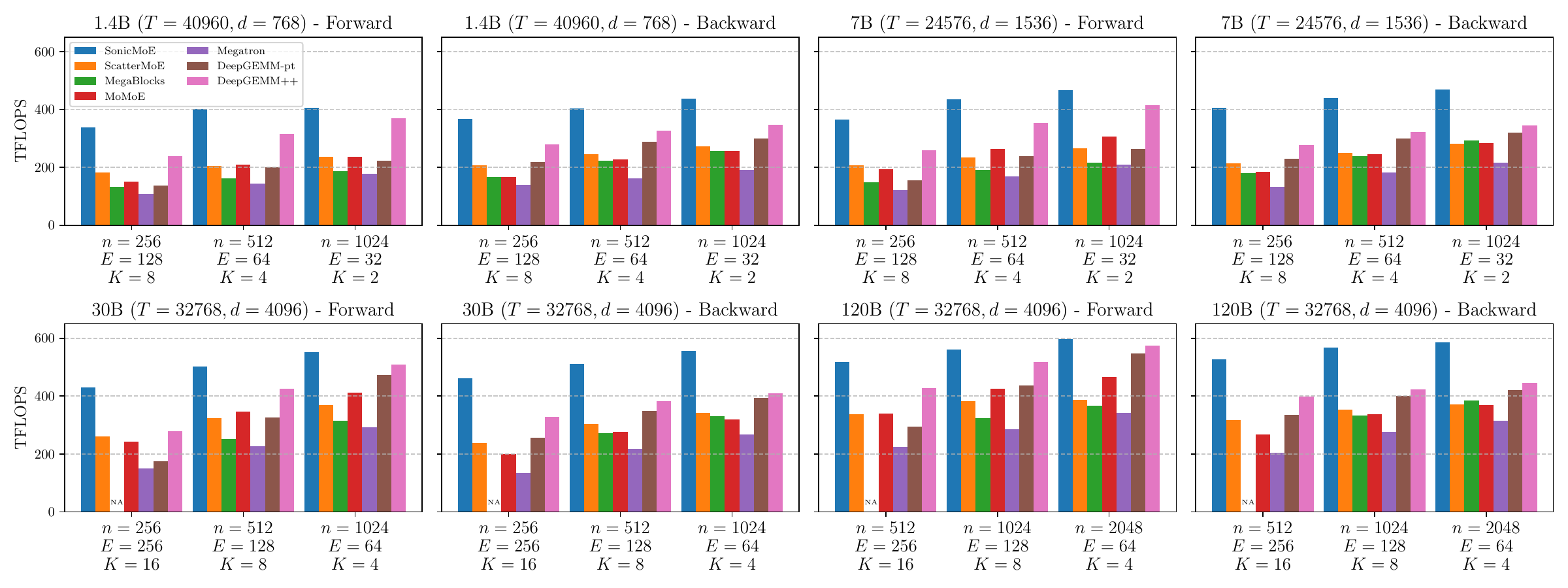}
        \caption{\small Forward \& backward TFLOPS for different MoE kernels on H100 GPUs.} 
        \label{fig:speed}
    \end{subfigure}
    \begin{subfigure}{\linewidth}
        \includegraphics[width=\linewidth]{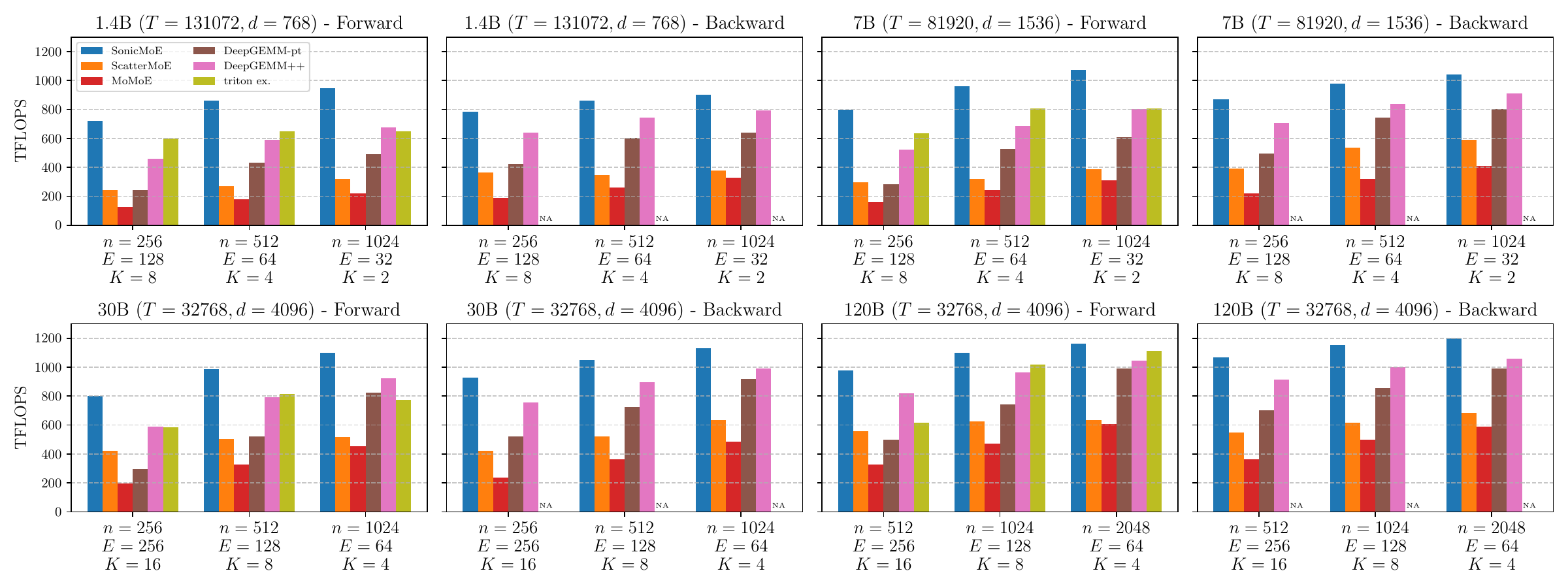}
        \caption{\small Forward \& backward TFLOPS for different MoE kernels on B300 GPUs.}
        \label{fig:B300-speed}
        \vspace{-1em}
    \end{subfigure}
    \caption{\small Forward \& backward TFLOPS for different MoE kernels on H100 and B300 GPUs. The definition of ``DeepGEMM++'', ``DeepGEMM-pt'', and ``triton ex.'' are the same as in Figure~\ref{fig:moe_breakdown-total}.}
\end{figure}

Figure~\ref{fig:speed} reports the compute throughput of forward and backward pass of one MoE layer in various MoE training configurations on H100 GPUs. Across all model scales, our method consistently achieves the highest TFLOPS. \textbf{For example, on a fine-grained 7B MoE model with $n=256$, \Algname~increases the \tflops~by 43\% on the forward pass compared to a highly optimized DeepGEMM baseline, and by 83\% and 115\% on the backward pass compared to ScatterMoE and MoMoE, respectively.} \Algname~also demonstrates speedup over DeepGEMM++ in the forward pass, which mainly arises from the gather $X$ kernel and Ping-Pong scheduling. The effect of both features increases as the MoE becomes more fine-grained and thus \Algname's relative speedup over DeepGEMM++ becomes larger.

\textbf{We further measure the real training throughput of a 7B MoE model with $n=256$ with FSDP-2: \algname\ on 64 H100s achieves 213 billion tokens per day, which achieves similar throughput to ScatterMoE on 96 H100s with 225 billion tokens per day.} The throughput for this is measured using the lm-engine codebase\footnote{\scriptsize\url{https://github.com/open-lm-engine/lm-engine}} \citep{Mishra_lm_engine_A_2024}. We shard the model using ZeRO-3 within a single node (8x H100s) and replicate this sharded unit across nodes for this experiment.

Besides H100 GPUs, we also measure \Algname's compute throughput on B300 GPUs in Figure~\ref{fig:B300-speed}. We mainly compare with DeepGEMM++ which is powered by DeepGEMM SM100 BF16 grouped GEMM kernels. \Algname~still demonstrates an overall speedup over DeepGEMM++, and such speedup is more pronounced when we increase expert granularity similar to the trend on H100 GPUs. For example, when we increase the expert granularity $d/n$ from 2 to 8 for 120B MoE training, \Algname~achieves a greater relative speedup over DeepGEMM++ from 11.6\% and 13.4\% in forward and backward to 19.6\% and 16.8\% respectively. \textbf{We also highlight that \Algname's forward pass has higher throughput than the triton official example, despite \Algname~storing both pre- and post-activation to GMEM whereas triton official example is designed for inference and only stores post-activation $A$ in the up-projection kernel. }

\begin{figure}[!ht]
    \centering
    \begin{subfigure}{\linewidth}
    \includegraphics[width=\linewidth]{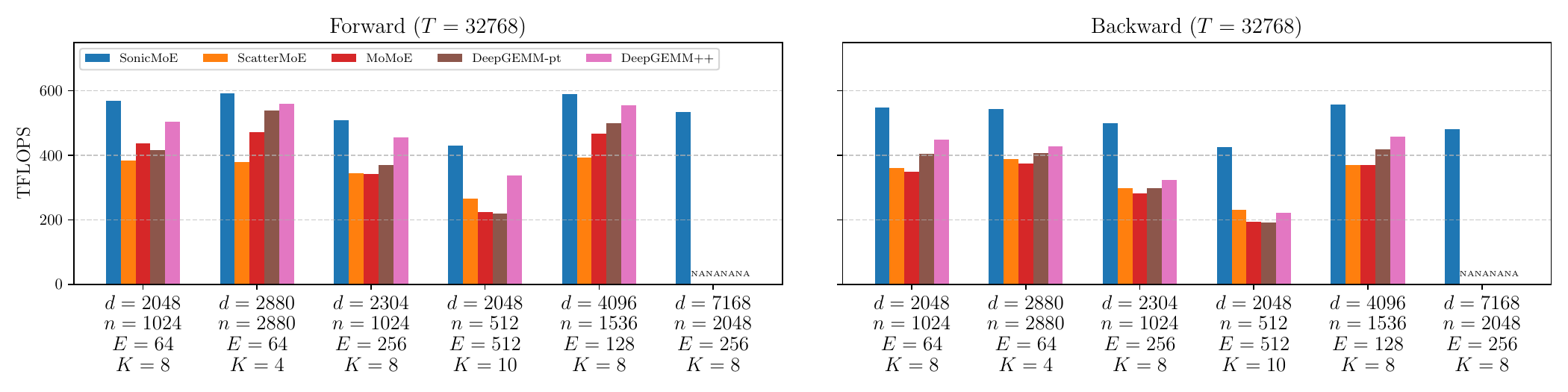}
    \caption{\small Forward \& backward TFLOPS of a single MoE layer on H100 GPUs. ScatterMoE, MoMoE, DeepGEMM-pt, and DeepGEMM++ all fail to run (either due to index overflow or CUDA OOM errors) for the DeepSeek-V3.2-Exp configuration.} \label{fig:real-moe-speed}
    \end{subfigure}
    \begin{subfigure}{\linewidth}
    \includegraphics[width=\linewidth]{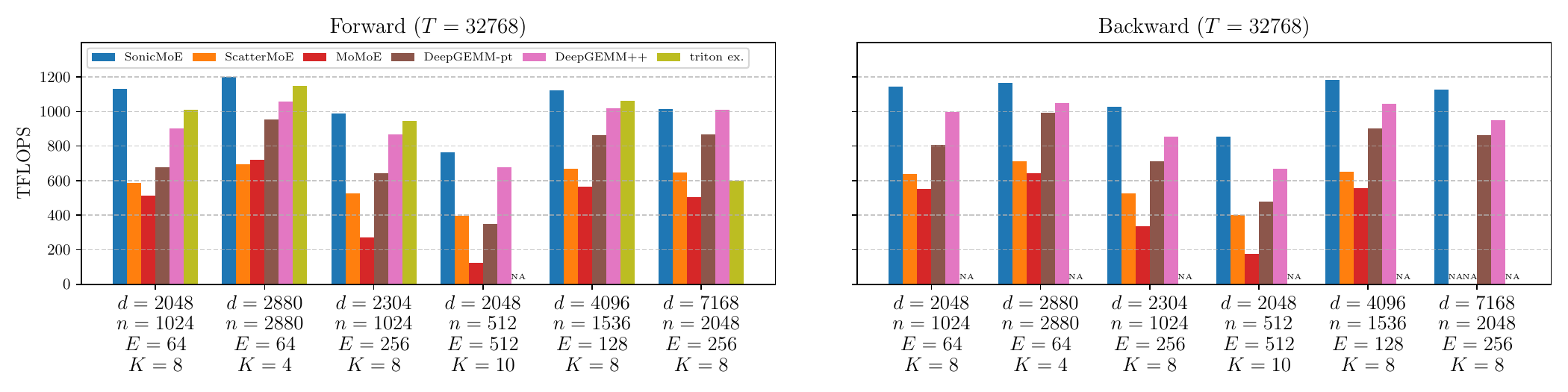}
    \caption{\small Forward \& backward TFLOPS of a single MoE layer on B300 GPUs. Triton official example for MoE does not support $K=10$ for the \textit{Qwen3-Next-80B-A3B-Thinking} configuration. } \label{fig:real-moe-speed-B300}
    \end{subfigure}
    \caption{\small Forward \& backward TFLOPS of a single MoE layer for different MoE kernels for different configurations ranging from 7B to 685B parameters on H100 and B300 GPUs. The MoE configurations from left to right adopt the model size of \model{OLMoE-1B-7B-0125}~\citep{muennighoff2024olmoe}, \model{gpt-oss-20b}~\citep{openai2025gptoss120bgptoss20bmodel}, \model{Kimi-Linear-48B-A3B-Base}~\citep{team2025kimi}, \model{Qwen3-Next-80B-A3B-Thinking}~\citep{qwen3technicalreport}, \model{Qwen3-235B-A22B-Thinking-2507}~\citep{qwen3technicalreport}, and \model{DeepSeek-V3.2-Exp}~\citep{deepseekai2024deepseekv32}. For fair comparison, we do not consider shared experts and expert biases, and we always use TC top-$K$ router with softmax scores. } 
\end{figure}

In addition, we measure the training throughput of a single MoE layer with configurations from recent open-source MoEs in Figure~\ref{fig:real-moe-speed} for H100 GPUs and in Figure~\ref{fig:real-moe-speed-B300} for B300 GPUs. On H100 GPUs, \Algname~generally achieves more than 550 \tflops~during both forward and backward pass, and consistently surpasses all baselines. We note that ScatterMoE, MoMoE, DeepGEMM-pt, and DeepGEMM++ all fail to run at the configuration for DeepSeek-V3.2-Exp, a 685B MoE model, while \Algname~successfully runs on a single H100 GPU. We also note that \Algname's IO-aware kernel design can achieve a greater relative speedup (e.g. Qwen3-Next-80B-A3B-Thinking) for sparse and fine-grained MoEs. 


On B300 GPUs, \Algname~generally achieves more than 1100 \tflops~during both forward and backward pass. \textbf{For the OLMoE-sized 7B MoE models, \Algname~demonstrates 25\% and 15\% higher \tflops~during forward and backward pass respectively than DeepGEMM++.} \Algname~also reaches 11.8\% higher \tflops~for forward pass than the triton official example. 


\subsection{Token rounding} 

\subsubsection{Token rounding's general task evaluation} \label{sec:exp:token_round_perf}

\begin{table}[!ht]
    \setlength{\tabcolsep}{2pt}
    \renewcommand{\arraystretch}{1.1}
    \fontsize{7pt}{8pt}\selectfont
    \centering
    \vspace{-0.6em}
    \caption{\small Comparison of different routing methods' task evaluation. ``Train'' and ``Val'' refer to the perplexity towards the end of training and on the validation set respectively. The next 11 columns are downstream tasks evaluated at the end of training and we report the accuracy for each. ``Avg'' is the mean accuracy across these 11 downstream tasks. We use TC top-$K$ routing for TR, token dropping, and EC baselines when evaluating validation perplexity and task performance. $\bar{T}_e$ represents the average number of received tokens in each microbatch per expert.}
    \label{tab:exp-sparse}

    \begin{subtable}[!ht]{\textwidth}
   \caption{\textbf{0.5B params, 20B tokens, 8/64 activated} \;($\bar{T}_e = 4096, \ \tileM = 128$)} \label{tab:sparsity:a}
    \vspace{-0.5em}
    \begin{tabular*}{\textwidth}{@{\extracolsep{\fill}}l!{\vrule width 0.4pt}cc!{\vrule width 0.4pt}ccccccccccc!{\vrule width 0.4pt}c}
    \toprule
    \textbf{Method} & \textbf{Train} & \textbf{Val} & \textbf{Wino} & \textbf{SIQA} & \textbf{SciQ} & \textbf{PIQA} & \textbf{OBQA} & \textbf{HS} & \textbf{COPA} & \textbf{CSQA} & \textbf{BoolQ} & \textbf{ArcE} & \textbf{ArcC} & \textbf{Avg} \\
    \midrule
    \textbf{TR} & \textbf{15.91} &  \textbf{15.94} & 51.9 & 41.3 & \textbf{80.8} & \textbf{65.5} & \textbf{35.0} & \textbf{38.7} & 63.0 & 31.2 & \textbf{61.4} & \textbf{58.9} & 27.1 & \textbf{50.4} \\\grayline

    TC top-$K$ & 16.04 &  16.01 & 51.0 & 41.4 & 79.2 & \textbf{65.5} & 31.6 & 38.4 & 66.0 & 31.5 & 60.2 & 57.5 & 25.7 & 49.8 \\

    TC (token drop) & 16.52 & 16.46 & 51.1 & 41.1 & 79.5 & 64.6 & 30.2 & 37.3 & 63.0 & \textbf{31.8} & 58.2 & 57.9 & 28.4 & 49.4  \\\grayline


    
    EC & 16.25 & 17.23 & 51.0 & 41.0 & 78.3 & 63.8 & 33.4 & 37.5 & \textbf{69.0} & 31.4 & 54.4 & 56.1 & 29.4 & 49.7 \\

    EC (aux router) & 16.25 & 17.40 & \textbf{52.6} & \textbf{41.5} & 77.3 & 64.4 & 31.4 & 37.5 & 65.0 & 30.9 & 55.4 & 55.8 & \textbf{30.4} & 49.3 \\

    EC (ft TC router) & 16.34 & 16.40 & 49.3 & 41.4 & 78.0 & 64.4 & 33.4 & 37.5 & 67.0 & 30.8 & 56.1 & 55.4 & 29.4 & 49.3 \\

    \bottomrule
    \end{tabular*}

    \vspace{0.5em}
   \caption{\textbf{0.5B params, 40B tokens, 2/64 activated} \;($\bar{T}_e = 512, \ \tileM = 128$)} \label{tab:sparsity:b}
    \vspace{-0.5em}
    \begin{tabular*}{\textwidth}{@{\extracolsep{\fill}}l!{\vrule width 0.4pt}cc!{\vrule width 0.4pt}ccccccccccc!{\vrule width 0.4pt}c}
    \toprule
    
    \textbf{TR} & \textbf{16.22} & \textbf{15.92} & \textbf{51.4} & 41.6 & 78.4 & \textbf{65.4} & 31.6 & \textbf{38.1} & 65.0 & 31.0 & 61.1 & 57.4 & 29.1 &  50.0 \\\grayline

    TC top-$K$ & 16.34 & 15.94 & 51.0 & \textbf{41.9} & 78.5 & 64.8 & 33.0 & \textbf{38.1} & \textbf{67.0} & 30.8 & 54.7 & 55.8 & 30.1 & 49.6 \\

    TC (token drop) & 16.44 & 16.10 & 51.1 & 41.4 & 78.7 & 64.9 & 31.6 & 38.0 & 62.0 & \textbf{32.8} & \textbf{61.9} & \textbf{58.9} & \textbf{30.8} & \textbf{50.2} \\\grayline



    EC & 16.83 & 18.61 & 49.6 & 41.4 & 79.1 & 64.4 & \textbf{33.4} & 36.9 & 62.0 & \textbf{32.8} & 60.2 & 55.8 & 29.1 & 49.5 \\

    EC (aux router) & 16.80 & 21.80 & 50.0 & 40.9 & 75.2 & 63.7 & 28.2 & 35.2 & 61.0 & 31.5 & 57.2 & 53.3 & 24.7 & 47.4 \\

    EC (ft TC router) & 16.81 & 16.98 & 50.0 & 41.7 & \textbf{79.7} & 64.9 & 31.6 & 36.8 & 63.0 & 32.1 & 60.7 & 54.6 & 27.4 & 49.3 \\


    \bottomrule
    \end{tabular*}

    \vspace{0.5em}
    \caption{\textbf{1.8B params, 40B tokens, 8/256 activated} \;($\bar{T}_e = 512, \ \tileM = 128$)} \label{tab:sparsity:c}
    \vspace{-0.5em}
    \begin{tabular*}{\textwidth}{@{\extracolsep{\fill}}l!{\vrule width 0.4pt}cc!{\vrule width 0.4pt}ccccccccccc!{\vrule width 0.4pt}c}
    \toprule
    
    \textbf{TR} & \textbf{13.34} & \textbf{13.10} & 53.4 & 42.1 & 81.7 & 69.6 & \textbf{35.2} & \textbf{45.3} & 70.0 & 33.2 & \textbf{61.4} & 63.0 & \textbf{33.4} & \textbf{53.5} \\\grayline

    TC top-$K$ & 13.51 &  13.12 & 50.1 & \textbf{42.9} & 81.3 & \textbf{69.8} & 33.8 & 45.2 & \textbf{71.0} & \textbf{34.1} & 56.7 & \textbf{64.6} & 31.1 & 52.8 \\

    TC (token drop) & 13.62 & 13.19 & \textbf{55.4} & 41.6 & \textbf{82.2} & 68.6 & 34.8 & 45.0 & 69.0 & 34.0 & 54.4 & 63.5 & 31.4 & 52.7 \\\grayline



    EC & 14.92 & 19.82 & 51.9 & 40.8 & 77.7 & 65.8 & 30.0 & 39.8 & 67.0 & 30.9 & 60.7 & 56.0 & 28.4 &  49.9 \\

    EC (aux router) & 14.94 & 18.01 & 50.6 & 41.8 & 79.8 & 65.8 & 31.6 & 39.3 & 62.0 & 31.8 & 59.7 & 55.8 & 29.8 & 49.8 \\

    EC (ft TC router) & 14.81 & 15.01 & 52.7 & 41.1 & 79.6 & 66.9 & 30.6 & 40.2 & 66.0 & 31.9 & 60.5 & 57.2 & 30.8 & 50.7 \\


    \bottomrule
    \end{tabular*}

    \vspace{0.5em}
    \caption{\textbf{1.4B params, 50B tokens, 8/128 activated} \;($\bar{T}_e = 2048, \ \tileM = 128$)}  \label{tab:sparsity:d}
    \vspace{-0.5em}
    \begin{tabular*}{\textwidth}{@{\extracolsep{\fill}}l!{\vrule width 0.4pt}cc!{\vrule width 0.4pt}ccccccccccc!{\vrule width 0.4pt}c}
    \toprule

    \textbf{TR} & 13.51 & \textbf{13.28} & \textbf{52.6} & \textbf{42.6} & 81.5 & \textbf{69.6} & 33.6 & \textbf{45.4} & 67.0 & 34.8 & 57.3 & 63.7 & 28.1 & 52.4 \\\grayline

    TC top-$K$ & \textbf{13.50} & 13.32 & 51.8 & 41.7 & 81.5 & 69.3 & 32.4 & 45.3 & 68.0 & 34.5 & 56.6 & 63.2 & 28.4 & 52.1 \\

    TC (token drop) & 13.52 & 13.30 & 51.8 & 42.2 & \textbf{84.1} & 69.2 & \textbf{34.4} & 45.2 & \textbf{70.0} & \textbf{35.1} & \textbf{61.2} & \textbf{64.2} & \textbf{31.4} & \textbf{53.5} \\\grayline




    EC & 14.41 & 17.37 & 51.4 & 42.0 & 79.7 & 66.3 & 32.2 & 40.7 & 64.0 & 31.8 & 59.0 & 57.4 & 27.4 & 50.2 \\

    EC (aux router) & 14.34 & 26.96 & 49.8 & 41.5 & 79.1 & 63.1 & 30.2 & 37.6 & 61.0 & 31.0 & 60.9 & 46.7 & 25.1 & 47.8 \\

    EC (ft TC router) & 14.67 & 14.90 & 51.9 & 41.8 & 80.1 & 66.4 & 32.6 & 41.1 & 65.0 & 32.4 & 57.7 & 57.7 & 27.8 & 50.4 \\


    \bottomrule
    \end{tabular*}

    \vspace{0.5em}
    \caption{\textbf{1.4B params, 100B tokens, 2/128 activated} \;($\bar{T}_e = 512, \ \tileM = 128$)}  \label{tab:sparsity:e}
    \vspace{-0.5em}
    \begin{tabular*}{\textwidth}{@{\extracolsep{\fill}}l!{\vrule width 0.4pt}cc!{\vrule width 0.4pt}ccccccccccc!{\vrule width 0.4pt}c}
    \toprule
    
    \textbf{TR} & \textbf{13.31} & \textbf{13.22} & \textbf{52.8} & 41.8 & 80.8 & \textbf{68.7} & 33.0 & \textbf{43.4} & \textbf{67.0} & 33.6 & \textbf{60.2} & 60.7 & 29.8 & \textbf{52.0} \\\grayline

    TC top-$K$  & 13.50 & 13.32 & 51.3 & 42.0 & \textbf{83.2} & 68.2 & \textbf{34.0} & \textbf{43.4} & 66.0 & \textbf{35.4} & 57.9 & \textbf{61.6} & 29.4 & \textbf{52.0} \\

    TC (token drop) & 13.35 & 13.29 & 50.0 & \textbf{42.2} & 81.7 & 68.3 & 31.2 & 43.3 & 66.0 & 34.3 & 56.6 & 59.5 & \textbf{30.8} & 51.3 \\\grayline



    EC & 14.08 & 24.79 & 51.5 & 41.7 & 81.0 & 66.1 & 33.2 & 40.6 & 64.0 & 34.0 & 56.3 & 56.5 & 27.4 & 50.2 \\

    EC (aux router) & 14.01 & 37.52 & 49.7 & 40.2 & 73.6 & 57.5 & 27.6 & 33.2 & 61.0 & 27.8 & 58.8 & 45.2 & 24.2 & 45.3 \\

    EC (ft TC router) & 14.24 & 14.75 & 52.2 & 42.6 & 79.4 & 65.7 & 32.8 & 40.8 & 64.0 & 34.9 & 58.3 & 57.2 & 27.1 & 50.5 \\


    \bottomrule
    \end{tabular*}
    \end{subtable}

\end{table}
\raggedbottom

In this section, we assess the quality of trained MoEs using our token rounding (``TR'') algorithm. We use TR for training and during evaluation we switch to token-choice top-$K$ (``TC top-$K$'') routing. This assesses the capability of replacement of TR with TC after training.\footnote{Token rounding is not a token-choice routing method which creates difficulty for autoregressive generation. Here we do not apply any adaptation and switch to vanilla token choice top-$K$ routing during evaluation/validation.} We use the OLMoE codebase and construct MoE models with OLMoE base architecture \citep{muennighoff2024olmoe}. We use a deduplicated version of FineWeb-Edu \citep{benallal2024smollmcorpus} for training all our models. More details are included in Appendix~\ref{sec:appendix:exp-details}.

We consistently use $\tileM=128$ in Table~\ref{tab:exp-sparse}, and the $\mathrm{round\_and\_sparsify}$ subroutine always rounds the expert frequency to the nearest multiple of $\tileM$ (``NR-f'', see Appendix~\ref{sec:appendix:ablation}). We also use softmax renormalization for TR. We compare TR to token-choice (TC) top-$K$ routing and expert-choice (EC) routing~\citep{zhou2022mixture}. However, EC routing results in future token leakage causing problems for autoregressive generation resulting in a performance drop during evaluation~\citep{raposo2024mixture,wang2024auxiliary}. To address this issue, we consider MoD's approach~\citep{raposo2024mixture} that trains an auxiliary router to predict the EC router's selection during inference\footnote{However, for MoE we are solving a harder \textit{$E$-label prediction problem} instead of MoD's\citep{raposo2024mixture} binary prediction problem. This is because EC router can activate arbitrary number of experts for each token, and we have to independently predict the label for each expert. This approach is likely not scalable for MoE as the prediction problem size scales with $E$. }. This baseline is referred to as ``EC (aux router)'' in each subtable in Table~\ref{tab:exp-sparse}. We also adapt EC routing to TC routing by finetuning a learned TC top-$K$ router and compare its task performance against TR's task performance \textit{without any adaptation}. This is the ``EC (ft TC router)'' baseline in Table~\ref{tab:exp-sparse}. Finally, we consider a token dropping baseline in which we set the capacity of each expert as the largest multiple of $M_\mathrm{tile}$ not exceeding its token frequencies and we discard the tokens with the lowest scores. This is the ``TC (token drop)'' baseline, and we note that this is equivalent as we always round down in TR.

\paragraph{TR's train-test gap.} We validate TR's performance on a 0.5B (subtable~\ref{tab:sparsity:a}) and 1.4B (subtable~\ref{tab:sparsity:d}) MoE model. We then increase the MoE sparsity by either decreasing $K$ while keeping $E$ constant (from~\ref{tab:sparsity:a} to~\ref{tab:sparsity:b}, from~\ref{tab:sparsity:d} to~\ref{tab:sparsity:e}) or increasing $E$ while keeping $K$ constant (from~\ref{tab:sparsity:a} to~\ref{tab:sparsity:c}). Across these sparse MoE configurations, we consistently observe similar model quality between TR and TC. In fact, TR achieves slightly lower validation perplexity and higher or the same average accuracy under the extremely sparse MoE ($K/E\leq 1/32$) settings for the~\ref{tab:sparsity:c} and~\ref{tab:sparsity:e}. There is a noticeable discrepancy between EC and TC as the train and val PPL for EC can have $\gtrapprox$3 gap for~\ref{tab:sparsity:c},\ref{tab:sparsity:d} and~\ref{tab:sparsity:e} compared to TC and TR's usual $\lessapprox$ 0.3 gap. TC finetuning is more effective than the auxiliary router to close this gap, but TR's task evaluation is still always better. In addition, when we compare TR with the token dropping baseline, we also find TR consistently yields lower validation perplexity, and has higher average task accuracy for~\ref{tab:sparsity:a},~\ref{tab:sparsity:c},~\ref{tab:sparsity:e}. In this case, TR can serve as an in-place substitute for TC during training.

\subsubsection{Ablation studies on token rounding routing} \label{sec:exp:token_rounding_ab}

There are 3 variables that can affect the trained MoE quality with token rounding routing: (1) rounding subroutine $\mathrm{round\_and\_sparsify}$ (2) microbatch size $T$, and (3) tile size $\tileM$ for rounding. We analyze their impacts:


\begin{itemize}
    \item \textbf{Choice of rounding subroutine}: In Table~\ref{tab:exp-ablation}, we assess the choice of different routing subroutines to train MoEs using TR. We find that our token rounding algorithm in general is robust to the specific rounding subroutines, and nearest rounding expert frequency to multiples of $\tileM$ (``NR-f'' in Table~\ref{tab:exp-ablation}) is often sufficient for providing an excellent downstream task performance despite its simplicity. Therefore, we choose NR-f as the default rounding subroutine.

    \item \textbf{Effect of microbatch size $T$ and tile size $\tileM$}: The token rounding is applied on the microbatch level so varying the microbatch size $T$ will result in different qualitative results for TR. This also holds true for EC routing. For example, EC over sequence will result in different model quality as EC over a text segment. Nevertheless, in Table~\ref{tab:exp-T-ablation}, we find that TR preserves its trained MoE quality when $\bar{T}_e/\tileM \ge 2$, and even if $\bar{T}_e/\tileM = 1$ (the last row in both subtables), the trained MoE inference quality is still better than training with EC and finetuning with TC top-$K$ routing. Similarly in Table~\ref{tab:exp-tileM-ablation}, we can find that TR is generally robust w.r.t. $\tileM$ when $\bar{T}_e/\tileM \geq 2$. However, when $\bar{T}_e/\tileM=1$ there is a noticeable degradation compared to TC baseline but the model quality is still better than the EC baseline.
\end{itemize}

\begin{figure}[!ht]
    \centering
    \includegraphics[width=\linewidth]{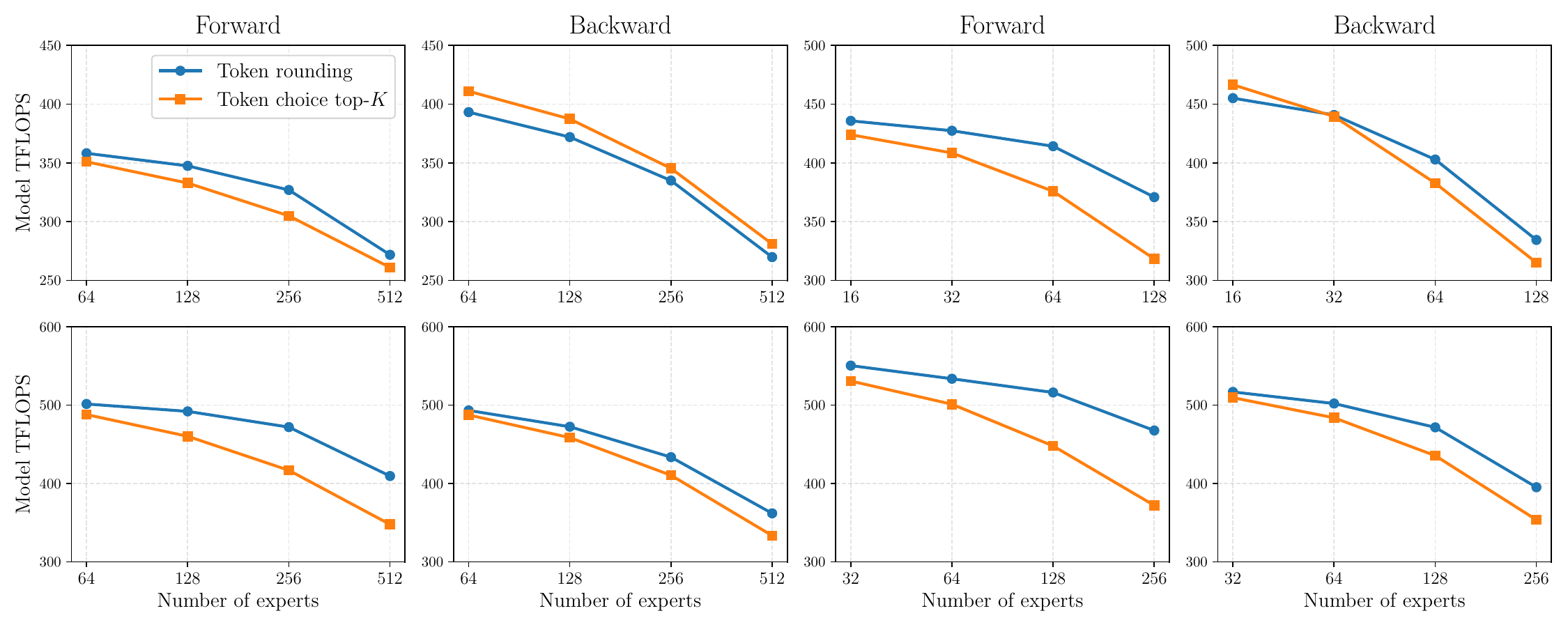}
    \caption{\small Forward \& backward model TFLOPS for \Algname~MoE kernels with different routing methods. We compare TR equipped with ``nearest rounding to $\tileM$-multiples via expert frequency" subroutine against TC top-$K$ routing. Configuration details are in Appendix~\ref{sec:appendix:config}.}
    \label{fig:token_rounding_speed}
\end{figure}

\subsubsection{Token rounding's training throughput} \label{sec:exp:token_round_throughput}

In Figure~\ref{fig:token_rounding_speed}, we benchmark the token rounding's MoE main kernel runtime (without router) against top-$K$ token choice routing. We focus on the iso-FLOPs setting by keeping $T$, $n$ and $K$ constant. We linearly increase the number of experts $E$ while keeping $K$ constant to increase MoE sparsity. As we linearly increase $E$, we observe a drop in TFLOPS for token-choice routing. This is due to the (1) tile quantization effect as the wasted FLOPs spent on padding roughly linearly increases with the MoE sparsity as shown in Figure~\ref{fig:wasted_flops} and (2) the linearly increased IO due to more expert weights. We observe a drop in FLOPs for both TC and TR as we increase $E$, but the drop is more pronounced for TC as shown in Figure~\ref{fig:token_rounding_speed}.


For the $3^\text{rd}$ and $4^\text{th}$ column in the top row in Figure~\ref{fig:token_rounding_speed}, an MoE model with 128 experts ($K/E = 1/64$) and $n = 1k$ with token rounding routing achieves 16.5\% model TFLOPS\footnote{Note that the consumed FLOPs are calculated from $(6+12) dn (\sum_e f_e)$ (note that $\sum_e f_e = TK$ for TC top-$K$ routing) as model FLOPs rather than hardware FLOPs. The speedup behind TR is to preserve the model FLOPs consumption on expectation but save the hardware FLOPs consumption by removing padding wastes, which in turn leads to model TFLOPS speedup. } improvement in forward and 6.1\% in backward, resulting in an end-to-end improvement of 9.4\%. For the $3^\text{rd}$ and $4^\text{th}$ column on the bottom row in Figure~\ref{fig:token_rounding_speed}, when we have a MoE with 256 experts ($K/E=1/64$), token rounding routing achieves a 25.7\% TFLOPS improvement in forward and 11.8\% in backward resulting in an end-to-end improvement of 15.9\%. In general, \textbf{we observe that as we move to larger intermediate sizes (more compute-bound) and higher MoE sparsity, the gap between TR and TC top-$K$ becomes larger.}

\begin{figure}[!ht]
    \centering
    \includegraphics[width=\linewidth]{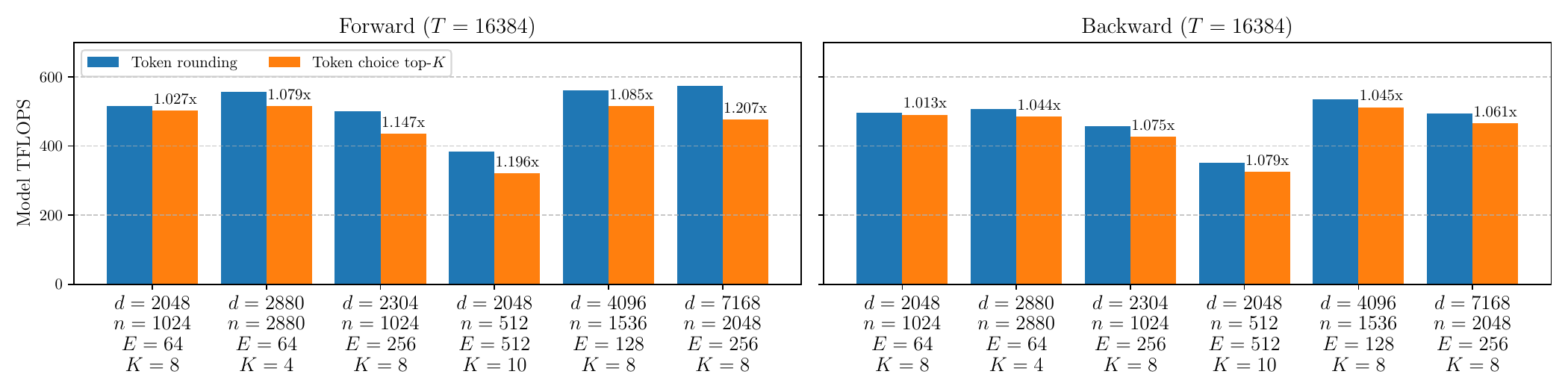}
    \caption{\small Forward \& backward TFLOPS of a single MoE layer of \Algname~equipped with different routing methods for different configurations ranging from 7B to 685B parameters on H100. The MoE configurations from left to right adopt the model size of \model{OLMoE-1B-7B-0125}, \model{gpt-oss-20b}, \model{Kimi-Linear-48B-A3B-Base}, \model{Qwen3-Next-80B-A3B-Thinking}, \model{Qwen3-235B-A22B-Thinking-2507}, and \model{DeepSeek-V3.2-Exp} (configurations identical to Figure~\ref{fig:real-moe-speed}). We compare TR equipped with ``nearest rounding to $\tileM$-multiples via expert frequency" subroutine against TC top-$K$ routing. }
    \label{fig:token_rounding_real_speed}
\end{figure}

This trend also holds with configurations from recent open-source MoEs in Figure~\ref{fig:token_rounding_real_speed}. When we equip \Algname's MoE kernel with TR router instead of TC top-$K$ router, we observe a larger relative speedup for highly sparse MoEs such as \model{Qwen3-Next-80B-A3B-Thinking} ($K/E=10/512$), where TR achieves 19.6\% and 7.9\% speedup over TC top-$K$ router during forward and backward pass respectively.

\section{Conclusion}
We present \Algname, a co-design solution that jointly optimizes MoE architecture and GPU kernels to address the training challenges of granular and sparse MoEs.
Our contributions include: (1) a memory-efficient algorithm that minimizes
activation size as MoEs become more fine-grained, (2) GPU kernels that overlap IO with computation for throughput improvement, and (3) tile-aware token rounding that yields additional speedup without quality loss.
Future directions include extending to low-precision and microscaling formats
(FP8, MXFP8, MXFP4) for further memory savings, and overlapping communication
with computation in distributed settings like expert parallelism. We envision future model architecture designs that optimize for quality per compute hour rather than just quality per FLOP—jointly considering algorithmic and hardware efficiency.

\section*{Acknowledgment}
We gratefully acknowledge the support of Schmidt Sciences AI2050 fellowship, the Google ML and Systems Junior Faculty Awards, and the Google Research Scholar program. We also thank the Princeton Language Intelligence program for the computing resources support. We thank Shawn Tan for his generous support on our experiments. We also thank Songlin Yang, Yilong Zhao, Bharat Runwal, Xinyu Yang, Andrew Sheinberg, Lijie Yang, Yongye Zhu, Zhuoqing Song and numerous anonymous reviewers for providing valuable feedback.

\printbibliography

@misc{shazeer2017outrageouslylargeneuralnetworks,
      title={Outrageously Large Neural Networks: The Sparsely-Gated Mixture-of-Experts Layer}, 
      author={Noam Shazeer and Azalia Mirhoseini and Krzysztof Maziarz and Andy Davis and Quoc Le and Geoffrey Hinton and Jeff Dean},
      year={2017},
      eprint={1701.06538},
      archivePrefix={arXiv},
      primaryClass={cs.LG},
      url={https://arxiv.org/abs/1701.06538}, 
}

@misc{vaswani2023attentionneed,
      title={Attention Is All You Need}, 
      author={Ashish Vaswani and Noam Shazeer and Niki Parmar and Jakob Uszkoreit and Llion Jones and Aidan N. Gomez and Lukasz Kaiser and Illia Polosukhin},
      year={2017},
      eprint={1706.03762},
      archivePrefix={arXiv},
      primaryClass={cs.CL},
      url={https://arxiv.org/abs/1706.03762}, 
}

@misc{kimiteam2025kimik2openagentic,
      title={Kimi K2: Open Agentic Intelligence}, 
      author={Team Kimi and Yifan Bai and Yiping Bao and Guanduo Chen and et al.},
      year={2025},
      eprint={2507.20534},
      archivePrefix={arXiv},
      primaryClass={cs.LG},
      url={https://arxiv.org/abs/2507.20534}, 
}

@misc{deepseekai2025deepseekv3technicalreport,
      title={DeepSeek-V3 Technical Report}, 
      author={DeepSeek-AI and Aixin Liu and Bei Feng and Bing Xue and et al.},
      year={2024},
      eprint={2412.19437},
      archivePrefix={arXiv},
      primaryClass={cs.CL},
      url={https://arxiv.org/abs/2412.19437}, 
}

@techreport{nvidia2022_h100,
  title        = {NVIDIA H100 Tensor Core GPU Architecture: Exceptional Performance, Scalability, and Security for the Data Center},
  author       = {NVIDIA},
  institution  = {NVIDIA},
  type         = {Whitepaper},
  number       = {V1.01},
  year         = {2022},
  month        = {March},
  note         = {Grace Hopper “Hopper” Architecture},
  url          = {https://www.advancedclustering.com/wp-content/uploads/2022/03/gtc22-whitepaper-hopper.pdf}
}

@techreport{nvidia2022_gb200,
  title        = {NVIDIA Blackwell Architecture Technical Brief},
  author       = {NVIDIA},
  institution  = {NVIDIA},
  type         = {Whitepaper},
  note         = {Blackwell Architecture},
  url          = {https://resources.nvidia.com/en-us-blackwell-architecture?ncid=no-ncid},
  year         = {2025}    
}

@article{gale2023megablocks,
  title={Megablocks: Efficient sparse training with mixture-of-experts},
  author={Gale, Trevor and Narayanan, Deepak and Young, Cliff and Zaharia, Matei},
  journal={Proceedings of Machine Learning and Systems},
  volume={5},
  pages={288--304},
  year={2023}
}

@article{tan2024scattered,
  title={Scattered mixture-of-experts implementation},
  author={Tan, Shawn and Shen, Yikang and Panda, Rameswar and Courville, Aaron},
  journal={arXiv preprint arXiv:2403.08245},
  year={2024}
}

@misc{paszke2019pytorchimperativestylehighperformance,
      title={PyTorch: An Imperative Style, High-Performance Deep Learning Library}, 
      author={Adam Paszke and Sam Gross and Francisco Massa and Adam Lerer and James Bradbury and Gregory Chanan and Trevor Killeen and Zeming Lin and Natalia Gimelshein and Luca Antiga and Alban Desmaison and Andreas Köpf and Edward Yang and Zach DeVito and Martin Raison and Alykhan Tejani and Sasank Chilamkurthy and Benoit Steiner and Lu Fang and Junjie Bai and Soumith Chintala},
      year={2019},
      eprint={1912.01703},
      archivePrefix={arXiv},
      primaryClass={cs.LG},
      url={https://arxiv.org/abs/1912.01703}, 
}

@inproceedings{triton,
author = {Tillet, Philippe and Kung, H. T. and Cox, David},
title = {Triton: an intermediate language and compiler for tiled neural network computations},
year = {2019},
isbn = {9781450367196},
publisher = {Association for Computing Machinery},
address = {New York, NY, USA},
url = {https://doi.org/10.1145/3315508.3329973},
doi = {10.1145/3315508.3329973},
booktitle = {Proceedings of the 3rd ACM SIGPLAN International Workshop on Machine Learning and Programming Languages},
pages = {10–19},
numpages = {10},
location = {Phoenix, AZ, USA},
series = {MAPL 2019}
}

@misc{shah2024flashattention3fastaccurateattention,
      title={FlashAttention-3: Fast and Accurate Attention with Asynchrony and Low-precision}, 
      author={Jay Shah and Ganesh Bikshandi and Ying Zhang and Vijay Thakkar and Pradeep Ramani and Tri Dao},
      year={2024},
      eprint={2407.08608},
      archivePrefix={arXiv},
      primaryClass={cs.LG},
      url={https://arxiv.org/abs/2407.08608}, 
}

@misc{dbrx2024,
  author       = {The Mosaic Research Team, Databricks},
  title        = {Introducing DBRX: A New State-of-the-Art Open LLM},
  year         = {2024},
  howpublished = {\url{https://www.databricks.com/blog/introducing-dbrx-new-state-art-open-llm}},
  note         = {Databricks Blog, March 27, 2024}
}

@misc{fedus2022switchtransformersscalingtrillion,
      title={Switch Transformers: Scaling to Trillion Parameter Models with Simple and Efficient Sparsity}, 
      author={William Fedus and Barret Zoph and Noam Shazeer},
      year={2022},
      eprint={2101.03961},
      archivePrefix={arXiv},
      primaryClass={cs.LG},
      url={https://arxiv.org/abs/2101.03961}, 
}

@misc{nvidia_cutlass,
  author       = {NVIDIA},
  title        = {{CUTLASS: CUDA Templates for Linear Algebra Subroutines}},
  year         = {2025},
  howpublished = {\url{https://github.com/NVIDIA/cutlass}},
  note         = {Version 4.2.0, Accessed: 2025-09-19}
}

@misc{mistral2024mixtral,
  author       = {Mistral},
  title        = {Mixtral of Experts: 
A high quality Sparse Mixture-of-Experts.},
  year         = {2024},
  howpublished = {\url{https://mistral.ai/news/mixtral-of-experts}},
  note         = {Mistral AI News}
}

@misc{azure2024phi35moe,
  author       = {Microsoft},
  title        = {Announcing the Availability of PHI-3.5 MoE in Azure AI Studio and GitHub},
  year         = {2024},
  howpublished = {\url{https://techcommunity.microsoft.com/blog/azure-ai-foundry-blog/announcing-the-availability-of-phi-3-5-moe-in-azure-ai-studio-and-github/4256278}},
  note         = {Microsoft Tech Community}
}

@misc{qwen3_2024,
  author       = {QwenLM},
  title        = {Qwen3: Think Deeper, Act Faster},
  year         = {2025},
  howpublished = {\url{https://qwenlm.github.io/blog/qwen3/}},
  note         = {Official Blog}
}

@misc{openai2025gptoss120bgptoss20bmodel,
      title={gpt-oss-120b \& gpt-oss-20b Model Card}, 
      author={OpenAI},
      year={2025},
      eprint={2508.10925},
      archivePrefix={arXiv},
      primaryClass={cs.CL},
      url={https://arxiv.org/abs/2508.10925}, 
}

@article{zeng2025glm,
  title={Glm-4.5: Agentic, reasoning, and coding (arc) foundation models},
  author={Zeng, Aohan and Lv, Xin and Zheng, Qinkai and Hou, Zhenyu and Chen, Bin and Xie, Chengxing and Wang, Cunxiang and Yin, Da and Zeng, Hao and Zhang, Jiajie and others},
  journal={arXiv preprint arXiv:2508.06471},
  year={2025}
}

@misc{granite3.1_language_models,
  author       = {IBM Granite},
  title        = {Granite 3.1 Language Models},
  year         = {2024},
  howpublished = {\url{https://github.com/ibm-granite/granite-3.1-language-models}},
  note         = {GitHub repository}
}

@article{tian2025towards,
  title={Towards greater leverage: Scaling laws for efficient mixture-of-experts language models},
  author={Tian, Changxin and Chen, Kunlong and Liu, Jia and Liu, Ziqi and Zhang, Zhiqiang and Zhou, Jun},
  journal={arXiv preprint arXiv:2507.17702},
  year={2025}
}

@article{krajewski2024scaling,
  title={Scaling laws for fine-grained mixture of experts},
  author={Krajewski, Jakub and Ludziejewski, Jan and Adamczewski, Kamil and Pi{\'o}ro, Maciej and Krutul, Micha{\l} and Antoniak, Szymon and Ciebiera, Kamil and Kr{\'o}l, Krystian and Odrzyg{\'o}{\'z}d{\'z}, Tomasz and Sankowski, Piotr and others},
  journal={arXiv preprint arXiv:2402.07871},
  year={2024}
}

@article{he2024mixture,
  title={Mixture of a million experts},
  author={He, Xu Owen},
  journal={arXiv preprint arXiv:2407.04153},
  year={2024}
}

@article{berges2024memory,
  title={Memory layers at scale},
  author={Berges, Vincent-Pierre and O{\u{g}}uz, Barlas and Haziza, Daniel and Yih, Wen-tau and Zettlemoyer, Luke and Ghosh, Gargi},
  journal={arXiv preprint arXiv:2412.09764},
  year={2024}
}

@inproceedings{huangultra,
  title={Ultra-Sparse Memory Network},
  author={Huang, Zihao and Min, Qiyang and Huang, Hongzhi and Zeng, Yutao and Zhu, Defa and Guo, Ran and others},
  booktitle={The Thirteenth International Conference on Learning Representations},
  year={2025}
}

@article{costin2025momoe,
    title={MoMoE: Memory optimized Mixture of Experts},
    author={Costin, Bobby and Averbuch, Timor and Pai, Dhruv and Chen, Nathan and Keigwin, Ben},
    journal={Tilde Research Blog},
    year={2025},
    month={7},
    url={https://www.tilderesearch.com/blog/momoe},
    note={Blog post}
}

@online{wright2024cutlass,
  author    = {Less Wright and Adnan Hoque},
  title     = {Deep Dive on CUTLASS Ping-Pong GEMM Kernel},
  year      = {2024},
  month     = {November},
  day       = {1},
  url       = {https://docs.pytorch.org/blog/cutlass-ping-pong-gemm-kernel/},
  note      = {Accessed: 2025-09-21}
}

@online{deepgemm,
  author    = {Chenggang Zhao and Liang Zhao and Jiashi Li and Zhean Xu},
  title     = {DeepGEMM: clean and efficient FP8 GEMM kernels with fine-grained scaling},
  year      = {2025},
  url       = {https://github.com/deepseek-ai/DeepGEMM},
}

@online{determinism,
  author    = {Horace He and Thinking Machines},
  title     = {Defeating Nondeterminism in LLM Inference},
  year      = {2025},
  url       = {https://thinkingmachines.ai/blog/defeating-nondeterminism-in-llm-inference/#true-on-policy-rl},
}

@inproceedings{deepseekv3insights,
author = {Zhao, Chenggang and Deng, Chengqi and Ruan, Chong and Dai, Damai and Gao, Huazuo and Li, Jiashi and Zhang, Liyue and Huang, Panpan and Zhou, Shangyan and Ma, Shirong and Liang, Wenfeng and He, Ying and Wang, Yuqing and Liu, Yuxuan and Wei, Y.X.},
title = {Insights into DeepSeek-V3: Scaling Challenges and Reflections on Hardware for AI Architectures},
year = {2025},
isbn = {9798400712616},
publisher = {Association for Computing Machinery},
address = {New York, NY, USA},
url = {https://doi.org/10.1145/3695053.3731412},
doi = {10.1145/3695053.3731412},
booktitle = {Proceedings of the 52nd Annual International Symposium on Computer Architecture},
pages = {1731–1745},
numpages = {15},
location = {
},
series = {ISCA '25}
}

@online{cutedsl,
  author    = {NVIDIA},
  title     = {NVIDIA CUTLASS Documentation},
  year      = {2025},
  url       = {https://docs.nvidia.com/cutlass/media/docs/pythonDSL/cute_dsl_general/dsl_introduction.html},
}

@inproceedings{muennighoff2024olmoe,
  title={OLMoE: Open Mixture-of-Experts Language Models},
  author={Muennighoff, Niklas and Soldaini, Luca and Groeneveld, Dirk and Lo, Kyle and Morrison, Jacob and Min, Sewon and Shi, Weijia and Walsh, Evan Pete and Tafjord, Oyvind and Lambert, Nathan and others},
  booktitle={The Thirteenth International Conference on Learning Representations},
  year={2025}
}

@inproceedings{sakaguchi2020winogrande,
  title={WinoGrande: An Adversarial Winograd Schema Challenge at Scale},
  author={Sakaguchi, Keisuke and Le Bras, Ronan and Bhagavatula, Chandra and Choi, Yejin},
  booktitle={Proceedings of the AAAI Conference on Artificial Intelligence},
  volume={34},
  pages={8732--8740},
  year={2020}
}

@inproceedings{sap2019socialiqa,
  title={SocialIQA: Commonsense Reasoning about Social Interactions},
  author={Sap, Maarten and Rashkin, Hannah and Chen, Derek and LeBras, Ronan and Choi, Yejin},
  booktitle={Conference on Empirical Methods in Natural Language Processing},
  year={2019}
}

@inproceedings{SciQ,
    title={Crowdsourcing Multiple Choice Science Questions},
    author={Johannes Welbl, Nelson F. Liu, Matt Gardner},
    year={2017},
    journal={arXiv:1707.06209v1}
}

@inproceedings{piqa,
  author = {Yonatan Bisk and Rowan Zellers and
            Ronan Le Bras and Jianfeng Gao
            and Yejin Choi},
  title = {PIQA: Reasoning about Physical Commonsense in
           Natural Language},
  booktitle = {Thirty-Fourth AAAI Conference on
               Artificial Intelligence},
  year = {2020},
}

@inproceedings{OpenBookQA2018,
 title={Can a Suit of Armor Conduct Electricity? A New Dataset for Open Book Question Answering},
 author={Todor Mihaylov and Peter Clark and Tushar Khot and Ashish Sabharwal},
 booktitle={EMNLP},
 year={2018}
}

@inproceedings{zellers2019hellaswag,
    title={HellaSwag: Can a Machine Really Finish Your Sentence?},
    author={Zellers, Rowan and Holtzman, Ari and Bisk, Yonatan and Farhadi, Ali and Choi, Yejin},
    booktitle ={Proceedings of the 57th Annual Meeting of the Association for Computational Linguistics},
    year={2019}
}

@inproceedings{roemmele2011choice,
  title={Choice of plausible alternatives: An evaluation of commonsense causal reasoning},
  author={Roemmele, Melissa and Bejan, Cosmin Adrian and Gordon, Andrew S},
  booktitle={2011 AAAI Spring Symposium Series},
  year={2011}
}

@inproceedings{csqa,
    title = "{C}ommonsense{QA}: A Question Answering Challenge Targeting Commonsense Knowledge",
    author = "Talmor, Alon  and
      Herzig, Jonathan  and
      Lourie, Nicholas  and
      Berant, Jonathan",
    booktitle = "Proceedings of the 2019 Conference of the North {A}merican Chapter of the Association for Computational Linguistics: Human Language Technologies, Volume 1 (Long and Short Papers)",
    month = jun,
    year = "2019",
    address = "Minneapolis, Minnesota",
    publisher = "Association for Computational Linguistics",
    url = "https://aclanthology.org/N19-1421",
    doi = "10.18653/v1/N19-1421",
    pages = "4149--4158",
    archivePrefix = "arXiv",
    eprint        = "1811.00937",
    primaryClass  = "cs",
}

@inproceedings{boolq,
  title={BoolQ: Exploring the Surprising Difficulty of Natural Yes/No Questions},
  author={Clark, Christopher and Lee, Kenton and Chang, Ming-Wei and Kwiatkowski, Tom and Collins, Michael and Toutanova, Kristina},
  booktitle={Proceedings of the 2019 Conference of the North American Chapter of the Association for Computational Linguistics: Human Language Technologies, Volume 1 (Long and Short Papers)},
  pages={2924--2936},
  year={2019}
}

@article{arc,
      author    = {Peter Clark  and Isaac Cowhey and Oren Etzioni and Tushar Khot and
                    Ashish Sabharwal and Carissa Schoenick and Oyvind Tafjord},
      title     = {Think you have Solved Question Answering? Try ARC, the AI2 Reasoning Challenge},
      journal   = {arXiv:1803.05457v1},
      year      = {2018},
}

@software{benallal2024smollmcorpus,
  author = {Ben Allal, Loubna and Lozhkov, Anton and Penedo, Guilherme and Wolf, Thomas and von Werra, Leandro},
  title = {SmolLM-Corpus},
  year = 2024,
  url = {https://huggingface.co/datasets/HuggingFaceTB/smollm-corpus}
}

@misc{qwen3technicalreport,
      title={Qwen3 Technical Report}, 
      author={Team Qwen},
      year={2025},
      eprint={2505.09388},
      archivePrefix={arXiv},
      primaryClass={cs.CL},
      url={https://arxiv.org/abs/2505.09388}, 
}

@article{zhou2022mixture,
  title={Mixture-of-experts with expert choice routing},
  author={Zhou, Yanqi and Lei, Tao and Liu, Hanxiao and Du, Nan and Huang, Yanping and Zhao, Vincent and Dai, Andrew M and Le, Quoc V and Laudon, James and others},
  journal={Advances in Neural Information Processing Systems},
  volume={35},
  pages={7103--7114},
  year={2022}
}

@inproceedings{zeng-etal-2024-adamoe,
    title = "{A}da{M}o{E}: Token-Adaptive Routing with Null Experts for Mixture-of-Experts Language Models",
    author = "Zeng, Zihao  and
      Miao, Yibo  and
      Gao, Hongcheng  and
      Zhang, Hao  and
      Deng, Zhijie",
    editor = "Al-Onaizan, Yaser  and
      Bansal, Mohit  and
      Chen, Yun-Nung",
    booktitle = "Findings of the Association for Computational Linguistics: EMNLP 2024",
    month = nov,
    year = "2024",
    address = "Miami, Florida, USA",
    publisher = "Association for Computational Linguistics",
    url = "https://aclanthology.org/2024.findings-emnlp.361/",
    doi = "10.18653/v1/2024.findings-emnlp.361",
    pages = "6223--6235"
}

@inproceedings{clark2022unified,
  title={Unified scaling laws for routed language models},
  author={Clark, Aidan and de Las Casas, Diego and Guy, Aurelia and Mensch, Arthur and Paganini, Michela and Hoffmann, Jordan and Damoc, Bogdan and Hechtman, Blake and Cai, Trevor and Borgeaud, Sebastian and others},
  booktitle={International conference on machine learning},
  pages={4057--4086},
  year={2022},
  organization={PMLR}
}

@article{raposo2024mixture,
  title={Mixture-of-depths: Dynamically allocating compute in transformer-based language models},
  author={Raposo, David and Ritter, Sam and Richards, Blake and Lillicrap, Timothy and Humphreys, Peter Conway and Santoro, Adam},
  journal={arXiv preprint arXiv:2404.02258},
  year={2024}
}

@article{wang2024auxiliary,
  title={Auxiliary-loss-free load balancing strategy for mixture-of-experts},
  author={Wang, Lean and Gao, Huazuo and Zhao, Chenggang and Sun, Xu and Dai, Damai},
  journal={arXiv preprint arXiv:2408.15664},
  year={2024}
}

@inproceedings{huang2024harder,
  title={Harder Task Needs More Experts: Dynamic Routing in MoE Models},
  author={Huang, Quzhe and An, Zhenwei and Zhuang, Nan and Tao, Mingxu and Zhang, Chen and Jin, Yang and Xu, Kun and Chen, Liwei and Huang, Songfang and Feng, Yansong},
  booktitle={Proceedings of the 62nd Annual Meeting of the Association for Computational Linguistics (Volume 1: Long Papers)},
  pages={12883--12895},
  year={2024}
}

@article{zoph2022st,
  title={St-moe: Designing stable and transferable sparse expert models},
  author={Zoph, Barret and Bello, Irwan and Kumar, Sameer and Du, Nan and Huang, Yanping and Dean, Jeff and Shazeer, Noam and Fedus, William},
  journal={arXiv preprint arXiv:2202.08906},
  year={2022}
}

@article{shoeybi2019megatron,
  title={Megatron-lm: Training multi-billion parameter language models using model parallelism},
  author={Shoeybi, Mohammad and Patwary, Mostofa and Puri, Raul and LeGresley, Patrick and Casper, Jared and Catanzaro, Bryan},
  journal={arXiv preprint arXiv:1909.08053},
  year={2019}
}

@inproceedings{
  xie2025rtopk,
  title={RTop-K: Ultra-Fast Row-Wise Top-K Selection for Neural Network Acceleration on GPUs},
  author={Xi Xie and Yuebo Luo and Hongwu Peng and Caiwen Ding},
  booktitle={The Thirteenth International Conference on Learning Representations},
  year={2025},
  url={https://openreview.net/forum?id=PHg4rAXFVH}
}

@misc{sorting, 
 url={https://bertdobbelaere.github.io/sorting_networks.html}, 
 journal={List of sorting networks}, 
 author={Dobbelaere, Bert},
 year={2025}
}

@misc{torch_cutlass_pingpong, 
    title={Deep dive on Cutlass Ping-Pong Gemm Kernel}, 
    url={https://pytorch.org/blog/cutlass-ping-pong-gemm-kernel/}, 
    journal={PyTorch}, 
    author={Wright, Less and Hoque, Adnan}, 
    year={2024}, 
    month={Nov}
}

@article{lu2022grab,
  title={Grab: Finding provably better data permutations than random reshuffling},
  author={Lu, Yucheng and Guo, Wentao and De Sa, Christopher M},
  journal={Advances in Neural Information Processing Systems},
  volume={35},
  pages={8969--8981},
  year={2022}
}

@article{dwivedi2024kernel,
  title={Kernel thinning},
  author={Dwivedi, Raaz and Mackey, Lester},
  journal={Journal of Machine Learning Research},
  volume={25},
  number={152},
  pages={1--77},
  year={2024}
}

@article{cooper2023coordinating,
  title={Coordinating distributed example orders for provably accelerated training},
  author={Cooper, A Feder and Guo, Wentao and Pham, Duc Khiem and Yuan, Tiancheng and Ruan, Charlie and Lu, Yucheng and De Sa, Christopher M},
  journal={Advances in Neural Information Processing Systems},
  volume={36},
  pages={56198--56210},
  year={2023}
}

@article{wang2025tilelang,
  title={TileLang: A Composable Tiled Programming Model for AI Systems},
  author={Wang, Lei and Cheng, Yu and Shi, Yining and Tang, Zhengju and Mo, Zhiwen and Xie, Wenhao and Ma, Lingxiao and Xia, Yuqing and Xue, Jilong and Yang, Fan and others},
  journal={arXiv preprint arXiv:2504.17577},
  year={2025}
}

@online{colfax_cutlass_tmem,
  title={CUTLASS Tutorial: Writing GEMM Kernels Using Tensor Memory For NVIDIA Blackwell GPUs},
  author={Colfax Research},
  year={2024},
  url={https://research.colfax-intl.com/cutlass-tutorial-writing-gemm-kernels-using-tensor-memory-for-nvidia-blackwell-gpus/},
  note={Accessed: 2025-09-21}
}

@misc{CUTLASS_gemm, 
    url={https://docs.nvidia.com/cutlass/media/docs/cpp/cutlass_3x_backwards_compatibility.html}, 
    journal={CUTLASS 3.0 GEMM Backwards Compatibility - NVIDIA CUTLASS Documentation}, 
    author={NVIDIA Corporation},
    year={2025}, 
    month={Sep}
}

@article{Blas,
  title={Basic linear algebra subprograms for Fortran usage},
  author={Lawson, Chuck L and Hanson, Richard J. and Kincaid, David R and Krogh, Fred T.},
  journal={ACM Transactions on Mathematical Software (TOMS)},
  volume={5},
  number={3},
  pages={308--323},
  year={1979},
  publisher={ACM New York, NY, USA}
}

@misc{pingpongandcooperative, 
    title={Efficient GEMM in CUDA}, 
    url={https://docs.nvidia.com/cutlass/media/docs/cpp/efficient_gemm.html#hopper-warp-specialization}, 
    journal={NVIDIA CUTLASS Documentation}, 
    publisher={NVIDIA}, 
    year={2025}
}

@misc{deepseekai2024deepseekv32,
      title={DeepSeek-V3.2-Exp: Boosting Long-Context Efficiency with DeepSeek Sparse Attention}, 
      author={DeepSeek-AI},
      year={2025},
}

@inproceedings{tmaadatpivefp8,
  title={TMA-Adaptive FP8 Grouped GEMM: Eliminating Padding Requirements in Low-Precision Training and Inference on Hopper},
  author={Fu, Rong and Cao, Weihan and Gao, Jianfei and Jin, Minxi and Wang, Hui and others},
  booktitle={ES-FoMo III: 3rd Workshop on Efficient Systems for Foundation Models},
  year={2025}
}

@inproceedings{zeng2024turn,
  title={Turn Waste into Worth: Rectifying Top-k Router of MoE},
  author={Zeng, Zhiyuan and Guo, Qipeng and Fei, Zhaoye and Yin, Zhangyue and Zhou, Yunhua and Li, Linyang and Sun, Tianxiang and Yan, Hang and Lin, Dahua and Qiu, Xipeng},
  booktitle={EMNLP},
  year={2024}
}

@software{Mishra_lm_engine_A_2024,
    author = {Mishra, Mayank},
    month = jun,
    title = {{LM Engine: A Hyper-Optimized Library for Pretraining and Finetuning}},
    url = {https://github.com/ibm/lm-engine},
    year = {2024}
}

@inproceedings{bitonicsort,
    author = {Batcher, K. E.},
    title = {Sorting networks and their applications},
    year = {1968},
    isbn = {9781450378970},
    publisher = {Association for Computing Machinery},
    address = {New York, NY, USA},
    url = {https://doi.org/10.1145/1468075.1468121},
    doi = {10.1145/1468075.1468121},
    booktitle = {Proceedings of the April 30--May 2, 1968, Spring Joint Computer Conference},
    pages = {307–314},
    numpages = {8},
    location = {Atlantic City, New Jersey},
    series = {AFIPS '68 (Spring)}
}

@misc{team2025kimi,
    title         = {Kimi Linear: An Expressive, Efficient Attention Architecture},
    author        = {Zhang, Yu  and Lin, Zongyu  and Yao, Xingcheng  and et al.},
    year          = {2025},
    eprint        = {2510.26692},
    archivePrefix = {arXiv},
    primaryClass  = {cs.CL}
}

@misc{deepep2025,
      title={DeepEP: an efficient expert-parallel communication library},
      author={Chenggang Zhao and Shangyan Zhou and Liyue Zhang and Chengqi Deng and Zhean Xu and Yuxuan Liu and Kuai Yu and Jiashi Li and Liang Zhao},
      year={2025},
      publisher = {GitHub},
      howpublished = {\url{https://github.com/deepseek-ai/DeepEP}},
}

@inproceedings{lepikhingshard,
  title={GShard: Scaling Giant Models with Conditional Computation and Automatic Sharding},
  author={Lepikhin, Dmitry and Lee, HyoukJoong and Xu, Yuanzhong and Chen, Dehao and Firat, Orhan and Huang, Yanping and Krikun, Maxim and Shazeer, Noam and Chen, Zhifeng},
  booktitle={International Conference on Learning Representations},
 year={2024}
}

\appendix
\section*{Appendix}
\appendix
We provide a table listing all notations and their explanations in Table~\ref{tab:notations}. In Section~\ref{sec:appendix:kernel_comparison}, we compare~\Algname's kernel design with other open-source MoE kernel designs. In Section~\ref{sec:appendix:dH_proof}, we elaborate on \Algname's computational path for $dS$ and $dH$ that does not use $Y$ and $dY$. In Section~\ref{sec:appendix:dS_computation}, we justify \Algname's computational path for $dS$ is both activation memory and computationally-efficient. In Section~\ref{sec:appendix:top-k}, We examine \Algname's top-$K$ sorting kernel. In Table~\ref{tab:moe-scaling}, we provide a trending overview for open-source frontier MoE models. We present \Algname's expert aggregation strategy in Figure~\ref{fig:expert-agg}. Figure~\ref{fig:expert-agg-why-slow} illustrates that on Hopper GPUs, asynchronous TMA store (top) has higher memory bandwidth and can naturally overlap with TensorCore MMA. In addition, \Algname's up-projection backward is included in Algorithm~\ref{alg:up-bwd-moe}. In Section~\ref{sec:appendix:kernel_ablation}, we present ablation studies of training throughput for \Algname's MoE computation kernels to examine the impact of each design choice made for~\Algname. In Section~\ref{sec:appendix:more_experiments}, we assess the quality improvements of MoE models trained by varying expert granularity. We then focus on various ablation studies on our token rounding routing algorithm to assess the quality difference of the trained MoE models from the choice of rounding subroutine. We also study the effect of microbatch size $T$ and tile size $\tileM$ on token rounding. In Section~\ref{sec:appendix:config}, we describe the configurations for benchmarking the memory and training throughput. In Section~\ref{sec:appendix:exp-details}, we include the details of model training and the evaluation setup. 


\clearpage
\section{Notations}\label{sec:appendix:notations}
In Table~\ref{tab:notations}, we describe the notations used in this paper.

\begin{table*}[!ht]
    \setlength{\tabcolsep}{2pt} 
    \renewcommand{\arraystretch}{0} 
    \fontsize{8pt}{8pt}\selectfont
    \caption{\small Notations and their explanations}\label{tab:notations}
    \begin{center}
    \centering
    \begin{tabular}{p{0.15\linewidth}p{0.0\linewidth}p{0.80\linewidth}}
    \toprule
    \textbf{Notations} && \textbf{Explanation} \\
    \midrule
     $T$ & & number of tokens in a microbatch \\\midrule
     
     $d$ & & model embedding dimension (hidden size)  \\\midrule

     $n$ & & each expert's intermediate dimension \\\midrule

     $E$ &&  total number of experts 
     \\ \midrule
  
    $K$ && number of activated experts \\ \midrule

    $\rho$ && $\rho = K/E$ represents MoE activation ratio \\ \midrule

     $\bar{T}_e$ & & $\bar{T}_e = \mathbb{E}_{e\in[E]} [T_e] =  T \rho$ represents the expected number of received tokens in each microbatch by expert $e$ \\\midrule

    $G$ && $G = d/n$ represents the MoE expert granularity. Greater $G$ means a MoE is more fine-grained \\ \midrule

    $\mathbf{M}, \mathbf{N}, \mathbf{K}$ && Dimensions for GEMM in CUTLASS. We define $A \in \mathbb{R}^{\mathbf{M} \times \mathbf{K}}$, $B \in \mathbb{R}^{\mathbf{K} \times \mathbf{N}}$, and $C \in \mathbb{R}^{\mathbf{M} \times \mathbf{N}}$ for $A B = C$  \\ \midrule

    $\tileM$, $\tileN$, $\tileK$ && tile size of $\mathbf{M}$, $\mathbf{N}$, $\mathbf{K}$ dimension for a single GEMM tile \\ \midrule

    $R_e$  &&  tile quantization residue $R_e := T_e \bmod \tileM$ \\ \midrule



    $X$ &&  $X \in \mathbb{R}^{T \times d}$, input token embeddings for an MoE layer \\ \midrule

  $\W{1}$ &&  $\W{1} \in \mathbb{R}^{E \times d\times 2n}$,  weight of up projection \\ \midrule
  
  $\W{2}$ &&  $\W{2} \in \mathbb{R}^{E \times n\times d}$,   weight of down projection \\ \midrule
  
  $\pi$ &&  $\pi \in \{0, 1\}^{T \times E}$, a binary-valued matrix where $\pi_{t, e}$ represents if token $t$ is routed to expert $e$   \\ \midrule
  
  $S$ &&  $S \in \mathbb{R}^{T \times E}$,  router scores. In practice, we only materialize the sparsified $S$ instead of the full $S$  \\ \midrule
  
  $H$ &&   $H \in \mathbb{R}^{TK\times 2n}$, output of up projection\\ \midrule
  
  $A$ &&   $A \in \mathbb{R}^{TK\times n}$,  output of SwiGLU\\ \midrule
  
  $Y$ &&   $Y \in \mathbb{R}^{TK\times d}$,  output of down projection\\ \midrule

  $O$ &&   $O\in \mathbb{R}^{T\times d}$,  output of expert aggregation, also output of the entire MoE layer\\ \midrule

  $dO$ &&   $dO \in \mathbb{R}^{T\times d}$, activation gradient for $O$ \\ \midrule

  $dA'$ &&   $dA' = dO \ \W{2}^\top \in \mathbb{R}^{T\times n}$,  GEMM output of $dO$ and $\W{2}$. Intermediate result for computing $dA$ and $dS$\\ \midrule

  $dA$ &&   $dA = \mathrm{Broadcast}(\vs) \ dA' \in \mathbb{R}^{T\times n}$,  activation gradient for $A$\\ \midrule

  $dY$ &&   $dY = \mathrm{Broadcast}(\vs) \ dO \in \mathbb{R}^{TK\times d}$,  activation gradient for $Y$. $dY$ is not used in \Algname. \\ \midrule

  $dS$ &&   $dS \in \mathbb{R}^{T\times E}$,  activation gradient for $S$\\ \midrule

  $A'$ &&   $A' = \mathrm{Broadcast}(\vs) \ A \in \mathbb{R}^{T\times n}$,  intermediate result and input for computing $d\W{2}$\\ \midrule

  $dH$ &&   $dH \in \mathbb{R}^{T\times 2n}$,  activation gradient for $H$\\ \midrule

  $d\tilde{X}$ &&   $d\tilde{X} \in \mathbb{R}^{TK\times d}$,  activation gradient for $X$ before aggregation \\ \midrule

  $dX$ &&   $dX \in \mathbb{R}^{T\times d}$,  activation gradient for $X$ after aggregation \\ \midrule

  $d\W{1}$ &&   $d\W{1} \in \mathbb{R}^{E \times d\times 2n}$,  weight gradient for $\W{1}$ \\ \midrule

  $d\W{2}$ &&   $d\W{2} \in \mathbb{R}^{E \times n\times d}$,  weight gradient for $\W{2}$ \\ \midrule

  $A$ kernel &&  forward up-proj kernel \\ \midrule

  $Y$ kernel &&  forward down-proj kernel \\ \midrule

  $O$ kernel &&  forward expert aggregation kernel where each token aggregates all routed expert's result as the final forward output \\ \midrule

  $dH$ kernel &&  backward down-proj activation gradient kernel \\ \midrule

  $d\W{2}$ kernel &&  backward down-proj weight gradient kernel \\ \midrule

  $d\tilde{X}$ kernel &&  backward up-proj activation gradient kernel \\ \midrule

  $d\W{1}$ kernel &&  backward up-proj weight gradient kernel \\ \midrule
  
  $dX$ kernel &&  backward expert aggregation kernel where each token aggregates the routed expert's $d\tilde{X}$ \\ \midrule

 $\roundup{f_e}, \rounddown{f_e}$ && $\tileM$-rounded multiples of expert frequency $f_e$.  $\roundup{f_e} = \lceil f_e / \tileM \rceil \cdot \tileM \quad \rounddown{f_e} = \lfloor f_e/\tileM \rfloor \cdot \tileM$ \\ \midrule

 $\round{S}, \round{f_e}$ && $\round{f_e}$ is $\tileM$-rounded multiples of expert frequency $f_e$.  $\round{f_e} \in \{ \roundup{f_e}, \rounddown{f_e} \}$, and $\round{S}$ is the score after rounding in Algorithm~\ref{alg:token_rounding}.
  \\

    \bottomrule
    \end{tabular}
    \end{center}
\end{table*}

\clearpage

\clearpage

\section{\Algname's comparison with existing MoE kernel design} \label{sec:appendix:kernel_comparison}

Existing efficient MoE kernels also frame MoE computation as a Grouped GEMM, but their ingredients are different from \Algname. Here we provide an overview (but not a complete list) of key differences:

\begin{itemize}
    \item \textbf{ScatterMoE} \citep{tan2024scattered}\footnote{\scriptsize \url{https://github.com/shawntan/scattermoe/blob/47b5e1502e5a10e82c8e5945d761b877849871e7/scattermoe/mlp.py\#L51}} implements gather fusion for varlen-$\mathbf{M}$ Grouped GEMM but not for varlen-$\mathbf{K}$ Grouped GEMM. ScatterMoE also does not overlap MMA computation with memory IO. Moreover, ScatterMoE is also built on older versions of Triton where TMA is not supported. ScatterMoE computes $dS$ as $dS = \langle dO, Y \rangle$ which requires caching $Y$. This results in large IO cost and activation memory requirement. Both ScatterMoE's forward and backward pass have limited fusion and hence ScatterMoE is much slower than~\Algname, especially for the backward pass.

    \item \textbf{MoMoE} \citep{costin2025momoe}\footnote{\scriptsize \url{https://github.com/tilde-research/MoMoE-impl/blob/d6e2d683185bfe4030265c3ca062564356faa61e/momoe/momoe.py\#L914}} also implements the gather fusion for varlen-$\mathbf{M}$ but not varlen-$\mathbf{K}$ Grouped GEMM similarly to ScatterMoE. Although fused with up-proj activation gradient, the $dS$ computation still utilizes $dS = \langle dO, Y \rangle$. Similar to ScatterMoE, MoMoE does not use TMA for IO. The scatter operation in MoMoE is (much) slower than~\Algname, as shown in Figure~\ref{fig:scatter-and-add-speed}.
    
    \item \textbf{MegaBlocks} \citep{gale2023megablocks} has multiple MoE implementations and we focus on $\mathrm{ParallelDroplessMLP}$\footnote{\scriptsize \url{https://github.com/databricks/megablocks/blob/78eea65fda01e638af36ae38853bc51efb04a4b4/megablocks/layers/dmoe.py\#L18}} which is built on top of block-sparse matrix multiplication\footnote{\scriptsize\url{https://github.com/databricks/megablocks/blob/78eea65fda01e638af36ae38853bc51efb04a4b4/megablocks/layers/mlp.py\#L308}}. $\mathrm{ParallelDroplessMLP}$ first gathers and pads the tokens and then launches block-sparse GEMM for up and down-proj. Then, it launches a scatter kernel before reducing across the expert results. These sparse matrix multiplications usually take a longer time than the highly-optimized Grouped GEMM, as shown in Figure~\ref{fig:moe_breakdown}, and the gather and scatter kernel have a total IO cost of $8TKd$ bytes which can be a bottleneck for fine-grained MoEs. We consider MegaBlocks's $\mathrm{ParallelDroplessMLP}$ as a block-sparse GEMM baseline in our benchmark and find that MoE implemented via Grouped GEMMs often have a higher training throughput than MoEs implemented via block-sparse GEMM. 

    \item \textbf{Megatron-LM} \citep{shoeybi2019megatron} also has multiple MoE implementations and we focus on $\mathrm{GroupedMLP}$\footnote{\scriptsize\url{https://github.com/NVIDIA/Megatron-LM/blob/610a75ef3a4a80c2ce2da436c19244e5362978d4/megatron/core/transformer/moe/experts.py\#L100}},  which uses Grouped GEMM\footnote{\scriptsize \url{https://github.com/fanshiqing/grouped_gemm/blob/main/csrc/grouped_gemm.cu}} from the CUTLASS library \citep{CUTLASS_gemm} with JIT epilogue fusion as the GEMM backend. Similar to DeepGEMM, $\mathrm{GroupedMLP}$ does not fuse gather with the prologue (it assumes contiguously-packed inputs). A recent memory-efficient patch\footnote{\scriptsize \url{https://github.com/NVIDIA/Megatron-LM/commit/2659630721ac87237c8cb772b1c2f1b34176f443}} fuses $S$ weighting with SwiGLU computation during forward, and during backward which allows the PyTorch autograd engine \citep{paszke2019pytorchimperativestylehighperformance} to follow a similar computational path as \Algname.

    Megatron-LM also implements $\mathrm{TEGroupedMLP}$\footnote{\scriptsize\url{https://github.com/NVIDIA/Megatron-LM/blob/610a75ef3a4a80c2ce2da436c19244e5362978d4/megatron/core/transformer/moe/experts.py\#L746}} which launches 4 CUDA streams to execute a list of GEMM (without contiguously-packed inputs, and without a persistent tile scheduler). In this case, each expert independently launches a new GEMM kernel leading to ``bubbles'' on the CUDA streams. This leads to underutilization of the GPU resources. We empirically find that $\mathrm{TEGroupedMLP}$ runs slower than $\mathrm{GroupedMLP}$ and so we use $\mathrm{GroupedMLP}$ across all benchmarks.

    \item \textbf{DeepGEMM} \citep{deepgemm} designs a Grouped GEMM kernel for contiguously-packed inputs. They also don't implement any other fusion for both SM90 (Hopper) and SM100 (Blackwell) BF16 Grouped GEMM. DeepGEMM specializes more on distributed training with expert parallelism \citep{deepep2025}, and it is common to launch a separate all2all kernel \citep{lepikhingshard} which is then followed by a contiguous Grouped GEMM. DeepGEMM SM90 BF16 kernel also assumes that each expert receives a multiple of $\tileM$ tokens as it does not implement the TMA tensor descriptor online update during the Grouped GEMM computation. DeepGEMM's BF16 GEMM on SM90 also does not employ Ping-Pong scheduling. DeepGEMM's varlen-$\mathbf{M}$ grouped GEMM usually uses tile shape (128, 256, 64) for both Hopper and Blackwell GPUs, while \Algname~often picks (256, 256, 64) with 2-CTA MMA for Blackwell GPUs. DeepGEMM does not use CLC persistent tile scheduler for grouped GEMM on Blackwell GPUs while \Algname~often adopts CLC persistent tile scheduler on Blackwell. 
\end{itemize}

Additionally, ScatterMoE and MoMoE are both implemented in Triton \citep{triton} for the ease of development at the expense of losing full programmability of the asynchronous compute and memory IO of Hopper and Blackwell GPUs \citep{nvidia2022_h100,nvidia2022_gb200}. For example, they cannot implement fine-grained control of asynchronous load and store during the GEMM's epilogue. They also cannot overlap MMA with heavy epilogue operations using Ping-Pong scheduling. It becomes increasingly important to overlap IO operations in epilogue when the GEMM computations are small in size (as in the case of fine-grained MoEs) to achieve high GPU utilization.

\section{Gradient computation} \label{sec:appendix:dH_proof}

\label{sec:appendix:proof}
For an expert $e$, let
\begin{equation}
    X_{e}\!\in\!\R^{T_e\times d},\quad
    W_{1,e}\!\in\!\R^{d\times 2n},\quad
    W_{2,e}\!\in\!\R^{n\times d}
\end{equation}

The forward activation computation is given by:
\begin{equation}
    H_e = X_{e} W_{1,e}\in\R^{T_e\times 2n},\qquad
    A_{e}=\mathrm{SwiGLU}(H_e)\in\R^{T_e\times n},\qquad
    Y_{e}=A_{e}W_{2,e}\in\R^{T_e\times d}.
\end{equation}

The token aggregation with scores $S=\{s_{t,e}\}$ is given by
\begin{equation}
    O_t=\sum_{e} s_{t,e}\,Y_{e,t},\qquad dO_t \in \mathbb{R}^{1\times d} \text{ as the gathered results from } dO.
\end{equation}

We know
\begin{align}
    d Y_{e,t} = s_{t,e}\ dO_t
    \quad\Longrightarrow\quad
    dY_{e} = \mathrm{Broadcast}(\vs_e)\ dO_{e}. \label{eq:dY}
\end{align}
Define the Grouped GEMM output as $dA'_{e} := dO_{e}\ W_{2,e}^{\top}\in\R^{T_e\times n}$.

Then from Equation~\ref{eq:dY}
\begin{equation}
    dA_{e} = dY_{e}\ W_{2,e}^{\top} = \mathrm{Broadcast}(\vs_e)\ dA'_{e}.
\end{equation}

The activation gradient for score $dS$ is\footnote{If we also consider expert biases, we will have  $dS_{e,t} = \langle dA'_{e,t},\ A_{e,t}\rangle + \langle dO_{e,t},\ \mathrm{Broadcast}(S_{e,t})\rangle$. The activation memory efficiency can still be preserved but we need to perform a separate computation for $\langle dO_{e,t},\ \mathrm{Broadcast}(S_{e,t})\rangle$ either via a separate kernel (\Algname's current choice) or fuse it with $dH$ Grouped GEMM mainloop. Please refer to the released code of~\Algname. }
\begin{equation}
\label{eq:ds-n-reduce}
    \boxed{dS_{t,e}=\left\langle dO_t,\ Y_{e,t}\right\rangle
      =\left\langle dO_t\ W_{2,e}^{\top},\ A_{e,t}\right\rangle
      =\left\langle dA'_{e,t},\ A_{e,t}\right\rangle }.
\end{equation}

In addition, we can derive $dH_e$ from $dA_e$ and $A_e$ (recomputed from $H_e$) as:

\begin{equation}
    dH_e=\mathrm{dSwiGLU}(dA_{e},\ H_e).
\end{equation}

Using Equation~\ref{eq:dY},
\begin{equation}
    dW_{2,e}=A_{e}^{\top}\ dY_{e}
        =A_{e}^{\top}\big(\mathrm{Broadcast}(\vs_e)\ dO_{e}\big)
        =\big(\underbrace{\mathrm{Broadcast}(\vs_e)\ A_{e}}_{A'_{e}}\big)^{\!\top} dO_{e}.
\end{equation}

\subsection{Computational choices for $dS$} \label{sec:appendix:dS_computation}
If we do not implement custom kernels and rely solely on PyTorch's autograd (AD) engine, we can add the expert weighting ($S$) either (1) before down-proj forward or (2) after down-proj forward. Both yield identical results for forward and backward, but the computation for $dS$ is different. For (1), we need to compute $\left\langle dA'_{e,t},\ A_{e,t}\right\rangle$ which is used by \Algname~and Megatron\footnote{\scriptsize \url{https://github.com/NVIDIA/Megatron-LM/blob/44af130cc4568d324860646996b6a5bfd6c5e3e6/megatron/core/transformer/moe/experts.py\#L284}}. MoMoE\footnote{\scriptsize \url{https://github.com/tilde-research/MoMoE-impl/blob/d6e2d683185bfe4030265c3ca062564356faa61e/momoe/momoe.py\#L702}}, ScatterMoE\footnote{\scriptsize \url{https://github.com/shawntan/scattermoe/blob/47b5e1502e5a10e82c8e5945d761b877849871e7/scattermoe/parallel_experts.py\#L79}}, and MegaBlocks\footnote{\scriptsize \url{https://github.com/databricks/megablocks/blob/78eea65fda01e638af36ae38853bc51efb04a4b4/megablocks/ops/binned_scatter.py\#L12}}\footnote{\scriptsize \url{https://github.com/databricks/megablocks/blob/78eea65fda01e638af36ae38853bc51efb04a4b4/megablocks/ops/binned_scatter.py\#L49}} compute $\left\langle dO_t,\ Y_{e,t}\right\rangle$ as required in (2).

Note that $dS$ can be computed as any of $dS_{t,e}=\left\langle dA'_{e,t},\ A_{e,t}\right\rangle = \left\langle dO_t,\ Y_{e,t}\right\rangle$, however computing it as $dS_{t,e}=\left\langle dA'_{e,t},\ A_{e,t}\right\rangle$ is a computationally and activation memory-efficient choice due to the following reasons:

\begin{itemize}
    \item \textbf{Additional HBM traffic ($0$ vs. $2TKd$ bytes)}: $\left\langle dA'_{e,t},\ A_{e,t}\right\rangle$ requires $dA'_{e,t}$ and $A_{e,t}$ are already computed during the $dH$ kernel, so we can avoid extra unnecessary loads.

    \item \textbf{Extra cached activation memory ($0$ vs. $2TKd$ bytes)}: One of the reasons why the cached activation memory for ScatterMoE, MoMoE and MegaBlocks fails to stay constant w.r.t. expert granularity is the required caching of $Y$ for computing $dS$.


    \item \textbf{Parallel reduction rounds ($\log_2(n)$ vs. $\log_2(d)$)}: $\left\langle dA'_{e,t},\ A_{e,t}\right\rangle$ reduces over $n$ while $\left\langle dO_t,\ Y_{e,t}\right\rangle$ reduces over $d$. This difference saves at least $\log_2(d/n)$ rounds of reduction.
\end{itemize}


\section{Efficient top-$K$ sorting kernel for MoE} \label{sec:appendix:top-k}

Existing MoE approaches such as ScatterMoE, MoMoE, and MegaBlocks use the PyTorch top-$K$ ($\mathrm{torch.topk}$) to compute the expert assignments for each token. We find that the PyTorch top-$K$ kernel can take approximately 40\% of the router's computation time. 
We implement an efficient top-$K$ kernel in \Algname~to reduce the overhead due to PyTorch top-$K$. 
Our top-$K$ kernel supports $E$ and $K$ when $E\leq4096$ and $K\leq16$ and is optimized for the case of a large number of tokens $T$. We also offer an optional softmax fusion on top-$K$ values within the top-$K$ kernel.  




\begin{wrapfigure}{r}{0.25\textwidth}
\vspace{-2em}
\begin{minipage}{0.25\textwidth}
    \centering
    \includegraphics[width=\linewidth]{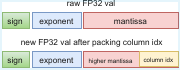}
    \caption{\small The sorting is conducted over values after we pack the column index bits into lower mantissa bits. This value format ensures a stable sorting result. Triton's official top-$K$ kernel follows a similar format.}
    \label{fig:col-bit-packing}
\end{minipage}
\vspace{-2em}
\end{wrapfigure}

The top-$K$ kernel accepts the router output with shape $(T, E)$ and parallelizes over $T$. The kernel uses bitonic sort \citep{bitonicsort} over every row (sorts $E$ values) and selects the first $K$ columns as the sort output. After loading the input, we pack the column indices of the first $K$ columns (for $\mathrm{argtopK}$) into the lower $\log_2(E)$ mantissa bits of FP32 values in registers, except that we specialize the base sorting cases (number of values $\leq 64$) to follow the comparison strategies obtained from optimal low-latency sorting networks \citep{sorting}, which provide the minimum number of parallel operation steps and required $\mathrm{compare\_and\_swap}$ calls. 



The bitonic compare and merging occurs within the same thread or the same warp via warp-shuffle. Therefore, every swap and merge operation only uses intra-thread or intra-warp registers. This achieves a higher memory bandwidth for the kernel over alternative kernel designs such as PyTorch TopK \citep{paszke2019pytorchimperativestylehighperformance}, the Triton \citep{triton} and Tilelang \citep{wang2025tilelang} official example, and RTop-K \citep{xie2025rtopk} in Figure~\ref{fig:topk-speed}.

Since the assigned column indices for values on each row are always unique, there will not be any equal numbers after we pack the column index to the lower mantissa bits. Therefore, \Algname's top-$K$ kernel is always stable as there will not be any tie-breaking scenarios during bitonic compare and merge.

\section{Referenced tables, figures, and \Algname~algorithms} \label{sec:appendix:more_kernel}

\begin{wrapfigure}{r}{0.35\textwidth}
\vspace{-3em}
\begin{minipage}{0.35\textwidth}

\begin{algorithm}[H]
    \caption{\Algname's MoE kernel backward pass of up-proj.}
    \DontPrintSemicolon
    \label{alg:up-bwd-moe}
     \fontsize{8pt}{9pt}\selectfont
    
    \SetKwInOut{Input}{Input}\SetKwInOut{Output}{Output}
    \Input{$\rX, \ \rpi, \ \rW_{1}, \ dH$}
    
    \Output{$dX, \ d\W{1}$.}
    
    \underline{\textcolor{red}{Up-proj act $d\tilde{X}$ kernel} $(\textcolor{blue}{dH, \ \W{1}}) \rightarrow \textcolor{blue}{d\tilde{X}}$:}  \;
    
    \note{// varlen-$\mathbf{M}$ Grouped GEMM}\;
    
    \Indp
    
    \textbf{Parallel} \For{$e \in [E]$}{
    
    $dH_{e}, \ \W{1,e} \gets \mathrm{load}(\textcolor{blue}{dH_{e}, \ \W{1,e}})$ \;
    
    $d\tilde{X}_e \gets dH_{e}\, \W{1,e}^\top$ \;
    
    $\textcolor{blue}{d\tilde{X}_e} \gets \mathrm{store}(d\tilde{X}_e)$
    
    }
    
    \Indm
    \underline{\textcolor{red}{Up-proj weight $d\W{1}$ kernel} $(\textcolor{blue}{X, \ dH, \ \pi}) \rightarrow \textcolor{blue}{d\W{1}}$}:\;
    
    \note{// Gather + varlen-$\mathbf{K}$ Grouped GEMM}\;
    
    \Indp

    \textbf{Parallel} \For{$e \in [E]$}{
    
    $X, \ \pi_{:,e}, \ dH_{e} \gets \mathrm{load}(\textcolor{blue}{X, \ \pi_{:,e}, \ dH_{e}})$ \;
    
    $\bX_e \gets \mathrm{Gather}(\rX, \ \bpi_{:, e})$\;
    
    $d\W{1,e} \gets X_{e}^{\top} dH_e$ \; 
    
    $\textcolor{blue}{d\W{1,e}} \gets \mathrm{store}(d\W{1,e})$
    }
    
    \Indm
    \underline{\textcolor{red}{Expert aggregation $dX$ kernel} $(\textcolor{blue}{d\tilde{X}, \ \pi}) \rightarrow \textcolor{blue}{dX}$}: \;
    
    \note{// Gather and sum}\;
    
    \Indp
    
    \textbf{Parallel} \For{$t \in [T]$}{
    { 
    
    $d\tilde{X}, \ \pi_{t,e} \gets \mathrm{load}(\textcolor{blue}{d\tilde{X}, \ \pi_{t,e}})$ \;
    
    $dX_t \gets \sum_{e \in [E]} \pi_{t, e} d\tilde{X}_{e,t}$}\; 
    
    $\textcolor{blue}{dX_t} \gets \mathrm{store}(dX_t)$ \;
    }    

\end{algorithm}

\end{minipage}
\end{wrapfigure}

\begin{figure}
    \vspace{0.5em}
    \centering
    \includegraphics[width=0.7\textwidth]{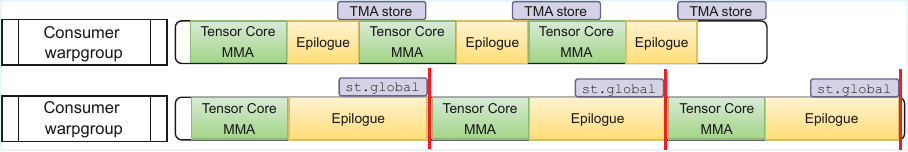}     
    \caption{\small Illustration to show asynchronous TMA store (top) has higher memory bandwidth and can naturally overlap with TensorCore MMA while synchronous $\mathrm{st.global}$ (bottom) PTX instruction, necessary for scatter fusion on Hopper GPUs, blocks the execution of next Tensor Core MMA tile and leads to longer kernel runtime. This figure is supported by the 20.1\% speedup on average of "\Algname~(gemm + gth w. sum)" (TMA store) over "\Algname~(gemm w. sct + sum)" ($\mathrm{st.global}$) in Figure~\ref{fig:scatter-and-add-speed}'s transparent bars. As a result, \Algname~does not fuse scatter with HBM store and instead, lets each token gather the expert results in the expert aggregation kernel. Both ScatterMoE and MoMoE do not adopt such design and \Algname~can achieve 1.75x and 3.11x speedup respectively on average during forward down-proj kernel in Figure~\ref{fig:moe_breakdown}.}
    \label{fig:expert-agg-why-slow}  
\end{figure}

\begin{figure}[!ht]
    \centering
    \includegraphics[width=\textwidth]{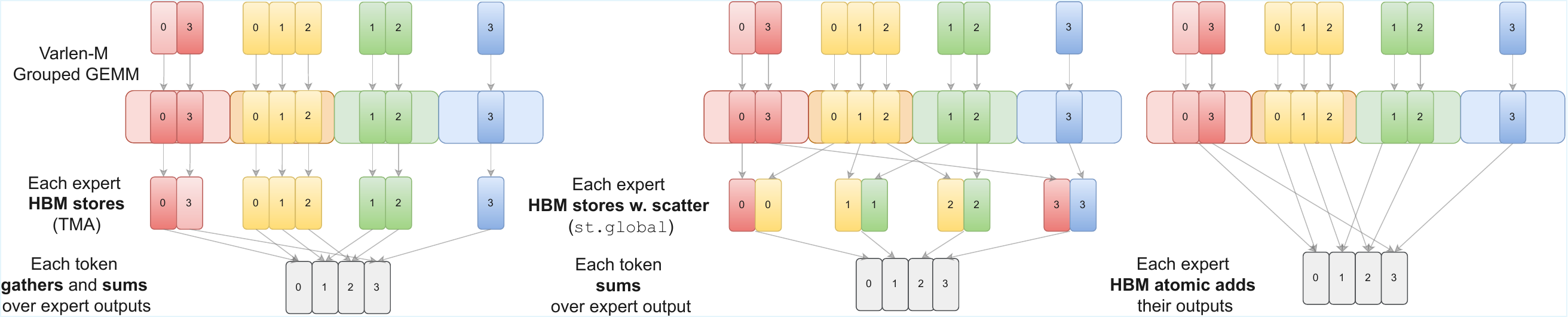}     
    \caption{\small Possible strategies for storing the results and aggregating the results for each token. \Algname~chooses the first strategy (left) in which each expert directly stores contiguously-packed outputs via TMA in the GEMM epilogue. In the expert aggregation kernel, each token gathers and sums over activated expert outputs. ScatterMoE and MoMoE (middle) choose to fuse HBM store with scatter in epilogue and launch a summation kernel afterwards. We note that each token gathering (left) the Grouped GEMM result is equivalent to each expert scattering (middle) the Grouped GEMM outputs. In Figure~\ref{fig:scatter-and-add-speed}, we implement both strategies on \Algname~and observe the left strategy can have 17\% speedup over the middle strategy. It is also possible to fuse atomic add in the epilogue to circumvent the requirement of an expert aggregation kernel as the right subfigure illustrated. However, this atomic add operation creates new issues like non-determinism \citep{determinism} and numerical accuracy (for BF16 atomic add). This figure is adapted from \citet{tan2024scattered}'s Figure 2.}
    \label{fig:expert-agg}  
\end{figure}

\begin{figure}[!ht]
    \includegraphics[width=\linewidth]{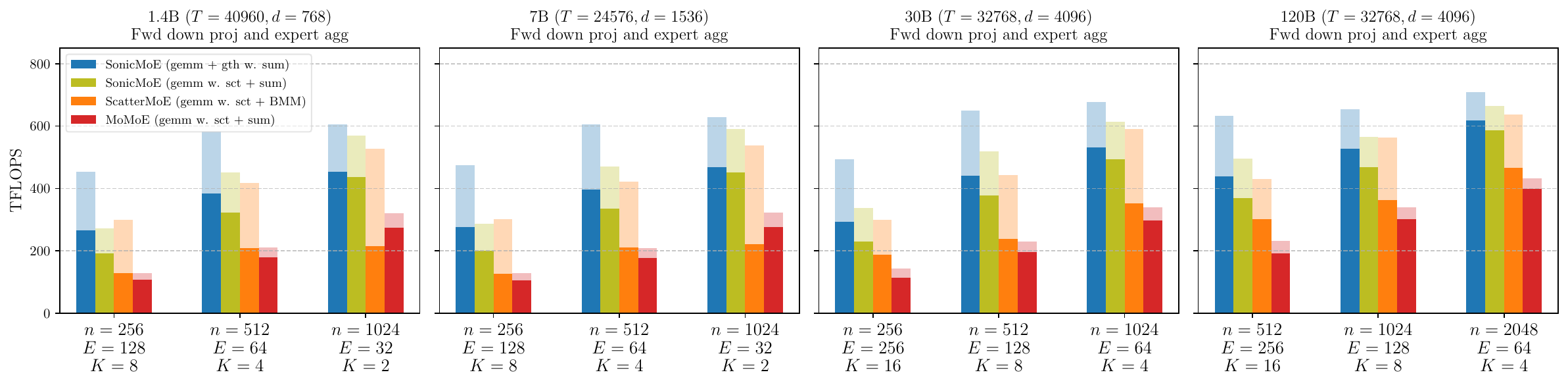}
    \caption{\small Throughput of Grouped GEMM and expert aggregation kernel on H100 GPUs. ``\Algname~(gemm + gth w. sum)'' is the final design choice for~\Algname~as illustrated in Figure~\ref{fig:expert-agg} left strategy. We compare this design against ``\Algname~(gemm w. sct + sum)'' that implements the Figure~\ref{fig:expert-agg} middle strategy on \Algname. We use identical tile sizes and other GEMM configs for both ``\Algname~(gemm + gth w. sum)'' and ``\Algname~(gemm w. sct + sum)''. We also compare with ScatterMoE's design (fused scatter with GEMM + $\mathrm{torch.bmm}$, labeled as ``ScatterMoE (gemm w. sct + BMM)'') and MoMoE's design (fused scatter with GEMM + $\mathrm{torch.sum}$, labeled as ``MoMoE (gemm w. sct + sum)''). For each method, we report the GEMM TFLOPS in transparent bars and TFLOPS of total runtime of GEMM and expert aggregation in the opaque bars.}
    \label{fig:scatter-and-add-speed}  
\end{figure}

In Table~\ref{tab:moe-scaling}, we provide a trending overview for open-source frontier MoE models. We present \Algname's expert aggregation strategy in Figure~\ref{fig:expert-agg}. Figure~\ref{fig:expert-agg-why-slow} illustrates that on Hopper GPUs, asynchronous TMA store (top) has higher memory bandwidth and can naturally overlap with TensorCore MMA whereas synchronous $\mathrm{st.global}$ (bottom) PTX instruction, necessary for scatter fusion on Hopper GPUs, blocks the execution of next Tensor Core MMA tile and leads to longer kernel runtime.


\begin{table}[!ht]
\centering
\caption{\small \textbf{MoE Scaling Trends:} Here, we show the activation ratio as experts activated per token $K$ / total experts $E$ and expert granularity is shown as model embedding dimension ($d$) / expert intermediate size ($n$) for frontier open source models. We do not include the shared experts for the MoE sparsity calculation. The trend indicates new open-source MoE models tend to be more granular and sparser. }

\label{tab:moe-scaling}

\setlength{\tabcolsep}{4pt}
\renewcommand{\arraystretch}{1.1}
\fontsize{8pt}{10pt}\selectfont
\begin{tabular}{lcccc}
\toprule
\textbf{Model} & \textbf{Release date} & \textbf{Parameters} & \textbf{Expert activation ratio ($K/E$)} & \textbf{Expert granularity ($d/n$)} \\
\midrule
Mixtral 8x22B \citep{mistral2024mixtral}&11/23 & 131B & 25.0\% (2/8) & $6144 / 16384 = 0.38$ \\

DBRX \citep{dbrx2024}&03/24 & 132B & 25.0\% (4/16) & $6144 / 10752 = 0.57$ \\

Phi-3.5-MoE \citep{azure2024phi35moe}& 09/24 & 42B & 12.5\% (2/16) & $4096 / 6400 = 0.64$ \\

OLMoE \citep{muennighoff2024olmoe} & 09/24 & 7B & 12.5\% (8/64) & $2048 / 1024 = 2.00$ \\

Granite 3.1-MoE \citep{granite3.1_language_models}& 12/24 & 3B & 20.0\% (8/40) & $1536 / 512 = 3.00$ \\
\textbf{DeepSeek-V3} \citep{deepseekai2025deepseekv3technicalreport}& \textbf{12/24} & \textbf{671B} & \textbf{3.13\% (8/256)} & \textbf{$7168 / 2048 = 3.50$} \\

\textbf{Qwen3 MoE} \citep{qwen3_2024} & \textbf{04/25} & \textbf{235B} & \textbf{6.25\% (8/128)} & \textbf{$4096 / 1536 = 2.67$} \\

\textbf{QWen3-30B-A3B} \citep{qwen3technicalreport} & \textbf{05/25} & \textbf{30.5B} & \textbf{6.25\% (8/128)} & \textbf{$2048 / 768 = 2.67$} \\

\textbf{Kimi K2} \citep{kimiteam2025kimik2openagentic} & \textbf{07/25} & \textbf{1.04T} & \textbf{2.08\% (8/384)} & \textbf{$7168 / 2048 = 3.50$} \\

gpt-oss-120b \citep{openai2025gptoss120bgptoss20bmodel} & 08/25 & 120B & 3.13\% (4/128) & $2880 / 2880 = 1.00$ \\

\textbf{GLM-4.5-Air} \citep{zeng2025glm}& \textbf{08/25} & \textbf{106B} & \textbf{6.25\% (8/128)} & \textbf{$4096 / 1408 = 2.91$} \\

\textbf{Qwen3-Next-80B-A3B-Instruct} \citep{qwen3technicalreport} & \textbf{09/25} & \textbf{81B} & \textbf{1.95\% (10/512)} & \textbf{$2048/512=4.00$} \\

\textbf{DeepSeek-V3.2-Exp} \citep{deepseekai2024deepseekv32} & \textbf{10/25} & \textbf{685B} & \textbf{3.13\% (8/256)} & \textbf{$7168/2048=3.50$} \\

\bottomrule
\end{tabular}
\end{table}
\raggedbottom


\clearpage
\section{\Algname's ablation study on kernel-level throughput} \label{sec:appendix:kernel_ablation}
In this section, we present kernel-level ablation studies on training throughput to examine the impact of each of the implemented features on~\Algname. In Section~\ref{sec:appendix:grouped_gemm}, we investigate the Grouped GEMM throughput with and without the gather fusion on Hopper and Blackwell GPUs. In Section~\ref{sec:appendix:expert_agg}, we profile the memory bandwidth of expert aggregation kernels. In section~\ref{sec:appendix:scatter_and_add}, we compare the left and middle expert aggregation strategy when both implemented on \Algname~on H100 GPUs in Figure~\ref{fig:expert-agg}. In Section~\ref{sec:appendix:topk}, we compare~\Algname's top-$K$ kernels with other efficient top-$K$ implementations.

\subsection{Grouped GEMM} \label{sec:appendix:grouped_gemm}

\begin{figure}[h]
    \centering
    \begin{subfigure}{\linewidth}
        \includegraphics[width=\linewidth]{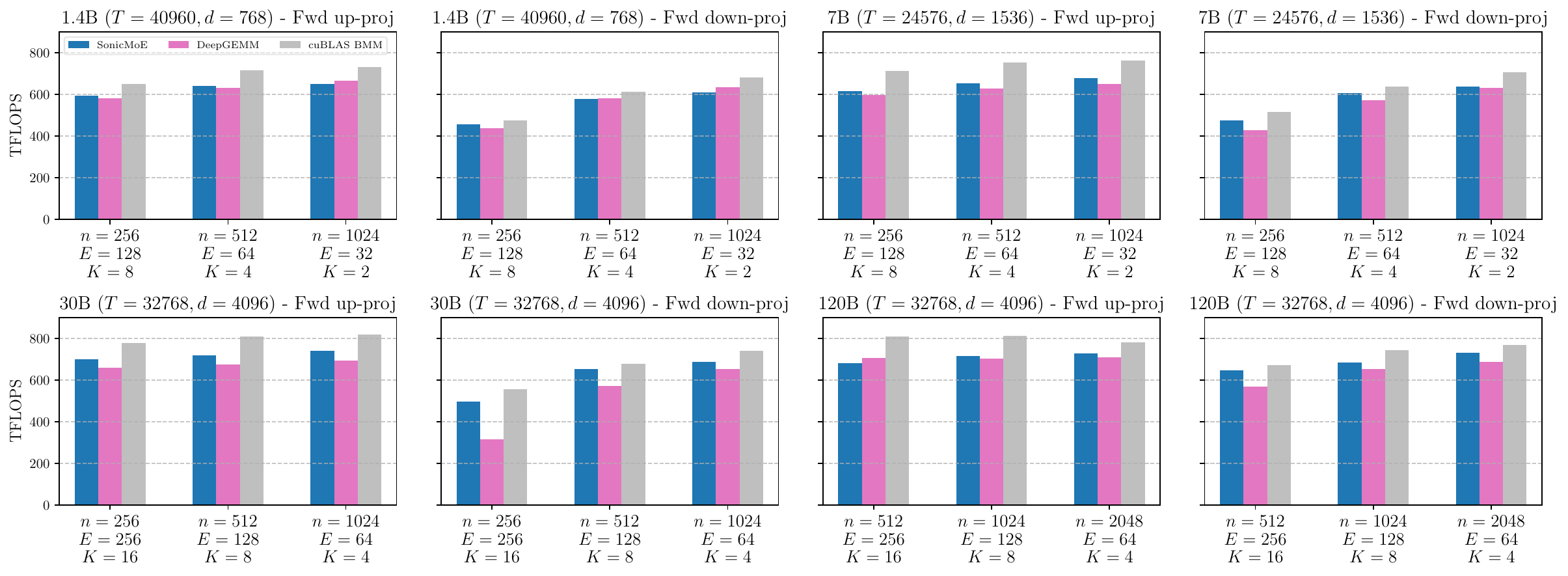}
        \caption{\small Varlen-$\mathbf{M}$ Grouped GEMM with contiguously-packed inputs to up and down-proj during forward pass on H100 GPUs. } 
        \label{fig:grouepd-gemm-contiguous-speed}
    \end{subfigure}
    \begin{subfigure}{\linewidth}
        \includegraphics[width=\linewidth]{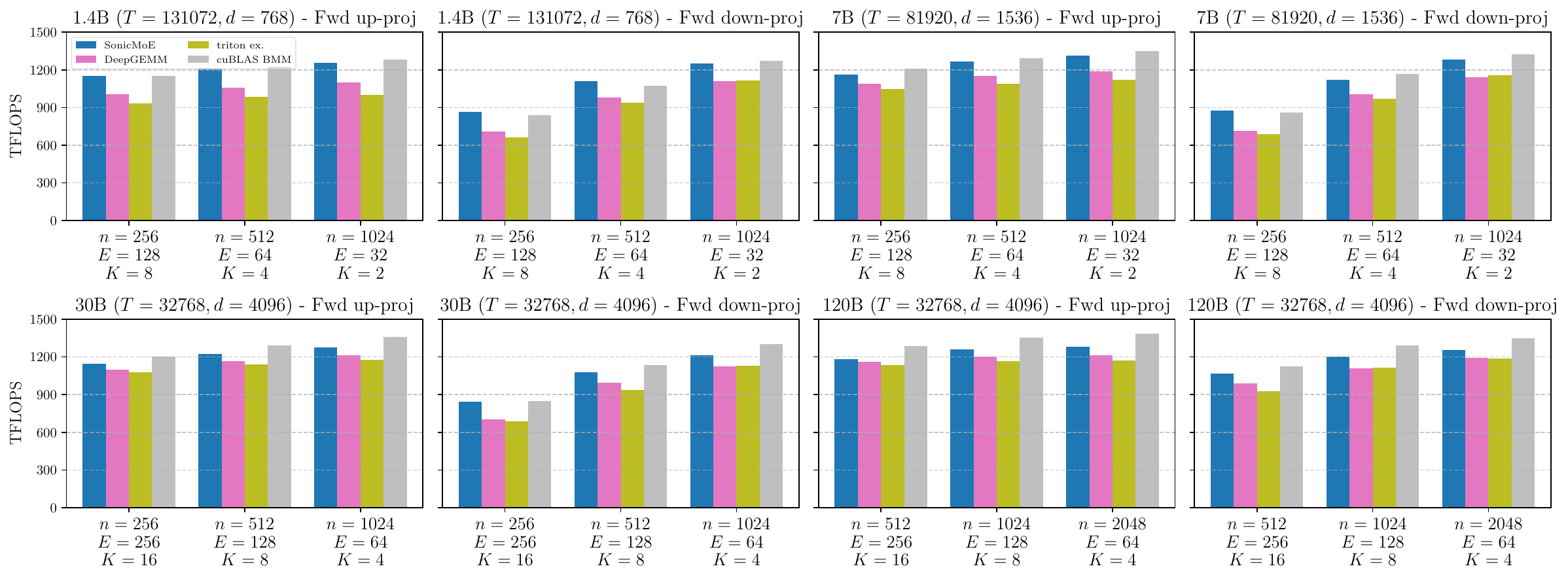}
        \caption{\small Varlen-$\mathbf{M}$ Grouped GEMM with contiguously-packed inputs to up and down-proj during forward pass on B300 GPUs. }
        \label{fig:grouepd-gemm-contiguous-speed-B300}
        \vspace{-1em}
    \end{subfigure}
    \caption{\small Varlen-$\mathbf{M}$ Grouped GEMM with contiguously-packed inputs on H100 and B300 GPUs. We use the same configurations as in Figure~\ref{fig:mem}. ``cuBLAS BMM'' is a \textit{dense} BMM baseline equivalent to all expert receiving equal number of tokens (perfectly load balanced), whose TFLOPS can be considered as an \textit{upper bound} for any Grouped GEMM kernel.}
\end{figure}

We also benchmark \Algname's base Grouped GEMM kernel with both contiguously-packed inputs and gathered inputs without any epilogue fusion on H100 and B300 GPUs. For contiguously-packed inputs, we mainly compare with DeepGEMM kernels ($\mathrm{sm90\_m\_grouped\_bf16\_gemm\_contiguous}$\footnote{\scriptsize\url{https://github.com/deepseek-ai/DeepGEMM/blob/9b680f428484625f4f35dc3617f134187c6bcd4a/csrc/jit_kernels/impls/sm90_bf16_gemm.hpp\#L127}} and $\mathrm{sm90\_bf16\_k\_grouped\_gemm}$\footnote{\scriptsize\url{https://github.com/deepseek-ai/DeepGEMM/blob/9b680f428484625f4f35dc3617f134187c6bcd4a/csrc/jit_kernels/impls/sm90_bf16_gemm.hpp\#L234}} on H100 GPUs, $\mathrm{sm100\_m\_grouped\_bf16\_gemm\_contiguous}$\footnote{\scriptsize\url{https://github.com/deepseek-ai/DeepGEMM/blob/35c4bc87713726d048f65275f6f1b551a4e7a6dc/csrc/jit_kernels/impls/sm100_bf16_gemm.hpp\#L124}} and $\mathrm{sm100\_bf16\_k\_grouped\_gemm}$\footnote{\scriptsize\url{https://github.com/deepseek-ai/DeepGEMM/blob/35c4bc87713726d048f65275f6f1b551a4e7a6dc/csrc/jit_kernels/impls/sm100_bf16_gemm.hpp\#L233}} on B300 GPUs). We note that at the time of writing, DeepGEMM's $\mathbf{k}$ Grouped GEMM kernel only accepts the form of $D = A B+C$ for GEMM\footnote{This formula naturally fits with the gradient accumulation.} where for correctness, we use a zero-filled FP32 $C$ weight gradient buffer as accumulator input, but we use $C$ as uninitialized weight gradient buffer during benchmarking. For inputs requiring a gather operation, we mainly compare with ScatterMoE and MoMoE. We also benchmark cuBLAS dense BMM (CUDA toolkit v12.9 for H100 GPUs and v13.0 for B300 GPUs) assuming each expert receives the same amount of tokens. On Blackwell GPUs, we also benchmark the GEMM example in triton official example\footnote{\scriptsize\url{https://github.com/triton-lang/triton/blob/c833d995f93d2eadb5627afae129496a852bc9fb/python/triton_kernels/bench/bench_mlp.py\#L53}}.

\paragraph{Grouped GEMM with contiguously-packed inputs.} We compare \Algname~with DeepGEMM on varlen-$\mathbf{M}$ Grouped GEMM without any other fusion. We also benchmark cuBLAS dense BMM (perfect load balance and no tensormap update needed) as a reference for the TFLOPS upper bound for Grouped GEMM.

\begin{itemize}
    \item \textbf{H100 GPUs}: In Figure~\ref{fig:grouepd-gemm-contiguous-speed}, we find that \Algname's up-proj has 2.7\% higher TFLOPS while down-proj has 10.0\% higher TFLOPS than DeepGEMM on average. We use Ping-Pong scheduling for down-proj with $n<1024$ while DeepGEMM~\citep{deepgemm} uses cooperative scheduling~\citep{deepgemm,pingpongandcooperative}, and as a result, we observe \Algname~has a greater speedup over DeepGEMM when intermediate size is small. For example, \Algname~has 57.4\%, 14.0\%, and 5.3\% more TFLOPS than DeepGEMM for 30B down-proj config. 
    
    \item \textbf{B300 GPUs}: In Figure~\ref{fig:grouepd-gemm-contiguous-speed-B300}, we find that \Algname's up-proj has 8.1\% higher TFLOPS while down-proj has 12.7\% higher TFLOPS than DeepGEMM on average. Compared to the triton official GEMM example, \Algname's up-proj has 13.3\% higher TFLOPS while down-proj has 15.6\% higher TFLOPS on average. \Algname's speedup for fine-grained MoEs are still maintained: for example, \Algname~has 20.2\%, 8.5\%, and 8.1\% higher TFLOPS than DeepGEMM, and 22.2\%, 14.9\%, and 7.5\% higher TFLOPS than triton official example for 30B down-proj config. 
\end{itemize}


\begin{figure}[!ht]
    \centering
    \begin{subfigure}{\linewidth}
        \includegraphics[width=\linewidth]{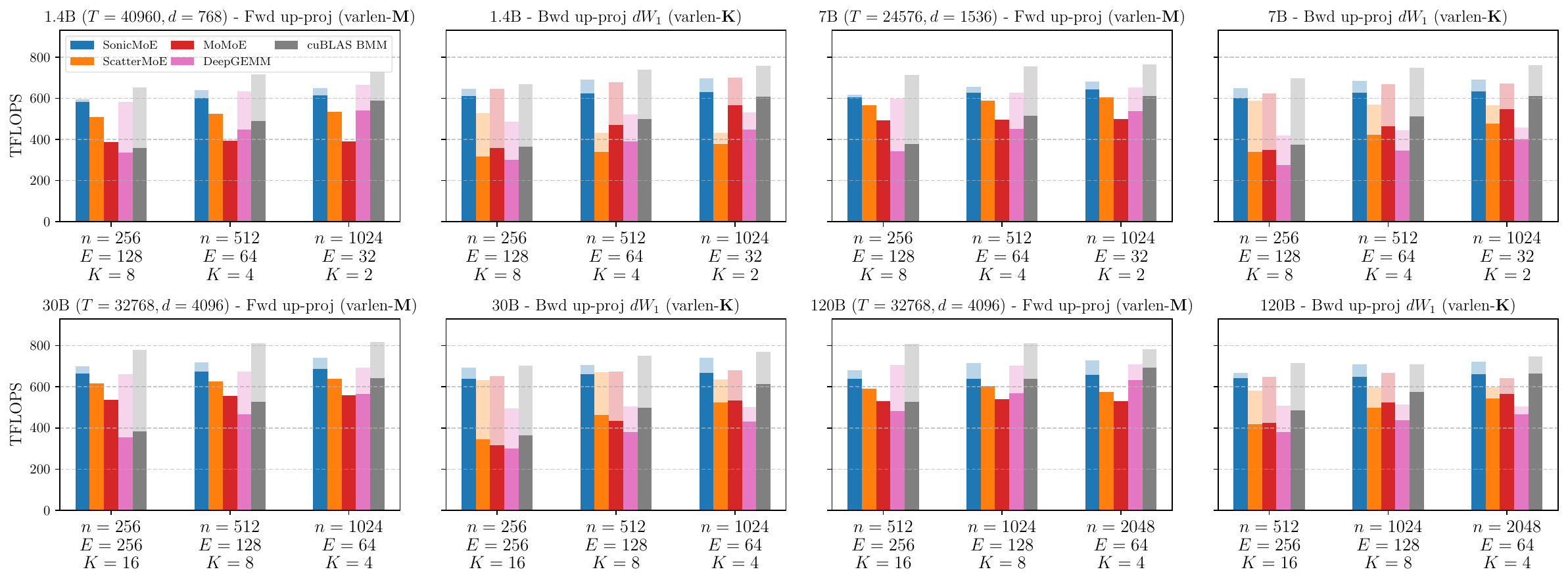}
        \caption{\small Forward pass up-proj (gather on $\mathbf{M}$ dim) and backward up-proj weight gradient $d\W{1}$ (gather on $\mathbf{K}$ dim) kernel on H100 GPUs. }
        \label{fig:gather-speed}
    \end{subfigure}
    \begin{subfigure}{\linewidth}
        \includegraphics[width=\linewidth]{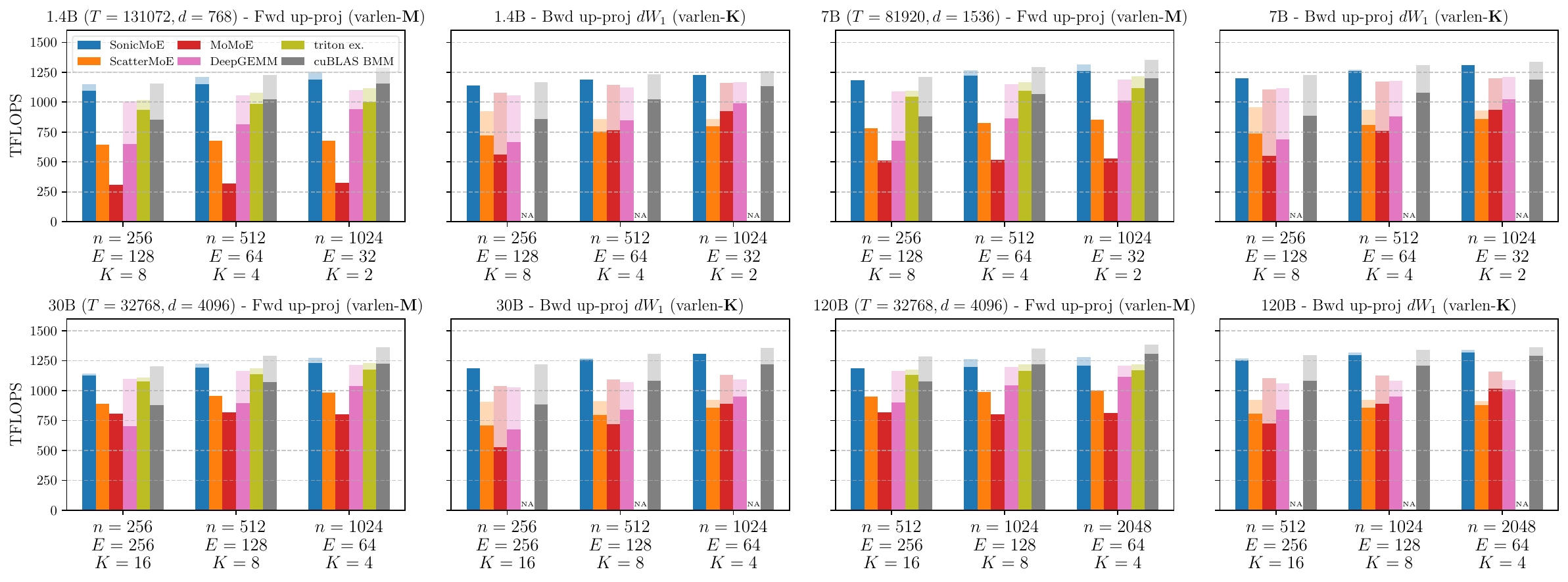}
        \caption{\small Forward pass up-proj (gather on $\mathbf{M}$ dim) and backward up-proj weight gradient $d\W{1}$ (gather on $\mathbf{K}$ dim) kernel on B300 GPUs.}
        \label{fig:gather-speed-B300}
    \end{subfigure}
    \caption{\small Forward pass up-proj (gather on $\mathbf{M}$ dim) and backward up-proj weight gradient $d\W{1}$ (gather on $\mathbf{K}$ dim) kernel on H100 and B300 GPUs. \Algname~supports both inputs gathered from different positions (opaque bars) and contiguously-packed inputs (transparent bars). ScatterMoE and MoMoE both have gather fusion for varlen-$\mathbf{M}$ but not varlen-$\mathbf{K}$, so we benchmark their gather with varlen-$\mathbf{K}$ Grouped GEMM time (opaque bar) by adding up the time of their contiguously-packed weight gradient kernel (transparent bar) with the time of their gather kernel. DeepGEMM does not have gather fusion for both varlen-$\mathbf{M}$ and varlen-$\mathbf{K}$ Grouped GEMM, so we provide an optimized gather kernel in both cases. We also provide a ``cuBLAS dense BMM" (transparent bar) baseline and the gather with GEMM time (opaque bar) by adding up the time with a heavily-tuned gather kernel's time with the same input shape, which can be considered as the upper bound TFLOPS for \emph{any} Grouped GEMM kernel without gather fusion.}
\end{figure}

\paragraph{Grouped GEMM with gather fusion.}

We report~\Algname, ScatterMoE, MoMoE, and DeepGEMM with and without gather fusion (as opaque and transparent bars for each method) on Hopper GPU. ScatterMoE and MoMoE both have gather fusion for varlen-$\mathbf{M}$ but not varlen-$\mathbf{K}$ Grouped GEMM, so we benchmark their gather with varlen-$\mathbf{K}$ Grouped GEMM time (opaque bar) by adding up the time of their contiguously-packed weight gradient kernel (transparent bar) with the time of their own gather kernel. We also adopt the similar benchmark approach for DeepGEMM and cuBLAS dense BMM. We note that the cuBLAS dense BMM results can be considered as the upper bound of the TFLOPS for any Grouped GEMM kernel without gather fusion. 

\begin{itemize}
    \item \textbf{Gather on $\mathbf{M}$ dimension.} 

    \begin{itemize}
        \item \textbf{H100 GPUs}: in Figure~\ref{fig:gather-speed}, the average relative TFLOPS difference of \Algname~with and without gather fusion on $\mathbf{M}$ dim is 6.3\%. \Algname~consistently achieves higher TFLOPS than ScatterMoE (avg 9.7\%), MoMoE (avg 30.9\%), and DeepGEMM (avg 38.3\%) with gather fusion.

        \item \textbf{B300 GPUs}: in Figure~\ref{fig:gather-speed-B300}, the average relative TFLOPS difference of \Algname~with and without gather fusion on $\mathbf{M}$ dim is 3.4\%. \textbf{Triton official example also implements gather fusion with TMA on Blackwell GPUs, but the difference of TFLOPS with and without gather fusion is 6.3\%, higher than \Algname.} Similar to the case on Hopper GPUs, \Algname~consistently achieves higher TFLOPS than ScatterMoE, MoMoE, and DeepGEMM with gather fusion.
    \end{itemize}

    \item \textbf{Gather on $\mathbf{K}$ dimension.} 

        \begin{itemize}
        \item \textbf{H100 GPUs}: In Figure~\ref{fig:gather-speed}, the average relative TFLOPS difference of \Algname~with and without gather fusion on $\mathbf{K}$ dim is 8.5\%. When we compare \Algname~with gather fusion (opaque bars) against ScatterMoE and MoMoE with their gather kernel together, we observe a wider gap as expert granularity increases (from right to left on each 3 bar groups). 
        
        

        \item \textbf{B300 GPUs}:  In Figure~\ref{fig:gather-speed-B300}, the average relative TFLOPS difference of \Algname~with and without gather fusion on $\mathbf{K}$ dim is -0.1\%. This means gather fusion on $\mathbf{K}$ dim virtually does not induce any throughput degradation on TFLOPS. We still observe a wider gap of TFLOPS between \Algname~with gather fusion and ScatterMoE and MoMoE with a separate gather kernel as expert granularity increases. 
    \end{itemize}
        
\end{itemize}


\subsection{Expert aggregation} \label{sec:appendix:expert_agg}

\begin{figure}[!ht]
    \centering
    \begin{subfigure}{\linewidth}
        \includegraphics[width=\linewidth]{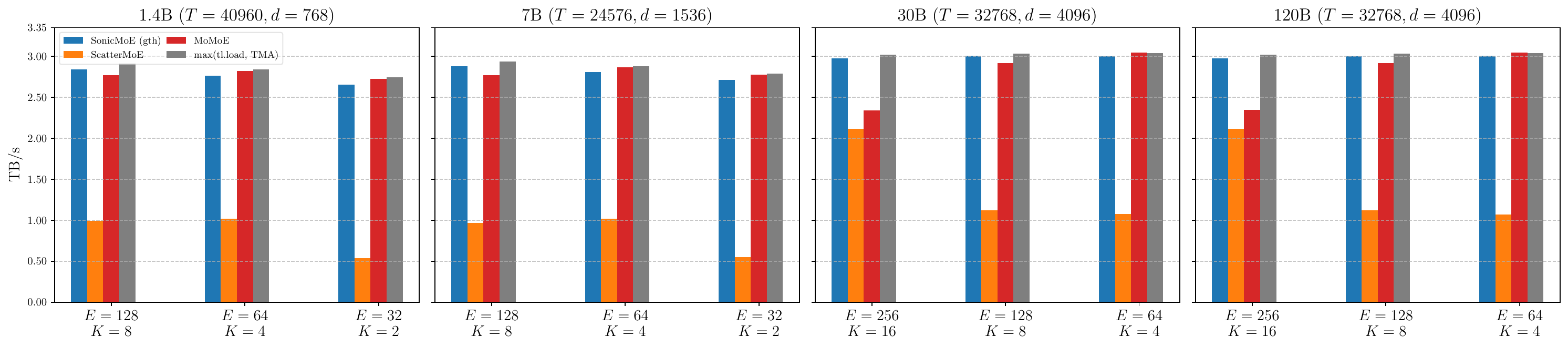}
        \caption{\small Expert aggregation kernels ($O$ kernel) during forward pass of MoE on H100 GPUs.}
        \label{fig:expert-agg-speed}
    \end{subfigure}
    \begin{subfigure}{\linewidth}
        \includegraphics[width=\linewidth]{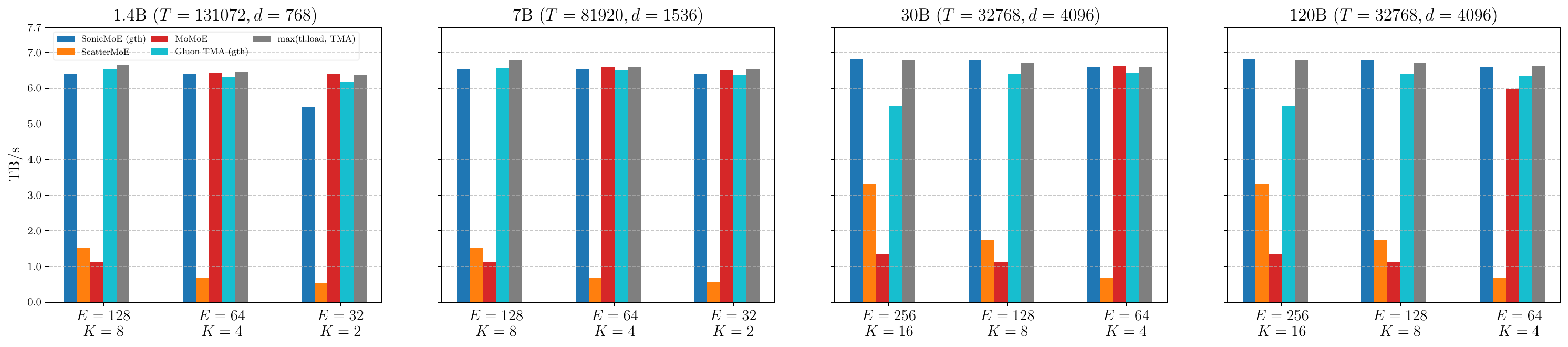}
        \caption{\small Expert aggregation kernels ($O$ kernel) during forward pass of MoE on B300 GPUs. }
        \label{fig:expert-agg-speed-B300}
    \end{subfigure}
    \caption{\small Expert aggregation kernel ($O$ kernel) on H100 and B300 GPUs. Same configurations as in Figure~\ref{fig:mem}. ScatterMoE uses $\mathrm{torch.bmm}$ call to reduce over $K$ for a contiguous $Y$ input, MoMoE uses a $\mathrm{torch.sum}$ call. We take the maximum bandwidth between PyTorch eager and PyTorch compile on default mode with PyTorch 2.9.0. We also implement 2 optimized triton kernels that either use vanilla triton $\mathrm{tl.load}$ or TMA load and they both sum over contiguous $Y$ inputs. We report the max bandwidth of both results as ``max(tl.load, TMA)''. On B300 GPUs, we also use Gluon to implement gather-and-sum aggregation with TMA gather. On H100 GPUs, we use Triton 3.3.1, and on B300 GPUs, we use Triton 3.6.0 for benchmarking. }
\end{figure}

In Figure~\ref{fig:expert-agg-speed} and~\ref{fig:expert-agg-speed-B300}, we compare the bandwidth of \Algname's gather-and-sum aggregation (Figure~\ref{fig:expert-agg} left) with ScatterMoE's $\mathrm{torch.bmm}$ and MoMoE $\mathrm{torch.sum}$ aggregation on contiguous $Y$ inputs (Figure~\ref{fig:expert-agg} middle). In addition, we implement 2 optimized triton kernels that either use vanilla triton $\mathrm{tl.load}$ or TMA load and they both sum over contiguous $Y$ inputs. We report the max bandwidth of both results as ``max(tl.load, TMA)''. This result achieves 85\%+ peak for most MoE configurations so we consider it as an upper bound for any aggregation kernel on both H100 and B300 GPUs. On B300 GPUs, we also use Gluon to implement gather-and-sum aggregation with TMA gather\footnote{adopted from \scriptsize\url{https://github.com/triton-lang/triton/blob/8ecbfb066a28e231e9fd54ca0ce6de8ace4950aa/python/tutorials/gluon/09-tma-gather-scatter.py\#L298}}.


\begin{itemize}
    \item \textbf{H100 GPUs}: although~\Algname's aggregation kernel requires a gather fusion during HBM load, the memory bandwidth of \Algname~still surpasses ScatterMoE (2.92x on average) and MoMoE (1.05x on average), and is only slightly slower (0.98x on average) than the triton upper bound of summing over contiguous $Y$.

    \item \textbf{B300 GPUs}: the results are similar to that of H100 GPUs. The memory bandwidth of \Algname~aggregation kernel still largely surpasses ScatterMoE (6.72x on average) and MoMoE (3.32x on average), and is only slightly slower (0.98x on average) than the triton upper bound of summing over contiguous $Y$. \textbf{We also note that \Algname~gather-and-sum is faster than Gluon TMA gather-and-sum (1.05x on average).}
\end{itemize}

\subsection{Strategies of combining grouped gemm and expert aggregation} \label{sec:appendix:scatter_and_add}

We compare the left and middle expert aggregation strategy when both are implemented on \Algname~on H100 GPUs in Figure~\ref{fig:expert-agg}. We observe the left strategy (gemm + gth w. sum) achieves a 20\% higher \tflops~than the middle strategy (gemm w. sct + sum) and we therefore choose the left one for forward down-proj and backward up-proj activation gradient kernel. 

\subsection{Top-$K$ sorting} \label{sec:appendix:topk}

\begin{figure}[!ht]
    \centering
    \begin{subfigure}{\linewidth}
        \includegraphics[width=\linewidth]{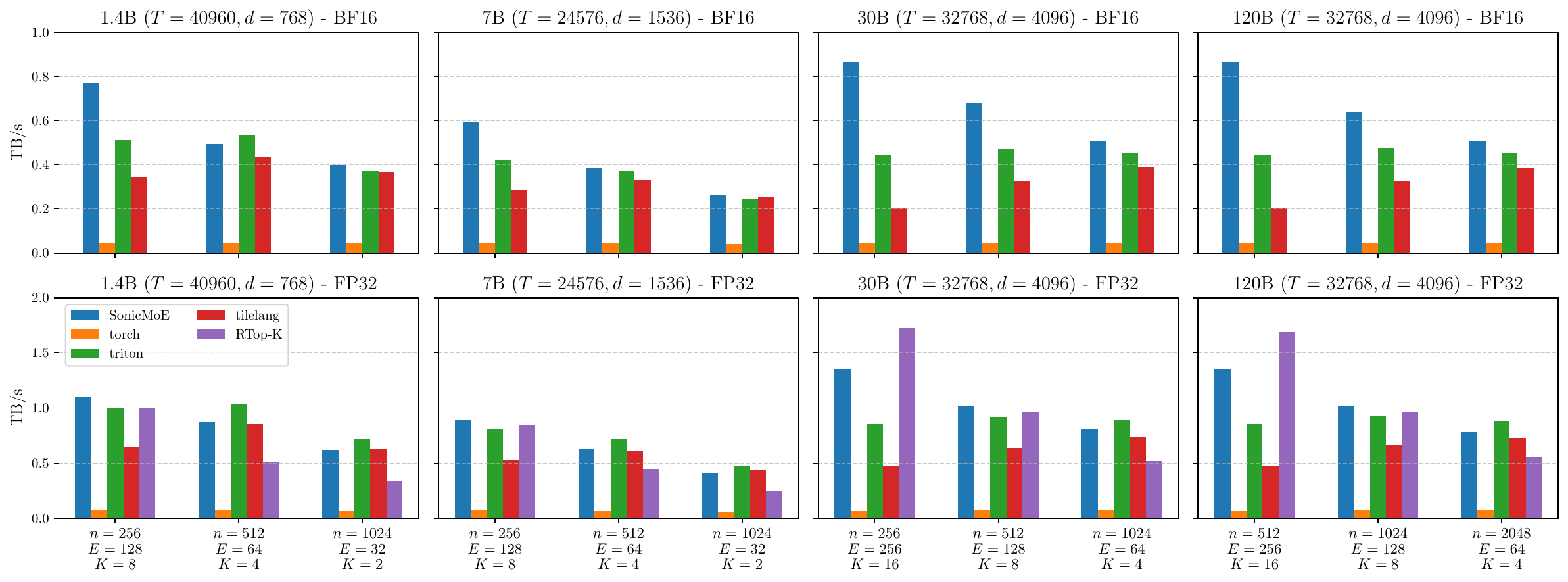}
        \caption{\small Top-$K$ kernels with BF16 inputs ($1^\text{st}$ row) and FP32 inputs ($2^\text{nd}$ row) during forward pass of MoE on H100 GPUs. }
    \label{fig:topk-speed}
    \end{subfigure}
    \begin{subfigure}{\linewidth}
        \includegraphics[width=\linewidth]{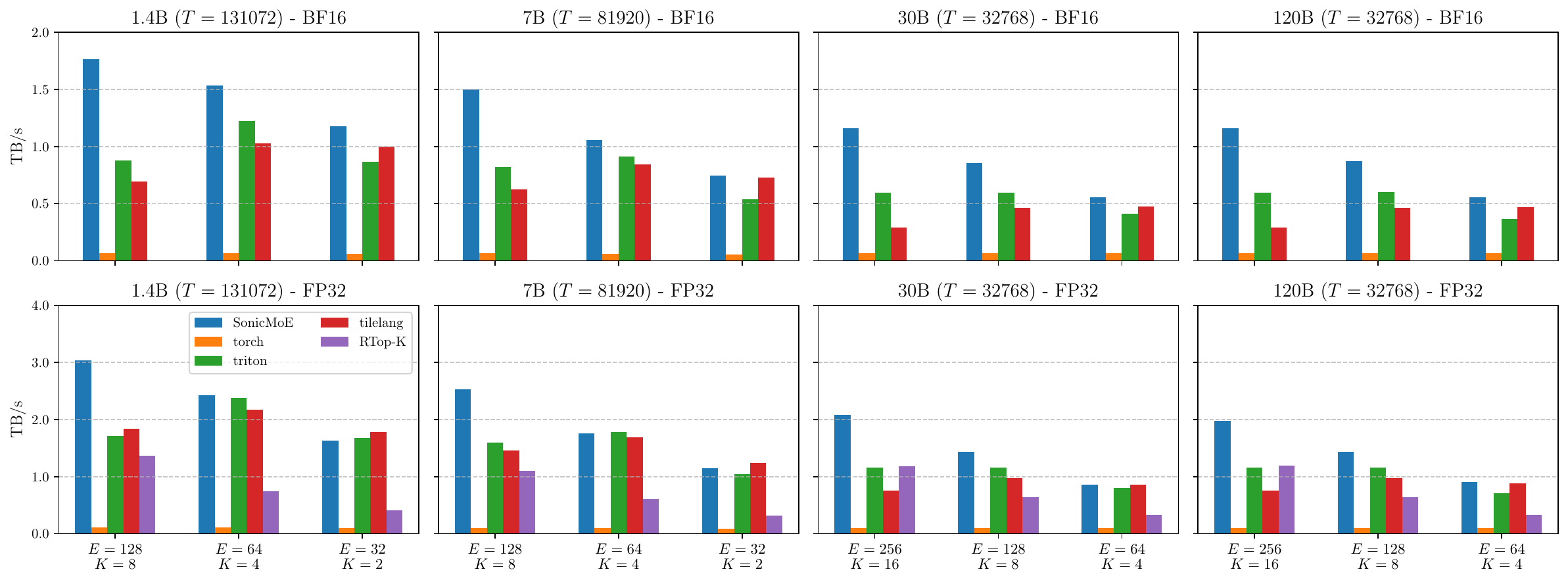}
        \caption{\small Top-$K$ kernels with BF16 inputs ($1^\text{st}$ row) and FP32 inputs ($2^\text{nd}$ row) during forward pass of MoE on B300 GPUs. }
    \label{fig:topk-speed-B300}
    \end{subfigure}
    \caption{\small Top-$K$ kernels during forward pass of MoE. Same configurations as in Figure~\ref{fig:mem}. ``torch'' is a direct $\mathrm{torch.topk}$ call. ``triton'' and ``tilelang'' are taken from their official examples with slight modifications to support BF16 inputs. For the triton official kernel, we remove the unnecessary bit matrix store and disable the softmax fusion in this example for a fair comparison. ``RTop-K''\citep{xie2025rtopk} only supports FP32 inputs. We set $\epsilon=0$ and maximum iteration as 8 for RTop-K.}
\end{figure}

We benchmark the bandwidth of \Algname's top-$K$ kernel in Figure~\ref{fig:topk-speed} and~\ref{fig:topk-speed-B300}. For each GPU, we compare \Algname~with PyTorch\footnote{\scriptsize \url{https://github.com/pytorch/pytorch/blob/b54e466fd04e5e736662a6206d81ab0d5fe85d91/aten/src/ATen/native/cuda/TensorTopK.cu\#L40}}, triton official example\footnote{\scriptsize \url{https://github.com/triton-lang/triton/blob/de8e71503fea971dfb65308147798657e18f8568/python/triton_kernels/triton_kernels/topk_details/_topk_forward.py\#L90}}, tilelang official example\footnote{\scriptsize \url{https://github.com/tile-ai/tilelang/blob/5c62d00a64f2f52cf6b2536a2492a29fc5323723/examples/topk/example_topk.py\#L18}}, and RTop-K\footnote{\scriptsize  \url{https://github.com/xiexi51/RTopK/blob/952f515321a4bdc5c4a57944bca0d32641052460/rtopk_kernel.cu\#L7}} \citep{xie2025rtopk} on BF16 and FP32 inputs.

\begin{itemize}
    \item \textbf{PyTorch single block Top-$K$}: PyTorch \citep{paszke2019pytorchimperativestylehighperformance} implements a radix-select followed by a gather algorithm for top-$K$, and it dispatches to a single or multiple block version depending on $T,E,K$. For large $T$ with modest $E$ and $K$, PyTorch uses the single-block version that performs 2 SMEM scans. In this case, \Algname's sorting networks with pure register-based communication are much faster.

    \item \textbf{Triton official example}: triton \citep{triton} provides a top-$K$ example kernel that is also based on bit packing and bitonic merge. The main algorithmic difference is that \Algname~relies on optimal sorting networks on the base cases while the Triton implementation directly calls $\mathrm{triton.language.topk}$. During the top-$K$ benchmark in Figure~\ref{fig:topk-speed}, we observe that Triton is much faster than PyTorch $\mathrm{torch.topk}$ but it is still consistently slower than \Algname's top-$K$ across all configurations.

    \item \textbf{Tilelang official example}: tilelang \citep{wang2025tilelang} provides a top-$K$ example kernel that performs $K$-pass maximum reduction. This design is more targeted for small $K$ and we observe that as both $E$ and $K$ become larger, the Tilelang top-$K$ kernel's throughput decreases compared to other baselines (and~\Algname)'s increasing trend. Such a trend makes tilelang's example top-$K$ kernel ($K$-pass top-$K$ kernel) unsuitable for fine-grained MoEs.

    \item \textbf{RTop-K}: RTop-K \citep{xie2025rtopk} follows a threshold-based binary search. Each bisection step utilizes warp-level primitives and follows a \textit{selection-by-count} method instead of \Algname's sorting network. RTop-K is also an iterative algorithm with iterations dependent on the value range and vector size. In addition, RTop-K heavily utilizes SMEM for scanning while \Algname~solely relies on registers for its $\mathrm{compare\_and\_swap}$ subroutine. We find that \Algname's top-$K$ generally achieves higher throughput on both H100 and B300 GPUs.
\end{itemize}



\section{More experiments}\label{sec:appendix:more_experiments}
In this section, we investigate the qualitative improvement from fine-grained MoE in Section~\ref{sec:appendix:granularity}. We also investigate the effect of rounding subroutines $\mathrm{round\_and\_sparsify}$ in Algorithm~\ref{alg:token_rounding} and the effects from microbatch size $T$ and tile size $\tileM$ for token rounding in Section~\ref{sec:appendix:ablation-microbatch-tileM}.

\subsection{Effect of expert granularity} \label{sec:appendix:granularity}

Here we validate the effectiveness of adopting fine-grained MoE. We fix the MoE activation ratio $\rho=K/E$ for the 0.5B and 1.4B model and we proportionally scale up $K$ and $E$ while linearly decreasing $n$ from row 1 to row 3 in Table~\ref{tab:moe-granularity:a} and~\ref{tab:moe-granularity:b}. 

In general, we observe a better performance for $n=256$ than $n=1024$ which is also consistent with the MoE scaling trends mentioned in Table~\ref{tab:moe-scaling}. 
In Figure~\ref{fig:teasor} right subfigure, we find both \Algname~and cuBLAS can still sustain the throughput from $n=1024$ to $n=256$ under iso-FLOPs, but starting from $n=256$ FLOPs will drop linearly w.r.t. granularity. Therefore, we choose $n=256$ for all experiments in Table~\ref{tab:exp-sparse}.

\begin{table}[!htbp]
    \setlength{\tabcolsep}{3pt}
    \renewcommand{\arraystretch}{1.15}
    \fontsize{8pt}{10pt}\selectfont
    \centering
    \vspace{-0.6em}
    \caption{\small Evaluation of MoE w.r.t. granularity with iso-FLOPs ($nK$ is constant) and iso-params ($nE$ is constant) settings. ``PPL'' refers to the validation perplexity at the end of training. ``Avg'' is the mean accuracy across the 11 downstream tasks. The ``dense, iso-FLOPs'' refers to a dense model with $nK$ as the intermediate size, while the ``dense, iso-params'' refers to a dense model with $nE$ as the intermediate size.}
    \label{tab:exp-granularity}
    
    \begin{subtable}[!ht]{\textwidth}
   \caption{\textbf{0.5B params, 20B tokens, 8/64 activated}} \label{tab:moe-granularity:a}
    \begin{tabular*}{\textwidth}{@{\extracolsep{\fill}}l!{\vrule width 0.4pt}c!{\vrule width 0.4pt}ccccccccccc!{\vrule width 0.4pt}c}
    \toprule
    \textbf{$(E, \, K, \,n)$} & \textbf{PPL} & \textbf{Wino} & \textbf{SIQA} & \textbf{SciQ} & \textbf{PIQA} & \textbf{OBQA} & \textbf{HS} & \textbf{COPA} & \textbf{CSQA} & \textbf{BoolQ} & \textbf{ArcE} & \textbf{ArcC} & \textbf{Avg} \\
    \midrule\midrule
    16, 2, 1024 & 16.23 & \textbf{53.0} & 41.3 & \textbf{79.8} & 65.0 & 32.6 & 37.8 & \textbf{66.0} & 32.2 & 55.8 & 53.9 & \textbf{29.1} & 49.7 \\
    64, 8, 256 & \textbf{16.01} & 51.0 & 41.4 & 79.2 & \textbf{65.5} & 31.6 & \textbf{38.4} & \textbf{66.0} & 31.5 & 60.2 & 57.5 & 25.7 & 49.8 \\
    256, 32, 64 & 16.13 & 51.2 & \textbf{41.5} & 78.9 & 65.3 & \textbf{34.2} & \textbf{38.4} & 63.0 & \textbf{32.4} & \textbf{60.6} & \textbf{59.5} & 28.1 & \textbf{50.3} \\ 
    \hline \hline

    Dense, iso-FLOPs & 19.90 & 48.9 & 41.4 & 74.9 & 62.2 & 30.2 & 32.6 & 62.0 & 31.6 & 61.7 & 53.2 & 27.1 & 47.8 \\

    Dense, iso-params & 15.46 & 52.1 & 41.5 & 78.9 & 65.3 & 34.0 & 39.2 & 69.0 & 32.2 & 58.5 & 59.3 & 28.8 & 50.8 \\

    \bottomrule
    \end{tabular*}
    \end{subtable}

    \vspace{0.5em}
    \begin{subtable}[!ht]{\textwidth}
   \caption{\textbf{1.4B params, 50B tokens, 8/128 activated}}  \label{tab:moe-granularity:b}
    \begin{tabular*}{\textwidth}{@{\extracolsep{\fill}}l!{\vrule width 0.4pt}c!{\vrule width 0.4pt}ccccccccccc!{\vrule width 0.4pt}c}
    \toprule
    32, 2, 1024 & 13.38 & \textbf{52.2} & \textbf{41.7} & 81.7 & 69.2 & 33.6 & 44.3 & 64.0 & 33.5 & \textbf{61.1} & 60.9 & 29.8 &  52.0 \\
    128, 8, 256 & \textbf{13.32} & 51.8 & \textbf{41.7} & 81.5 & \textbf{69.3} & 32.4 & \textbf{45.3} & 68.0 & \textbf{34.5} & 56.6 & \textbf{63.2} & 28.4 & 52.1 \\
    512, 32, 64 &  13.50  & 52.5 & 41.2 & \textbf{82.9} & 68.9 & \textbf{34.4} & 44.7 & \textbf{69.0} & 33.6 & 58.7 & 62.6 & \textbf{30.1} & \textbf{52.6} \\ 
    \hline \hline

    Dense, iso-FLOPs &  17.90 & 52.2 & 41.0 & 79.2 & 63.4 & 31.0 & 34.7 & 61.0 & 30.5 & 60.3 & 51.8 & 25.1 & 48.2 \\

    Dense, iso-params & 12.74  & 52.2  & 42.6 & 83.3 & 70.1 & 34.8 & 46.8 & 67.0 & 35.1 & 61.7 & 63.5 & 31.8 & 53.5 \\

    \bottomrule
    \end{tabular*}
    \end{subtable}
\end{table}
\raggedbottom


\subsection{Ablation study on different rounding subroutines for token rounding} \label{sec:appendix:ablation}
We conduct ablation studies to study the effect of the different rounding subroutines on the trained MoE by TR. We compare token rounding with nearest rounding (``NR'') on per-expert token counts alongside other rounding methods. Specifically, we compare against stochastic rounding with per-expert token count (``SR''), always round up (``UP''), and always round down (``DOWN''). The results are shown in Table~\ref{tab:exp-ablation} and we find that our token rounding algorithm in general is robust to the specific rounding subroutines. 

Following Algorithm~\ref{alg:token_rounding}, for expert $e$, we denote the expert frequency from the TC sorting as $f_e$, and the last $\tileM$-divisible expert frequency as $\rounddown{f_e}$, and the next $\tileM$-divisible expert frequency as $\roundup{f_e}$. We also denote the expert scores from TC sorting as $s_e$, the expert scores from the selected tokens in $\pi_e[:\rounddown{f_e}]$ as $\rounddown{s_e}$ and the scores for $\pi_e[:\roundup{f_e}]$ as $\roundup{s_e}$. We note that all rounding algorithms only make a \textit{binary decision} between discarding TC tokens and padding EC tokens for each expert. The following are simple heuristics to perform rounding:

\begin{figure}[t]
\begin{algorithm}[H]
    \fontsize{8pt}{10pt}\selectfont

    \caption{Balanced rounding to $\tileM$-multiples via expert frequency (``Balance-f'' in Table~\ref{tab:exp-ablation}). 
    \\ This algorithm satisfies $\max_{e\in [E]} \bigl|\lceil f_e\rfloor_{\tileM} - f_e \bigr| \leq \tileM/2 $ and $\Bigl|\sum_{e=1}^E\lceil f_e\rfloor_{\tileM} - \sum_{e=1}^E f_e \Bigr| \leq \tileM/2$} \label{alg:balance-freq}
    \DontPrintSemicolon
    \SetKwInOut{Input}{Input}\SetKwInOut{Output}{Output}
    \Input{$f^\text{TC} = \{ f_e \}_{e \in [E]}$ as a list of expert frequency with TC top-$K$ routing, $\{ \lceil f_e \rceil_{\tileM} \}_{e \in [E]}$ as a list of expert frequency with TC top-$K$ routing and with potential EC padding to ensure each expert receives a multiple of $\tileM$ tokens, $\{ \lfloor f_e \rfloor_{\tileM} \}_{e \in [E]}$ as a list of expert frequency with TC top-$K$ routing and with potential token dropping to ensure each expert receives a multiple of $\tileM$ tokens; We should ensure $\max_{e\in[E]} \Bigl(\lceil f_e \rceil_{\tileM} - \lfloor f_e \rfloor_{\tileM} \Bigr) \leq \tileM $.}
    \Output{$f^\text{TR} = \{ \lceil f_e \rfloor_{\tileM} \}_{e \in [E]}$ as a list of expert frequency that ensures each expert receives a multiple of $\tileM$ tokens}

    \note{// an accumulator that ensures the preservation of total expert frequency} 
    
    $z \gets 0; $  

     \For{$e \in [E]$}{
        \note{// calculate the residual error of both rounding choice}
        
        $r_e^\text{up} \gets \lceil f_e \rceil_{\tileM} - f_e ;$ 

        $r_e^\text{down} \gets \lfloor f_e \rfloor_{\tileM} - f_e ;$ 

        \If{$ \bigl| r_e^\text{up} + z \bigr| < \bigl| r_e^\text{down} + z \bigr|$}{

            \note{// choose to pad with EC tokens} 

            $\lceil f_e\rfloor_{\tileM} \gets \lceil f_e \rceil_{\tileM};$
            
            $z \gets z + r_e^\text{up};$
        }
        \Else{

            \note{// choose to discard TC tokens} 

            $\lceil f_e\rfloor_{\tileM} \gets \lfloor f_e \rfloor_{\tileM};$
            
            $z \gets z + r_e^\text{down};$
        }

     }

    \Indm
\end{algorithm}
\end{figure}

\begin{itemize}
    \item \textbf{NR-f: nearest rounding to $\tileM$-multiples via expert frequency}: We pad EC tokens if $\roundup{f_e} - f_e < f_e - \rounddown{f_e}$. ``NR-f'' is our default choice of token rounding and we use it for Table~\ref{tab:exp-sparse},~\ref{tab:exp-T-ablation},~\ref{tab:exp-tileM-ablation}, and Figures~\ref{fig:wasted_flops} and~\ref{fig:token_rounding_speed}.

    \item \textbf{SR-f: stochastic rounding to $\tileM$-multiples via expert frequency}: We sample from $\mathrm{Bernoulli}\left(\dfrac{f_e-\rounddown{f_e}}{\tileM}\right)$ distribution for deciding whether to pad EC tokens for expert $e$.

    \item \textbf{NR-s: nearest rounding to $\tileM$-multiples via expert scores}: We sample from the following distribution for deciding whether to pad EC tokens for expert $e$:

    \vspace{-1.5em}
    \begin{equation}
        \mathrm{Bernoulli}\left(\dfrac{\sum_t s_{e,t}- \sum_t \rounddown{s_{e,t}}}{\sum_t \roundup{s_{e,t}} - \sum_t \rounddown{s_{e,t}}}\right)
    \end{equation}

    \item \textbf{Balance-f: balanced rounding to $\tileM$-multiples via expert frequency}: The Balance algorithm \citep{dwivedi2024kernel,lu2022grab,cooper2023coordinating} can be adapted to ensure the total number of routed tokens to all experts after tile-rounding is preserved regardless of the number of experts $E$. Algorithm~\ref{alg:balance-freq} is such an example that ensures

    \vspace{-2em}
    \begin{equation}
        \max_{e\in [E]} \left|\lceil f_e\rfloor_{\tileM} - f_e \right| \leq \tileM/2, \qquad \left|\sum_{e=1}^E\lceil f_e\rfloor_{\tileM} - \sum_{e=1}^E f_e \right| \leq \tileM/2 
    \end{equation}
    \vspace{-1em}

    where the other rounding subroutine will have an expected deviation of $O \left( \tileM\sqrt{E} \right)$ for $\sum_{e=1}^E\lceil f_e\rfloor_{\tileM}$.

    \item \textbf{UP: always round up expert frequency as $\roundup{f_e}$}: We always pad EC tokens chosen in the second step of sorting in Algorithm~\ref{alg:token_rounding}. This gives a model TFLOPS lower-bound for Figure~\ref{fig:token_rounding_speed}. 

    \item \textbf{DOWN: always round down expert frequency as $\rounddown{f_e}$}: We always discard TC top-$K$ tokens chosen in the first step of sorting in Algorithm~\ref{alg:token_rounding}. This gives a model TFLOPS upper-bound for Figure~\ref{fig:token_rounding_speed}.
\end{itemize}

\begin{table}[!htbp]
    \setlength{\tabcolsep}{2pt}
    \renewcommand{\arraystretch}{1.1}
    \fontsize{7pt}{8pt}\selectfont
    \centering
    \vspace{-0.6em}
    \caption{\small Evaluation of token rounding algorithms equipped with different $\mathrm{round\_and\_sparsify}$ subroutines in Algorithm~\ref{alg:token_rounding}. ``PPL'' refers to the validation perplexity at the end of training. ``Avg'' is the mean accuracy across the 11 downstream tasks. }
    \label{tab:exp-ablation}
    
    \begin{subtable}[!ht]{\textwidth}
   \caption{\textbf{0.5B params, 40B tokens, 2/64 activated} ($\bar{T}_e = 512, \ \tileM = 128$)}
    \vspace{-0.5em}
    \begin{tabular*}{\textwidth}{@{\extracolsep{\fill}}l!{\vrule width 0.4pt}c!{\vrule width 0.4pt}ccccccccccc!{\vrule width 0.4pt}c}
    \toprule
    \textbf{Method} & \textbf{PPL} & \textbf{Wino} & \textbf{SIQA} & \textbf{SciQ} & \textbf{PIQA} & \textbf{OBQA} & \textbf{HS} & \textbf{COPA} & \textbf{CSQA} & \textbf{BoolQ} & \textbf{ArcE} & \textbf{ArcC} & \textbf{Avg} \\
    \midrule \midrule
    TR (NR-f) & 15.92 & 51.4 & 41.6 & 78.4 & 65.4 & 31.6 & 38.1 & 65.0 & 31.0 & 61.1 & 57.4 & 29.1 &  50.0 \\
    TR (SR-f) & 15.93 & 50.8 & 40.9 & 77.4 & \textbf{66.9} & \textbf{33.0} & \textbf{38.4} & 64.0 & 31.1 & 60.7 & 55.8 & 28.1 & 49.7 \\
    TR (NR-s) & 15.91 & 51.3 & 40.9 & \textbf{80.3} & 65.4 & 30.8 & 37.7 & 67.0 & 31.0 & 61.6 & 55.4 & 28.4 & 50.0 \\ 
    TR (Balance-f) & 15.93 & \textbf{51.9} & \textbf{41.8} & 78.8 & 65.9 & 32.6 & \textbf{38.4} & 66.0 & 31.6 & 60.3 & 56.8 & 27.1 & 50.1 \\ \grayline
    TR (UP) & \textbf{15.89} & 50.5 & 40.9 & 78.6 & 64.5 & 32.2 & 38.2 & \textbf{68.0} & 29.9 & 55.2 & 54.2 & 30.1 & 49.3 \\  
    TR (DOWN) & 16.10 & 51.1 & 41.4 & 78.7 & 64.9 & 31.6 & 38.0 & 62.0 & \textbf{32.8} & \textbf{61.9} & \textbf{58.9} & \textbf{30.8} & \textbf{50.2}  \\ \midrule \midrule
    TC top-$K$ & 15.94 & 51.0 & 41.9 & 78.5 & 64.8 & 33.0 & 38.1 & 67.0 & 30.8 & 54.7 & 55.8 & 30.1 & 49.6 \\
    \bottomrule
    \end{tabular*}
    \end{subtable}

    \vspace{0.5em}
    \begin{subtable}[!ht]{\textwidth}
   \caption{\textbf{1.8B params, 40B tokens, 8/256 activated} ($\bar{T}_e = 512, \ \tileM = 128$)}
    \vspace{-0.5em}
    \begin{tabular*}{\textwidth}{@{\extracolsep{\fill}}l!{\vrule width 0.4pt}c!{\vrule width 0.4pt}ccccccccccc!{\vrule width 0.4pt}c}
    \toprule
    TR (NR-f) & 13.10 & 53.4 & 42.1 & 81.7 & 69.6 & 35.2 & 45.3 & \textbf{70.0} & 33.2 & \textbf{61.4} & 63.0 & 33.4 & \textbf{53.5} \\
    TR (SR-f) & 13.08 & 52.7 & 41.6 & 82.6 & 69.4 & 34.4 & 45.6 & \textbf{70.0} & 33.0 & 59.1 & 62.5 & \textbf{34.8} & 53.2 \\
    TR (NR-s) & 13.09 & 54.1 & \textbf{42.3} & \textbf{82.8} & 69.3 & 33.8 & \textbf{45.7} & \textbf{70.0} & 34.1 & 59.0 & \textbf{64.6} & 32.4 & \textbf{53.5} \\ 
    TR (Balance-f) & 13.08 & 52.5 & 42.0 & 82.7 & \textbf{70.0} & 33.2 & 45.3 & 68.0 & \textbf{34.6} & 59.4 & 63.3 & 33.4 & 53.1 \\ \grayline
    TR (UP) & \textbf{13.07} & 50.4 & 41.7 & 81.4 & 68.4 & \textbf{37.2} & 45.4 & 69.0 & 31.9 & 51.7 & 62.2 & 33.4 & 52.1  \\ 
    TR (DOWN) & 13.19 & \textbf{55.4} & 41.6 & 82.2 & 68.6 & 34.8 & 45.0 & 69.0 & 34.0 & 54.4 & 63.5 & 31.4 & 52.7  \\ \midrule \midrule
    TC top-$K$ & 13.12 & 50.1 & 42.9 & 81.3 & 69.8 & 33.8 & 45.2 & 71.0 & 34.1 & 56.7 & 64.6 & 31.1 & 52.8 \\
    \bottomrule
    \end{tabular*}
    \end{subtable}
\end{table}
\raggedbottom

\paragraph{Always discarding TC tokens (``DOWN'').} ``DOWN'' is a baseline in which we always drop the last TC tile if the expert frequency is not $\tileM$-divisible. This idea is similar to the idea of \textit{token dropping} in expert parallelism where the expert will sort and drop the token with the lowest scores when it receives too many tokens \citep{fedus2022switchtransformersscalingtrillion}. We note that ``DOWN'' produces the shortest MoE kernel runtime for any rounding algorithm. However, in Table~\ref{tab:exp-ablation}, we observe that ``DOWN'' yields a much higher validation perplexity than ``NR-f'', ``SR-f'' and ``NR-s''. Although we can expect a shorter MoE kernel runtime by always discarding TC tokens, such quality degradation might not be acceptable in practice.

\paragraph{Always padding EC tokens (``UP'').} ``UP'' is a baseline in which we always pad extra EC tokens to the last TC tile if the expert frequency is not $\tileM$-divisible. Contrary to ``DOWN'', ``UP'' produces the longest MoE kernel runtime for any rounding algorithm. In Table~\ref{tab:exp-ablation}, we find that ``UP'' often produces lower validation perplexity, but the average downstream task accuracy is not necessarily higher than other rounding algorithms. Given the longer MoE kernel runtime but not necessarily better trained MoE quality, we do not recommend the usage of always rounding up. \textit{We speculate this is due to the train-test gap between TC and EC routing and ``UP'' reinforces the bias towards EC more strongly than other TR algorithms.}

For a balance between training efficiency and trained MoE quality, neither always discarding TC tokens nor padding EC tokens is the right solution. In Table~\ref{tab:exp-sparse}, we pick ``NR-f'' as the $\mathrm{round\_and\_sparsify}$ subroutine for TR's main experiments.

\subsection{Ablation study on the effects of microbatch size $T$ and tile size $\tileM$} \label{sec:appendix:ablation-microbatch-tileM}

\paragraph{Effect of microbatch size $T$.} Since the token rounding is applied on the microbatch level, the choice of microbatch size $T$ will result in different qualitative results for TR. Note that this also holds true for EC routing. For example, EC over sequence will result in different model quality as EC over a text segment. In Table~\ref{tab:exp-T-ablation}, we vary the microbatch size while keeping the minibatch size (consumed tokens per optimization step) constant. 

We find that TR will preserve its trained MoE quality when $\bar{T}_e/\tileM \ge 2$, but if $\bar{T}_e/\tileM = 1$ (the last row in both subtables), there is a noticeable quality degradation for both validation perplexity and downstream task performance. However, the trained MoE quality with $\bar{T}_e/\tileM = 1$ is still better than training with EC and finetuning with TC top-$K$ routing.

\paragraph{Effect of the tile quantization size $\tileM$.} Similarly in Table~\ref{tab:exp-tileM-ablation}, we can find that TR is generally robust w.r.t. $\tileM$ when $\bar{T}_e/\tileM \geq 2$, and when $\bar{T}_e/\tileM=1$ there is a noticeable degradation but the overall result is still better than EC baseline. 

\begin{table}[!ht]
    \setlength{\tabcolsep}{2pt}
    \renewcommand{\arraystretch}{1.1}
    \fontsize{7pt}{8pt}\selectfont
    \centering
    \vspace{-0.6em}
    \caption{\small Evaluation of token rounding algorithms when we vary microbatch size $T$ to change average number of tokens per expert ($\bar{T}_e$). For each trial, we vary the microbatch size from 4 ($\bar{T}_e=512$) to 1 ($\bar{T}_e=128$) and keep minibatch size constant. The $\tileM$ is always kept as 128. ``PPL'' refers to the validation perplexity at the end of training. ``Avg'' is the mean accuracy across the 11 downstream tasks. }
    \label{tab:exp-T-ablation}
    
    \begin{subtable}[!ht]{\textwidth}
   \caption{\textbf{0.5B params, 40B tokens, 2/64 activated} ($\tileM = 128$)}
    \vspace{-0.5em}
    \begin{tabular*}{\textwidth}{@{\extracolsep{\fill}}l!{\vrule width 0.4pt}c!{\vrule width 0.4pt}ccccccccccc!{\vrule width 0.4pt}c}
    \toprule
    \textbf{Method} & \textbf{PPL} & \textbf{Wino} & \textbf{SIQA} & \textbf{SciQ} & \textbf{PIQA} & \textbf{OBQA} & \textbf{HS} & \textbf{COPA} & \textbf{CSQA} & \textbf{BoolQ} & \textbf{ArcE} & \textbf{ArcC} & \textbf{Avg} \\
    \midrule \midrule
    TR ($\bar{T}_e=1024$) & \textbf{15.91} & \textbf{52.9} & 41.0 & \textbf{80.1} & 65.1 & 31.0 & 37.9 & 63.0 & \textbf{32.3} & 59.3 & 54.9 & 28.1 & 49.6 \\
    \textbf{TR ($\bar{T}_e=512$)} & 15.92 & 51.4 & 41.6 & 78.4 & 65.4 & 31.6 & \textbf{38.1} & 65.0 & 31.0 & 61.1 & \textbf{57.4} & 29.1 &  50.0 \\
    TR ($\bar{T}_e=256$) & 15.98 & 52.2 & 41.4 & 77.7 & \textbf{66.1} & \textbf{32.2} & 37.9 & 66.0 & 31.0 & 59.6 & 57.2 & \textbf{30.1} & \textbf{50.1} \\
    TR ($\bar{T}_e=128$) & 16.11 & 51.7 & \textbf{41.7} & 77.9 & \textbf{66.1} & 30.8 & 37.7 & \textbf{67.0} & 31.9 & \textbf{61.2} & 54.7 & 29.1 & 50.0 \\\midrule\midrule
    TC top-$K$ & 15.94 & 51.0 & 41.9 & 78.5 & 64.8 & 33.0 & 38.1 & 67.0 & 30.8 & 54.7 & 55.8 & 30.1 & 49.6 \\\grayline
    EC (ft TC router) & 16.98 & 50.0 & 41.7 & 79.7 & 64.9 & 31.6 & 36.8 & 63.0 & 32.1 & 60.7 & 54.6 & 27.4 & 49.3 \\
    \bottomrule
    \end{tabular*}
    \end{subtable}

    \vspace{0.5em}
    \begin{subtable}[!ht]{\textwidth}
   \caption{\textbf{1.8B params, 40B tokens, 8/256 activated} ($\tileM = 128$)}
    \vspace{-0.5em}
    \begin{tabular*}{\textwidth}{@{\extracolsep{\fill}}l!{\vrule width 0.4pt}c!{\vrule width 0.4pt}ccccccccccc!{\vrule width 0.4pt}c}
    \toprule
    TR ($\bar{T}_e=1024$) & \textbf{13.08} & 51.5 & 42.0 & 81.7 & 68.9 & 34.8 & \textbf{45.7} & 72.0 & 32.6 & 59.5 & 61.4 & 32.1 & 52.9  \\
    \textbf{TR ($\bar{T}_e=512$)} & 13.10 & \textbf{53.4} & \textbf{42.1} & 81.7 & 69.6 & \textbf{35.2} & 45.3 & 70.0 & 33.2 & \textbf{61.4} & 63.0 & 33.4 & \textbf{53.5} \\
    TR ($\bar{T}_e=256$) &  13.12 & 51.9 & 41.2 & 81.8 & \textbf{69.7} & 33.6 & 45.2 & \textbf{73.0} & 34.2 & 56.9 & 63.2 & \textbf{34.1} & 53.2  \\
    TR ($\bar{T}_e=128$) & 13.55 & 51.9 & 41.5 & \textbf{82.0} & 69.2 & 32.8 & 44.4 & 69.0 & \textbf{34.4} & 59.8 & \textbf{64.0} & 30.4 & 52.7  \\\midrule\midrule
    TC top-$K$ & 13.12 & 50.1 & 42.9 & 81.3 & 69.8 & 33.8 & 45.2 & 71.0 & 34.1 & 56.7 & 64.6 & 31.1 & 52.8 \\\grayline
    EC (ft TC router) & 15.01 & 52.7 & 41.1 & 79.6 & 66.9 & 30.6 & 40.2 & 66.0 & 31.9 & 60.5 & 57.2 & 30.8 & 50.7 \\
    \bottomrule
    \end{tabular*}
    \end{subtable}
\end{table}
\raggedbottom

\begin{table}[!ht]
    \setlength{\tabcolsep}{2pt}
    \renewcommand{\arraystretch}{1.1}
    \fontsize{7pt}{8pt}\selectfont
    \centering
    \vspace{-0.6em}
    \caption{\small Evaluation of token rounding algorithms when we vary the size of tile $\tileM$ for token rounding. ``PPL'' refers to the validation perplexity at the end of training. ``Avg'' is the mean accuracy across the 11 downstream tasks. }
    \label{tab:exp-tileM-ablation}
    
    \begin{subtable}[!ht]{\textwidth}
   \caption{\textbf{0.5B params, 40B tokens, 2/64 activated ($\bar{T}_e = 512$)}}
    \vspace{-0.5em}
    \begin{tabular*}{\textwidth}{@{\extracolsep{\fill}}l!{\vrule width 0.4pt}c!{\vrule width 0.4pt}ccccccccccc!{\vrule width 0.4pt}c}
    \toprule
    \textbf{Method} & \textbf{PPL} & \textbf{Wino} & \textbf{SIQA} & \textbf{SciQ} & \textbf{PIQA} & \textbf{OBQA} & \textbf{HS} & \textbf{COPA} & \textbf{CSQA} & \textbf{BoolQ} & \textbf{ArcE} & \textbf{ArcC} & \textbf{Avg} \\
    \midrule \midrule
    TR ($\tileM=64$) & \textbf{15.90} & 51.3 & \textbf{41.7} & 78.1 & \textbf{65.6} & 31.4 & 37.9 & \textbf{67.0} & \textbf{32.4} & 59.8 & 57.2 & 28.8 & 50.1 \\
    \textbf{TR ($\tileM=128$)} & 15.92 & 51.4 & 41.6 & 78.4 & 65.4 & 31.6 & \textbf{38.1} & 65.0 & 31.0 & \textbf{61.1} & 57.4 & 29.1 &  50.0 \\
    TR ($\tileM=256$) & 16.00 & 51.7 & 41.4 & 78.7 & 66.3 & \textbf{32.4} & 37.7 & \textbf{67.0} & 31.3 & 60.1 & \textbf{58.2} & 29.1 & \textbf{50.4}  \\
    TR ($\tileM=512$) & 16.17 & \textbf{52.5} & 41.2 & \textbf{80.2} & 65.2 & 32.0 & 37.9 & 62.0 & 31.0 & 59.4 & 57.2 & \textbf{30.4} & 49.9  \\\midrule\midrule
    TC top-$K$ & 15.94 & 51.0 & 41.9 & 78.5 & 64.8 & 33.0 & 38.1 & 67.0 & 30.8 & 54.7 & 55.8 & 30.1 & 49.6 \\\grayline
    EC (ft TC router) & 16.98 & 50.0 & 41.7 & 79.7 & 64.9 & 31.6 & 36.8 & 63.0 & 32.1 & 60.7 & 54.6 & 27.4 & 49.3 \\
    \bottomrule
    \end{tabular*}
    \end{subtable}

    \vspace{0.5em}
    \begin{subtable}[!ht]{\textwidth}
   \caption{\textbf{1.8B params, 40B tokens, 8/256 activated ($\bar{T}_e = 512$)}}
    \vspace{-0.5em}
    \begin{tabular*}{\textwidth}{@{\extracolsep{\fill}}l!{\vrule width 0.4pt}c!{\vrule width 0.4pt}ccccccccccc!{\vrule width 0.4pt}c}
    \toprule
    TR ($\tileM=64$) & \textbf{13.07} & 52.3 & \textbf{42.9} & \textbf{82.7} & 69.4 & \textbf{35.4} & \textbf{45.6} & \textbf{70.0} & 32.4 & 56.6 & 64.4 & 31.4 & 53.0 \\
    \textbf{TR ($\tileM=128$)} & 13.10 & \textbf{53.4} & 42.1 & 81.7 & \textbf{69.6} & 35.2 & 45.3 & \textbf{70.0} & 33.2 & \textbf{61.4} & 63.0 & \textbf{33.4} & \textbf{53.5} \\
    TR ($\tileM=256$) & 13.13 & 52.0 & 41.6 & 82.1 & 69.2 & \textbf{35.4} & 45.3 & 69.0 & \textbf{34.2} & 58.0 & \textbf{65.6} & 32.1  & 53.1  \\
    TR ($\tileM=512$) & 13.56 &  53.0 & 41.8 & 81.2 & 68.4 & 34.0 & 44.2 & 68.0 & 33.3 & 58.1 & 59.5 & 30.1 & 52.0  \\\midrule\midrule
    TC top-$K$ & 13.12 & 50.1 & 42.9 & 81.3 & 69.8 & 33.8 & 45.2 & 71.0 & 34.1 & 56.7 & 64.6 & 31.1 & 52.8 \\\grayline
    EC (ft TC router) & 15.01 & 52.7 & 41.1 & 79.6 & 66.9 & 30.6 & 40.2 & 66.0 & 31.9 & 60.5 & 57.2 & 30.8 & 50.7 \\
    \bottomrule
    \end{tabular*}
    \end{subtable}
\end{table}
\raggedbottom

\section{Activation memory and training throughput benchmark configurations} \label{sec:appendix:config}


The configurations of Figure~\ref{fig:mem} and~\ref{fig:speed} are included in Table~\ref{tab:config_benchmark}.



\begin{table}[!ht]
\caption{Benchmark configurations used by Figure~\ref{fig:mem}, ~\ref{fig:speed}, and~\ref{fig:B300-speed}, and all other kernel-level ablation studies on H100 and B300 GPUs.}
\begin{subtable}[t]{0.4\textwidth}
\centering
\fontsize{7pt}{10pt}\selectfont
\setlength{\tabcolsep}{3pt}
\renewcommand{\arraystretch}{1.1}
\caption{Benchmark configurations used by Figure~\ref{fig:mem} and~\ref{fig:speed}, and all other ablation studies on H100 GPUs.}
\label{tab:config_benchmark}
\begin{tabular}{l!{\vrule width 0.4pt}ccccc}
\toprule
Model Size & $T$ & $d$ & $n$ & $E$ & $K$ \\
\midrule
\multirow{3}{*}{1.4B}
& 40960 & 768  & 256  & 128 & 8  \\
& 40960 & 768  & 512  & 64 & 4  \\
& 40960 & 768  & 1024 & 32  & 2  \\
\midrule
\multirow{3}{*}{7B}
& 24576 & 1536 & 256  & 128 & 8  \\
& 24576 & 1536 & 512  & 64 & 4  \\
& 24576 & 1536 & 1024 & 32  & 2  \\
\midrule
\multirow{3}{*}{30B}
& 32768 & 4096 & 256  & 256 & 16 \\
& 32768 & 4096 & 512  & 128 & 8  \\
& 32768 & 4096 & 1024 & 64  & 4  \\
\midrule
\multirow{3}{*}{120B}
& 32768 & 4096 & 512  & 256 & 16 \\
& 32768 & 4096 & 1024 & 128 & 8  \\
& 32768 & 4096 & 2048 & 64  & 4  \\
\bottomrule
\end{tabular}
\end{subtable}
\hfill
\begin{subtable}[t]{0.4\textwidth}
\centering
\fontsize{7pt}{10pt}\selectfont
\setlength{\tabcolsep}{3pt}
\renewcommand{\arraystretch}{1.1}
\caption{Benchmark configurations used by Figure~\ref{fig:B300-speed}, and all other ablation studies on B300 GPUs.}
\begin{tabular}{l!{\vrule width 0.4pt}ccccc}
\toprule
Model Size & $T$ & $d$ & $n$ & $E$ & $K$ \\
\midrule
\multirow{3}{*}{1.4B}
& 131072 & 768  & 256  & 128 & 8  \\
& 131072 & 768  & 512  & 64 & 4  \\
& 131072 & 768  & 1024 & 32  & 2  \\
\midrule
\multirow{3}{*}{7B}
& 81920 & 1536 & 256  & 128 & 8  \\
& 81920 & 1536 & 512  & 64 & 4  \\
& 81920 & 1536 & 1024 & 32  & 2  \\
\midrule
\multirow{3}{*}{30B}
& 32768 & 4096 & 256  & 256 & 16 \\
& 32768 & 4096 & 512  & 128 & 8  \\
& 32768 & 4096 & 1024 & 64  & 4  \\
\midrule
\multirow{3}{*}{120B}
& 32768 & 4096 & 512  & 256 & 16 \\
& 32768 & 4096 & 1024 & 128 & 8  \\
& 32768 & 4096 & 2048 & 64  & 4  \\
\bottomrule
\end{tabular}
\end{subtable}
\end{table}

The configurations for the 4 subfigures in Figure~\ref{fig:token_rounding_speed} are listed below. Notice that we consistently use $\tileM$ as 128 when we benchmark the TR's speed.
\begin{itemize}
    \item \textbf{Top-left 2 subfigures}: We use $(T,d,n,K)=(16384,1536,256,8)$ and we vary $E$ from 64 to 512.

    \item \textbf{Top-right 2 subfigures}: We use $(T,d,n,K)=(16384,1536,1024,2)$ and we vary $E$ from 16 to 128.

    \item \textbf{Bottom-left 2 subfigures}: We use $(T,d,n,K)=(16384,4096,512,8)$ and we vary $E$ from 64 to 512.

    \item \textbf{Bottom-right 2 subfigures}: We use $(T,d,n,K)=(16384,4096,1024,4)$ and we vary $E$ from 32 to 256.
\end{itemize}

\section{Hyperparameter details for LM training} \label{sec:appendix:exp-details}
We use the OLMoE codebase and its downstream tasks in the default configuration\footnote{\scriptsize \url{https://github.com/allenai/OLMoE/blob/357454f4f647385839c0ff6b99a688dc7cd9c13f/configs/OLMoE-1B-7B-0924.yml}} except for MMLU: WinoGrande (``wino'')~\citep{sakaguchi2020winogrande}, Social IQA (``SIQA'')~\citep{sap2019socialiqa}, SciQ~\citep{SciQ}, PIQA~\citep{piqa}, OpenBookQA (``OBQA'')~\citep{OpenBookQA2018}, HellaSwag (``HS'')~\citep{zellers2019hellaswag}, COPA~\citep{roemmele2011choice}, CommonsenseQA (``CSQA'')~\citep{csqa}, BoolQ~\citep{boolq}, Arc-Easy and Arc-Challenge (``ArcE'' and ``ArcC'')~\citep{arc} datasets. We use a deduplicated version of FineWeb-Edu \citep{benallal2024smollmcorpus}\footnote{\scriptsize \url{https://huggingface.co/datasets/HuggingFaceTB/smollm-corpus}} as the pretraining corpus, and train all models with a context length of 4096 tokens.

We always use MoE with SwiGLU for the MoE layers and we use an auxiliary load balancing loss \citep{shazeer2017outrageouslylargeneuralnetworks} with coefficient 0.01 but we do not use the router Z loss \citep{zoph2022st}. Our attention block architecture is identical to OLMoE's attention block. We always tie the weight of the LM head with the weight of the token embedding matrix. 

\begin{table}[!ht]
\setlength{\tabcolsep}{3pt}
\renewcommand{\arraystretch}{1.15}
\fontsize{7pt}{8pt}\selectfont
\centering
\vspace{-0.6em}
\caption{Common configurations for MoE pretraining experiment}

\label{tab:config_tr}
\begin{tabular}{l!{\vrule width 0.4pt}ccccccccccc}
    \toprule
    \textbf{Config name in Tables~\ref{tab:exp-sparse} and~\ref{tab:exp-ablation}} & \textbf{\# layers} & \textbf{\# attn heads} & \textbf{$d$} & \textbf{$n$} & \textbf{$E$} & \textbf{$K$} & \textbf{\# tokens in a minibatch} & \textbf{LR} & \textbf{WD} & \textbf{LR scheduler}  \\
    
    \midrule
    
    0.5B params, 20B tokens, 8/64 activated
     & 12 & 12 & 768 & 256 & 64  & 8  & 0.5M & 6e-4 & 0.01 & cosine w/. warmup (10\% steps) \\
    
    0.5B params, 40B tokens, 2/64 activated
     & 12 & 12 & 768 & 256 & 64  & 2 & 1M & 6e-4 & 0.01 & cosine w/. warmup (10\% steps) \\
    
    1.8B params, 40B tokens, 8/256 activated
     & 12 & 12 & 768 & 256 & 256  & 8  & 1M& 6e-4 & 0.01 & cosine w/. warmup (10\% steps)  \\
    
    1.4B params, 50B tokens, 8/128 activated
     & 18 & 12 & 768 & 256 & 128  & 8  & 1M& 4e-4 & 0.01 & cosine w/. warmup (10\% steps) \\
    
    1.4B params, 100B tokens, 2/128 activated  & 18 & 12 & 768 & 256 & 128  & 2  & 2M & 4e-4 & 0.01 & cosine w/. warmup (10\% steps) \\
    
    \bottomrule
\end{tabular}
\end{table}


For all EC with finetuned TC router experiments in Table~\ref{tab:exp-sparse}, we use an additional 4B tokens and we only finetune the router weights with TC top-$K$ routing (all other parameters are frozen). We always use a learning rate of 2e-4, weight decay of 0.01 and cosine learning rate scheduler with 10\% warmup steps. The number of tokens per minibatch during finetuning is 1M. We disable auxiliary load balancing loss during TC finetuning. 

For all EC with auxiliary router experiments, we use a 2-layer MLP (each linear layer has size $E\times E$ with SiLU activation) which takes as input the raw router logits and makes $E$ independent binary predictions for all experts. We compute the averaged binary cross entropy loss over $E$ labels using the multi-label prediction loss, and scale the loss by 0.01. During the evaluation, we will let the EC router compute the raw logits and raw scores and let the auxiliary router mask the token-expert pair with its own confidence score.

We implement ``TC (token drop)'' by discarding tokens selected from the TC top-$K$ sorting.


\end{document}